\documentclass{article}

\usepackage{microtype}
\usepackage{graphicx}
\usepackage{subcaption}

\usepackage{hyperref}
\usepackage{colortbl}

\usepackage[preprint]{icml2026}

\usepackage{amsmath}
\usepackage{amssymb}
\usepackage{mathtools}
\usepackage{amsthm}
\usepackage{hyphenat}

\usepackage{threeparttable}
\usepackage[table]{xcolor}
\usepackage{array}
\usepackage{booktabs,multirow,makecell,pifont,siunitx}

\definecolor{highlightrow}{gray}{0.92}
\definecolor{bestblue}{RGB}{0, 102, 204}
\definecolor{oursrow}{RGB}{235, 240, 250} 
\definecolor{oursblue}{RGB}{235, 245, 255}
\definecolor{badcaseorange}{RGB}{255, 245, 235} 
\definecolor{improvegreen}{RGB}{240, 255, 240}
\definecolor{textred}{RGB}{200, 0, 0}
\definecolor{textblue}{RGB}{0, 0, 150}

\definecolor{mydarkgray}{gray}{0.4}
\definecolor{mylightgray}{gray}{0.7}
\newcommand{\cmark}{\textcolor{mydarkgray}{\ding{51}}}
\newcommand{\xmark}{\textcolor{mylightgray}{\ding{55}}}

\newcommand{\best}[1]{\textbf{#1}}
\newcommand{\secbest}[1]{\underline{#1}}

\newcommand{\Enc}{\mathrm{Enc}}
\newcommand{\Dec}{\mathrm{Dec}}
\newcommand{\edgeemb}{\mathbf{e}} 
\newcommand{\VectorWorld}{VectorWorld}
\newcommand{\DeltaSim}{\(\Delta\)Sim}
\newcommand{\EdgeDiT}{edge-gated relational DiT}
\newcommand{\MotionVAE}{motion-aware gated VAE}
\newcommand{\MeanFlow}{MeanFlow}
\newcommand{\InteractionState}{interaction-state interface}

\usepackage[capitalize,noabbrev]{cleveref}

\theoremstyle{plain}

\theoremstyle{definition}

\theoremstyle{remark}

\newcommand{\projectpage}{%
  \href{https://jiangchaokang.github.io/VectorWorld/}{\nolinkurl{jiangchaokang.github.io/VectorWorld/}}%
}

\usepackage[textsize=tiny]{todonotes}

\icmltitlerunning{VectorWorld: Efficient Streaming World Model via Diffusion Flow on Vector Graphs}

\begin{document}

\twocolumn[
  \icmltitle{VectorWorld: \texorpdfstring{\\}{ } Efficient Streaming World Model via Diffusion Flow on Vector Graphs}




  \begin{icmlauthorlist}
    \icmlauthor{Chaokang Jiang}{bosch}
    \icmlauthor{Desen Zhou}{bosch}
    \icmlauthor{Jiuming Liu}{cambridge}
    \icmlauthor{Kevin Li Sun}{bosch}
  \end{icmlauthorlist}

  \icmlaffiliation{bosch}{Bosch, Shanghai, Shanghai, China}
  \icmlaffiliation{cambridge}{University of Cambridge, Cambridge, United Kingdom}

  \icmlcorrespondingauthor{Kevin Li Sun}{sen5sgh@bosch.com}

  \icmlkeywords{Machine Learning, ICML}

  \vskip 0.3in
]



\printAffiliationsAndNotice{Project page: \projectpage.}  

\begin{abstract}
Closed-loop evaluation of autonomous-driving policies requires interactive simulation beyond log replay. However, existing generative world models often degrade in closed loop due to (i) history-free initialization that mismatches policy inputs, (ii) multi-step sampling latency that violates real-time budgets, and (iii) compounding kinematic infeasibility over long horizons. We propose VectorWorld, a streaming world model that incrementally generates ego-centric $64 \mathrm{m}\times 64\mathrm{m}$ lane--agent vector-graph tiles during rollout. VectorWorld aligns initialization with history-conditioned policies by producing a policy-compatible interaction state via a motion-aware gated VAE. It enables real-time outpainting via solver-free one-step masked completion with an edge-gated relational DiT trained with interval-conditioned MeanFlow and JVP-based large-step supervision. To stabilize long-horizon rollouts, we introduce $\Delta$Sim, a physics-aligned non-ego (NPC) policy with hybrid discrete--continuous actions and differentiable kinematic logit shaping. On Waymo open motion and nuPlan, VectorWorld improves map-structure fidelity and initialization validity, and supports stable, real-time $1\mathrm{km}+$ closed-loop rollouts (\href{https://github.com/jiangchaokang/VectorWorld}{code}).
\end{abstract}

\begin{figure}[!ht]
  \centering
  \includegraphics[width=1.0\linewidth]{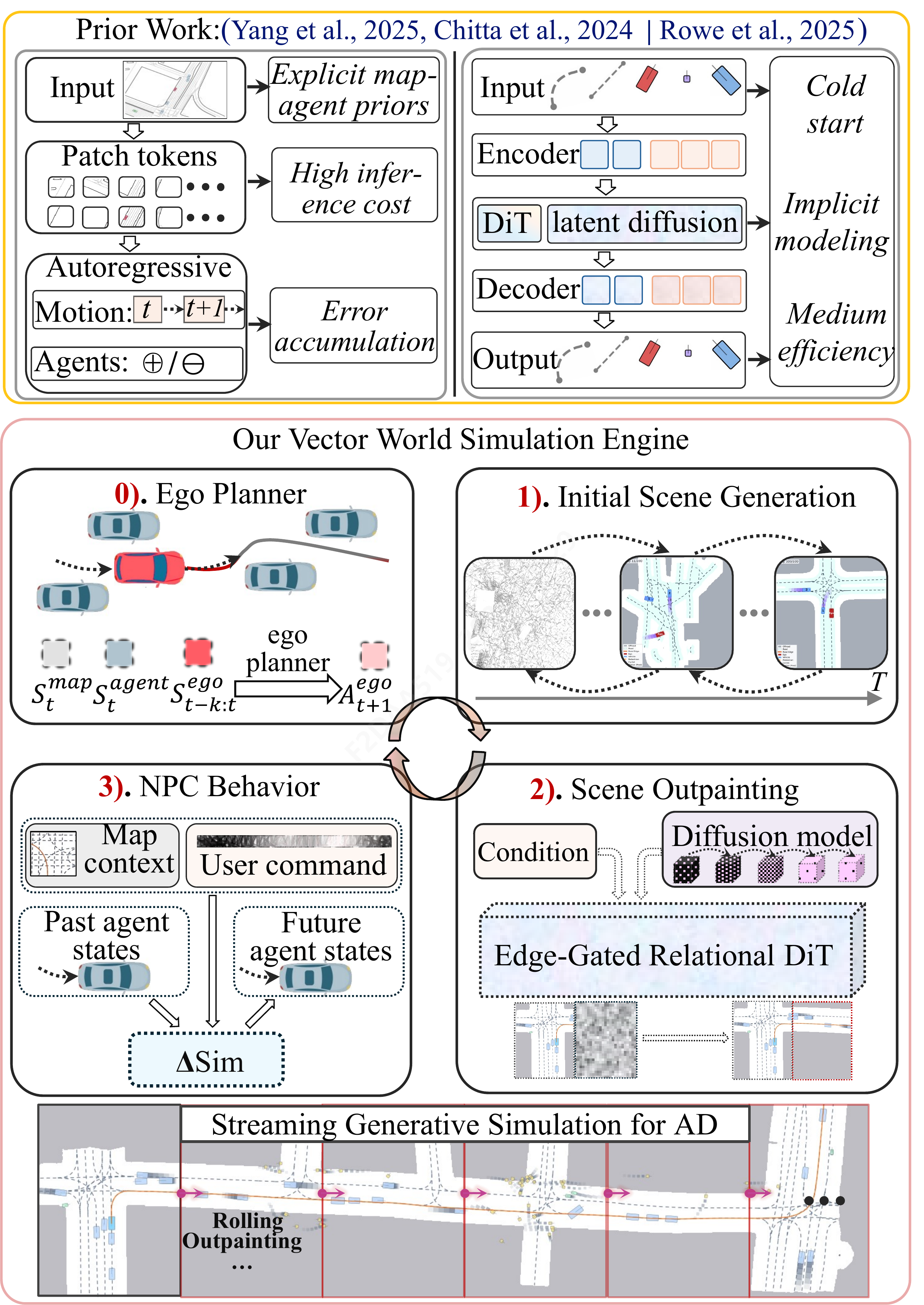}
  \caption{\textbf{Comparison with Prior Works.} Unlike prior pipelines that degrade in closed loop due to accumulated drift (autoregressive rollout) or history-free cold-start initialization, \VectorWorld supports kilometer-scale closed-loop simulation via streaming outpainting in our experiments. The engine operates in a closed loop: 0) An Ego Planner drives within the scene; 1) Initial Scene Generation creates the base tile; 2) Scene Outpainting dynamically extends the map frontier using \EdgeDiT; and 3) NPC Behavior (\DeltaSim) ensures reactive interactions. This mechanism enables consistent, kilometer-scale driving simulation by progressively outpainting the world ahead.}
  \label{fig:system}
\end{figure}

\vspace{-2mm}
\section{Introduction}
\label{sec:intro}

Closed-loop evaluation and training of autonomous-driving policies require interactive simulations that extend beyond recorded logs.
Log replay can neither answer counterfactual queries nor extend the horizon once the recorded context ends~\cite{caesar2021nuplan,gulino2023waymax}. Although generative world models~\cite{ha2018world,feng2023trafficgen,tan2023language,yang2025wcdt} can synthesize diverse lane--agent scenes, deploying them as an \emph{online} simulator backend introduces constraints (initialization, latency, and long-horizon stability) that standard open-loop metrics do not capture~\cite{tan2025scenediffuserpp}.

We identify three deployment constraints that dominate closed-loop simulation but are weakly reflected by standard open-loop metrics. First, history-free initialization: many generators output a static snapshot at \(t=0\)~\cite{rowe2025scenario}, while reactive policies are conditioned on short motion histories, inducing cold-start transients (e.g., jerk spikes) and unstable early interactions. Second, sampling latency: diffusion-based solvers require many denoising steps \cite{chitta2024sledge,tan2025scenediffuserpp} for structured lane--agent graph generation, which is incompatible with streaming outpainting under a real-time budget. Third, long-horizon feasibility: minor per-step kinematic violations, while negligible in short-horizon metrics, accumulate over kilometer-scale rollouts through compounding drift \cite{rowe2024ctrlsim, lin2025causal}.

To address these constraints, we propose \VectorWorld\ (Figure~\ref{fig:system}), a streaming world model built on vector-graph representations.
For initialization, we introduce an \InteractionState\ that augments each agent with a compact motion-history code learned by a \MotionVAE, enabling warm-start inputs for history-conditioned policies.
For real-time outpainting, we propose an \EdgeDiT\ trained with interval-conditioned \MeanFlow\ \cite{geng2025mean} and a JVP-based large-step objective, enabling solver-free (without iterative diffusion/ODE solvers) one-step latent sampling while preserving lane connectivity and agent--lane relations.


Overall, our contributions are as follows:
\vspace{-3mm}
\begin{itemize}
    \item An \InteractionState\ for lane--agent world models, implemented by a \MotionVAE\ that selectively encodes motion history for warm-start initialization.
    \vspace{-2mm}
    \item An \EdgeDiT\ for heterogeneous vector graph generation with edge-conditioned attention bias and value gating to preserve lane topology and cross-type relations.
    \vspace{-2mm}
    \item Interval-conditioned \MeanFlow\ with JVP-based large-step supervision to enable high-fidelity solver-free one-step completion for streaming outpainting.
    \vspace{-2mm}
    \item A physics-aligned NPC policy, \DeltaSim, that stabilizes long-horizon multi-agent rollouts via hybrid actions and differentiable kinematic shaping.
    \vspace{-2mm}
    \item Experiments on Waymo Open Motion and nuPlan demonstrate that \VectorWorld\ improves map-structure fidelity and initialization validity, supports stable \(1\,\mathrm{km}+\) closed-loop rollouts, and operates at \(\approx 6\)\,ms per \(64\,\mathrm{m}\times 64\,\mathrm{m}\) tile. Policy retraining in \VectorWorld\ increases stress-test success from 25.0\% to 56.0\%.
\end{itemize}


\section{Related Work}
\paragraph{Driving World Models and Vectorized Scene Generation.}
Recent driving world models generate future observations from history sensor inputs (e.g., videos) \cite{gao2024vista,chen2025drivinggpt,ni2025maskgwm}, while other approaches generate simulation environments from perception outputs, such as bounding boxes and lane graphs \cite{chitta2024sledge,sun2024drivescenegen,jiang2024scenediffuser,tan2025scenediffuserpp}. Several methods focus on initial scene generation \cite{rowe2025scenario}, generating agents, lanes, or both, sometimes via rasterization, which increases computational cost. Vectorized representations \cite{jiang2023vad,chen2024vadv2} better match map geometry and can be more efficient for structured generation. Scenario-centric systems further combine initial scene generation with closed-loop behavior simulation, often decoupling the two for flexibility. Our approach complements these lines by introducing an \emph{interaction-state} representation that reduces initialization mismatches, and by enabling \emph{one-step} (or few-step) graph-latent generation suitable for online inpainting and outpainting. Figure~\ref{fig:system} summarizes these deployment gaps (cold-start mismatch, solver latency, and long-horizon drift) and positions VectorWorld relative to representative prior pipelines.

\paragraph{Fast Sampling in Diffusion and Flow-based Methods.}
Denoising diffusion probabilistic models (DDPM) \cite{ho2020denoising} and classifier-free guidance \cite{ho2022classifier} have become a standard approach for high-fidelity generation, and latent diffusion \cite{podell2023sdxl} reduces sampling costs by operating in a compressed space. Transformer backbones such as DiT \cite{peebles2023scalable} improve scalability and conditioning flexibility. A large body of work accelerates sampling, including DDIM \cite{song2020denoising}, DPM-Solver \cite{lu2022dpm}, EDM \cite{karras2022elucidating}, and one-step consistency-style methods \cite{song2023consistency}. Flow Matching \cite{lipman2022flow} and Rectified Flow (straight-line probability paths) \cite{lee2024improving} provide a training alternative that often admits fewer solver steps. MeanFlow \cite{geng2025mean} aims to learn interval-conditioned velocity parameterizations that enable solver-free one-step sampling.
Our work differs in two ways: (i) we operate on \emph{heterogeneous vector-graph} latents for maps and agents, and (ii) we unify DDPM/Flow/MeanFlow in one \EdgeDiT\ backbone with dual-time conditioning, using a JVP-augmented MeanFlow objective to make one-step structured completion practical.

\paragraph{Closed-Loop Behavior Simulation.}
Closed-loop multi-agent simulation requires reactive behavior models that remain stable under rollout. Prior work spans learned agent simulators \cite{igl2022symphony,suo2023mixsim,yang2025long}, next-token traffic modeling \cite{philion2024trajeglish}, and return-conditioned or reactive policies such as CtRL-Sim \cite{rowe2024ctrlsim}. While large Transformers and diffusion-based policies can improve realism, their autoregressive paradigm can incur high costs for long rollouts. Classical simulators (e.g., CARLA \cite{dosovitskiy2017carla}, SUMO \cite{krajzewicz2012sumo}) and procedural systems (e.g., MetaDrive \cite{li2021metadrive}) often rely on heuristics for agent insertion/removal, which can miss complex real-world interactions. Compared to them, VectorWorld focuses on stabilizing the full closed loop by (i) aligning the world-model output with policy inputs via interaction states, and (ii) enforcing feasibility via differentiable physics-guided logit shaping and inference-time constraints, with training--inference alignment to mitigate exposure bias.

\begin{figure}[t]
  \centering
  \includegraphics[width=0.89\linewidth]{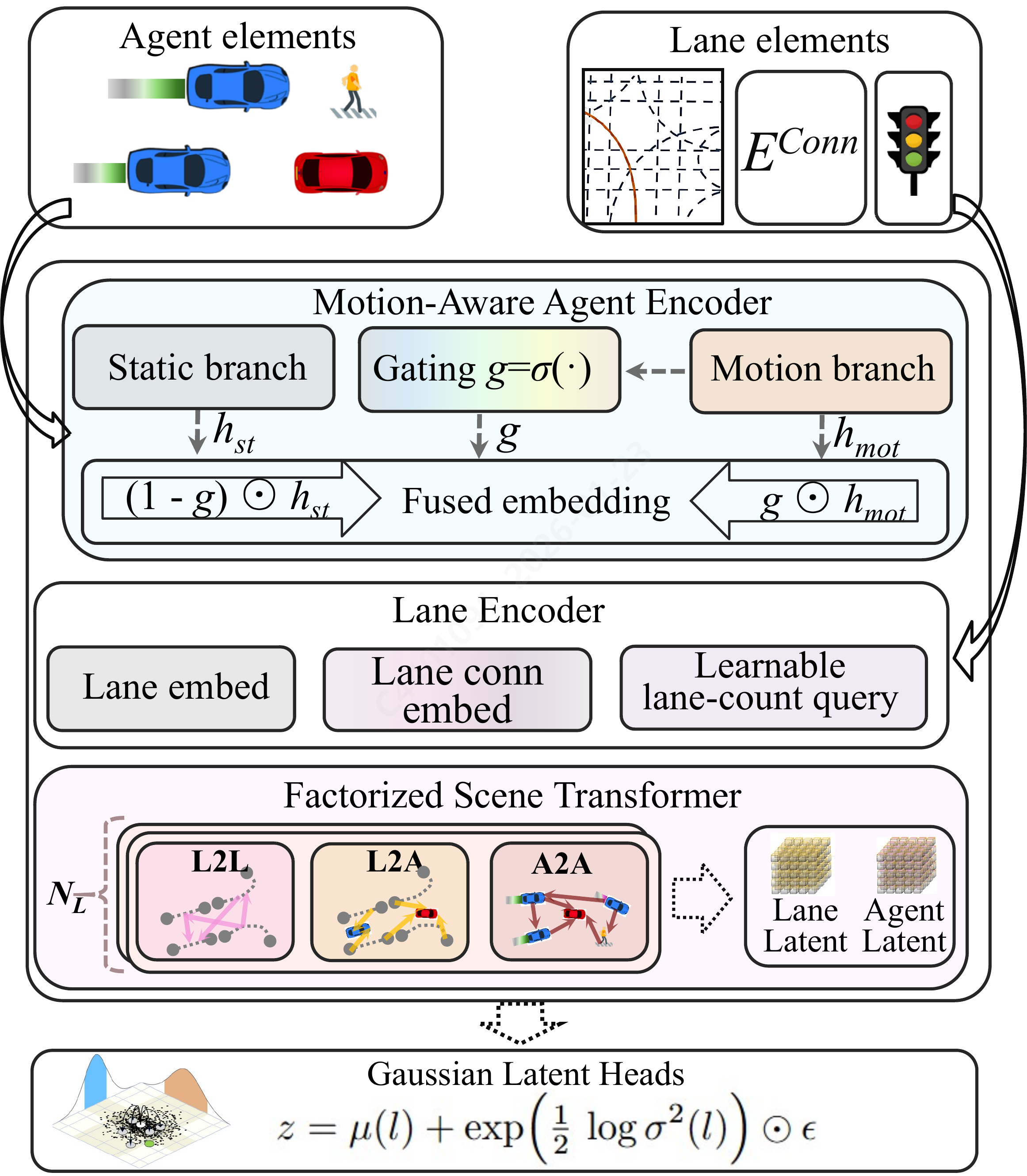}
  \caption{\textbf{Interaction-state interface with motion-aware gated VAE.} A motion-aware gate \(g_i\) fuses static state \(s_i\) and motion code \(m_i\), suppressing history noise for stationary agents while preserving dynamics for moving ones. A factorized scene transformer aggregates typed relations (L2L/L2A/A2A) and outputs Gaussian latents \(z=(z_{\mathrm{lane}},z_{\mathrm{agent}})\) for downstream generation. Warm-started interaction states align initialization with history-conditioned policies.}
  \label{fig:ae}
\end{figure}

\section{Method}
\label{sec:method}

\paragraph{Closed-Loop Streaming Formulation.}
We formulate streaming simulation as a sliding-window generation task over a dynamic vector graph $\mathcal{G}_k$. At simulation step $k$, the system maintains an ego-centered local tile and an \InteractionState\ for each agent. An ego planner $\pi_{\mathrm{ego}}$ and our reactive NPC policy $\pi_{\Delta}$ (Section~\ref{sec:deltasim}) act upon this state. To sustain kilometer-scale rollout without pre-generation, we trigger \textit{outpainting} when the ego approaches the tile boundary: the system encodes the current state via $\mathrm{Enc}$, and generates the frontier via a solver-free MeanFlow generator (Section~\ref{sec:meanflow}). This pipeline relies on three coupled innovations to ensure consistency and speed: (i) a \textbf{\MotionVAE} for history-aligned initialization (Section~\ref{sec:vae}); (ii) an \textbf{\EdgeDiT} for structural consistency (Section~\ref{sec:dit}); and (iii) \textbf{\DeltaSim} for kinematic feasibility (Section~\ref{sec:deltasim}).

\begin{figure*}[t]
  \centering
  \includegraphics[width=1.0\linewidth]{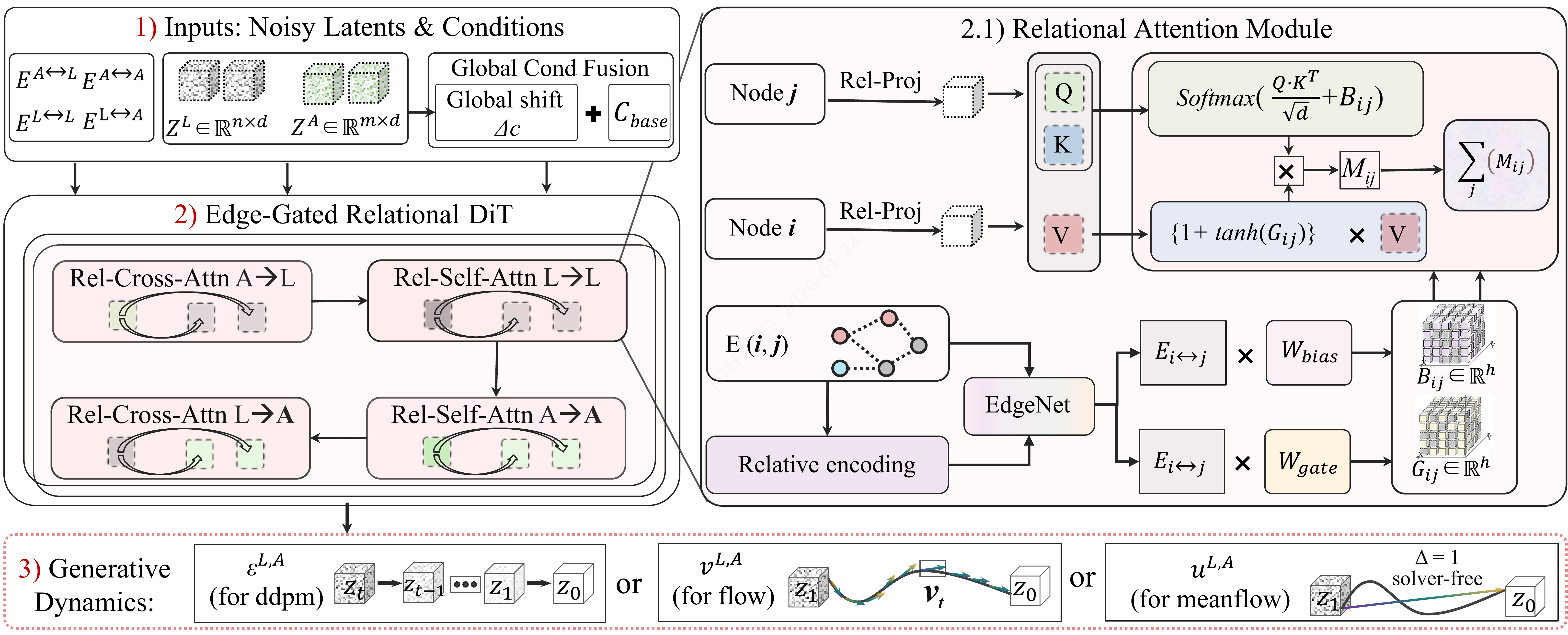}
  \caption{\textbf{Edge-gated relational DiT for vector-graph latent generation.} The framework generates vector-graph latents via a factorized transformer backbone. To capture complex graph dependencies, we introduce a Relational Attention Module (Right). Unlike standard attention, this module conditions message passing on edge features \(\edgeemb_{ij}\) via: 1) An additive bias $B_{ij}$ applied to attention scores to regulate connectivity; 2) A multiplicative gate $G_{ij}$ applied to value vectors to modulate feature aggregation. This design enables precise structural control across heterogeneous agents and map elements, compatible with various generative dynamics (DDPM, Flow, or MeanFlow).}
  \label{fig:generator}
\end{figure*}


\subsection{Vector-Graph Representation}
\label{sec:repr}

We represent each ego-centric scene tile as a heterogeneous vector graph
\(\mathcal{G}=(\mathcal{V},\mathcal{E})\) with lane nodes \(\mathcal{V}_{\mathrm{lane}}\) and agent nodes \(\mathcal{V}_{\mathrm{agent}}\).
Edges are typed by interaction semantics: \(\mathcal{E}=\mathcal{E}_{\mathrm{L2L}}\cup\mathcal{E}_{\mathrm{A2A}}\cup\mathcal{E}_{\mathrm{L2A}}\cup\mathcal{E}_{\mathrm{A2L}}\),
capturing lane connectivity (predecessor/successor/adjacent) and lane--agent relations (proximity and alignment).
Compared to rasterization \cite{chitta2024sledge}, vector graphs keep topology explicit and make missing regions well-posed: we can clamp observed tokens and complete only frontier tokens, which is the key operation in streaming outpainting.

We learn a latent interface for \(\mathcal{G}\) via an autoencoder.
Let \(z=(z_{\mathrm{lane}}, z_{\mathrm{agent}})\) denote the lane/agent latent tokens produced by the encoder.
Generation, inpainting, and outpainting are all implemented as masked completion in this latent space, enabling a single generator to serve both initialization (\(t=0\)) and rollout-time outpainting.

\subsection{Interaction-State Interface}
\label{sec:vae}
\paragraph{Motivation: History-Conditioned Policies.}
Closed-loop behavior policies (including many ego planners and NPC models) are history-conditioned,
because intent inference depends on recent velocity and curvature trends.
Consequently, a history-free snapshot at initialization induces an implicit near-zero history prior for downstream policies,
often resulting in large jerk and unstable early interactions.

\paragraph{Interaction State: Static + Motion Code.}
We therefore represent each agent \(i\) by an \InteractionState\ consisting of
a static state \(s_i\in\mathbb{R}^{7}\) in the unified format
\([x, y, \mathrm{speed}, \cos\theta, \sin\theta, \mathrm{length}, \mathrm{width}]\),
a type token \(\tau_i\), and a motion-history code \(m_i\in\mathbb{R}^{2K_{\mathrm{mot}}}\) represented as a \(K_{\mathrm{mot}}\)-point polyline in the agent frame.
At simulation start, \(m_i\) is deterministically unrolled into a short pseudo-history sequence to fill the policy context window (\Cref{app:motion}).

\paragraph{Selective Motion Encoding.}
Motion history exhibits sparsity in information content: for parked agents, aggressively reconstructing \(m_i\) injects noise into the latent interface.
We adopt the motion-aware gating mechanism in \Cref{fig:ae}, and empirically observe that gate activation correlates with motion magnitude (Appendix Figure~\ref{fig:app:gate}), resulting in an interpretable separation between static and dynamic agents:
\begin{equation}
\label{eq:vae_core}
\begin{gathered}
g_i = \sigma\!\Big(f_{\mathrm{gate}}\big([s_i, m_i, \tau_i]\big)\Big), \\
h_i = (1-g_i)\odot f_{\mathrm{st}}\big([s_i,\tau_i]\big)\;+\;g_i\odot f_{\mathrm{mot}}\big([m_i,\tau_i]\big), \\
q_\phi(z_i \mid s_i, m_i, \tau_i) = \mathcal{N}\!\Big(\mu_\phi(h_i),\ \mathrm{diag}\!\big(\sigma_\phi(h_i)^2\big)\Big),
\end{gathered}
\end{equation}
where \(g_i\in(0,1)^{d}\) is a learned gate, \(\sigma(\cdot)\) is the sigmoid, and \(\odot\) denotes element-wise product.

\paragraph{Scene-Level Relational Encoding.}
Following \Cref{fig:ae}, lane and agent tokens are processed by a factorized scene transformer that alternates
\(\mathrm{L2L}\), \(\mathrm{L2A}\), and \(\mathrm{A2A}\) interactions, and uses a learnable lane-count query for masked completion.
This design produces generation-friendly Gaussian latents \(z\) whose agent component is selectively history-aware,
enabling warm starts without introducing motion-code noise for static agents.
This VAE defines the latent interface \(z\) that the outpainting generator completes in \Cref{sec:dit}.
We provide the full \(\beta\)-VAE objective in Appendix~\ref{app:vae}.

\subsection{Edge-Gated Relational DiT}
\label{sec:dit}
\paragraph{Edge-Consistent Completion.}
Let \(z=(z_{\mathrm{lane}}, \allowbreak z_{\mathrm{agent}})\) be VAE latents.
We learn a generator that completes missing latent tokens given observed tokens and context \(c\) (counts, and optional scene labels). The challenge is that topology and cross-type alignment are edge-defined:
lane traversability depends on \(\mathcal{E}_{\mathrm{L2L}}\), while realistic interactions require agent--lane consistency.
Node-only attention can match marginal token statistics yet violate these constraints,
and such violations are amplified when masked completion is invoked repeatedly during streaming outpainting.

\paragraph{Factorized Typed Attention.}
To this end, we adopt a factorized DiT backbone (\Cref{fig:generator}) that alternates attention blocks over typed edge sets
\(\mathcal{E}_{\mathrm{L2L}},\mathcal{E}_{\mathrm{A2A}},\mathcal{E}_{\mathrm{L2A}},\mathcal{E}_{\mathrm{A2L}}\).
This explicitly matches heterogeneous interactions while keeping complexity compatible with real-time outpainting.

\begin{figure*}[t]
  \centering
  \includegraphics[width=1.0\linewidth]{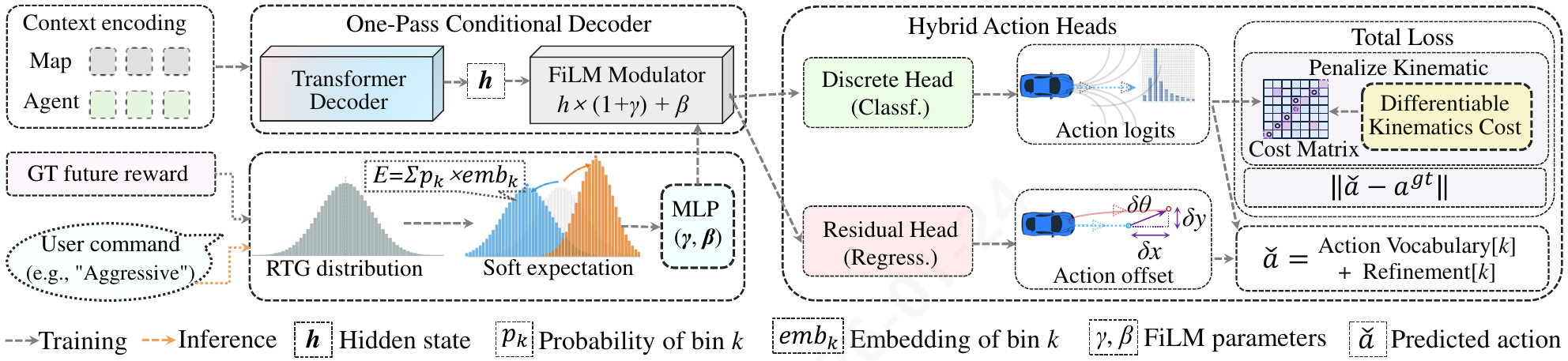}
  \caption{\textbf{\(\Delta\)Sim: physics-aligned NPC policy.} A single-pass return-to-go (RTG) control embedding modulates the decoder via FiLM \((\gamma,\beta)\). Actions use a hybrid head (discrete k-disks token plus continuous residual) with differentiable kinematic logit shaping and DKAL regularization. This design reduces feasibility violations that compound under kilometer-scale closed-loop simulation.}
  \label{fig:phygsim}
\end{figure*}

\paragraph{Edge-Conditioned Bias and Value Gating.}
For each typed edge \((i,j)\), we compute a relative edge embedding \(\edgeemb_{ij}\) and modulate attention through an additive bias and a multiplicative value gate:
\begin{equation}
\label{eq:rel_attn}
\begin{gathered}
\alpha_{ij}^{(h)} = \operatorname{softmax}_{j}\!\left(
\frac{\langle q_i^{(h)}, k_j^{(h)}\rangle}{\sqrt{d_h}}
\;+\; B^{(h)}\!\big(\edgeemb_{ij}\big)
\right), \\
y_i^{(h)} = \sum_{j}\alpha_{ij}^{(h)}
\Big(G^{(h)}\!\big(\edgeemb_{ij}\big)\odot v_j^{(h)}\Big),
\end{gathered}
\end{equation}
where \(h\) indexes attention heads, \(d_h\) is the head dimension,
\(B^{(h)}(\cdot)\) is an edge-conditioned logit bias, and \(G^{(h)}(\cdot)\) is an edge-conditioned gate
(initialized as identity to preserve optimization stability).

\paragraph{Mechanistic Effect.}
The bias term shapes \emph{connectivity preferences}, while the gate controls \emph{message strength} by suppressing edge-inconsistent feature aggregation.
This directly targets the main failure mode of streaming: cumulative structural drift under repeated masked completion, improving lane connectivity preservation and lane--agent alignment without rasterization or heuristic stitching.

\subsection{Interval-Conditioned MeanFlow}
\label{sec:meanflow}

\paragraph{Constraint: Real-time Large-Step Transport.}
Streaming outpainting calls the generator during rollout and thus requires real-time inference. Multi-step diffusion or ODE solvers are too costly to invoke repeatedly.
We therefore train the generator to support \emph{large-step} transport, including the solver-free one-step case that we deploy online (\Cref{fig:efficiency_quality}). We use \(k\) for discrete simulation steps and \(t\in[0,1]\) for latent-generation time; \(\Delta\) denotes the \MeanFlow\ interval length (full notation in Appendix~\ref{app:notation}).

\paragraph{JVP-based large-step calibration.}
Standard rectified flow matching supervises an instantaneous velocity field, which is accurate under small solver steps but does not directly constrain single large-step transport.
\MeanFlow\ instead predicts an \emph{interval-averaged} velocity \(u_\theta(z_t;\,t,\Delta,c)\).
To improve one-step accuracy, we supervise a first-order corrected predictor
\begin{equation}
\begin{gathered}
V_\theta(z_t;\,t,\Delta,c) = u_\theta(z_t;\,t,\Delta,c) + \\
\Delta\,\frac{\mathrm{d}}{\mathrm{d}t}u_\theta(z_t;\,t,\Delta,c),
\end{gathered}
\end{equation}
where \(\frac{\mathrm{d}}{\mathrm{d}t}u_\theta\) denotes the total derivative along the rectified path and is computed efficiently via a Jacobian--vector product (JVP) (Appendix~\ref{app:meanflow}). Here the total derivative differentiates through both the explicit time input and the implicit dependence on \(z_t\) along the rectified path.

\paragraph{Rectified Path with Masked Clamping.}
We use the straight-line probability path between data latents \(z\) and Gaussian noise \(\boldsymbol{\epsilon}\sim\mathcal{N}(\mathbf{0},\mathbf{I})\):
\begin{equation}
\label{eq:mf_path}
z_t = (1-t)\,z + t\,\boldsymbol{\epsilon},\qquad t\in[0,1].
\end{equation}
For masked (conditioned) tokens, we enforce a constant path by setting \(\boldsymbol{\epsilon}=z\),
which implies a zero target velocity and prevents conditioned structure from drifting during completion.
Along this path, the rectified-flow target velocity for unmasked tokens is \(v^\star = \boldsymbol{\epsilon} - z\).

\paragraph{Interval-Conditioned Mean Velocity \& JVP.}
Let \(u_\theta(z_t;\, \allowbreak t,\Delta,c)\) be the predicted \emph{mean velocity} over an interval of length \(\Delta=t-r\) (with \(r\le t\)),
conditioned on scene context \(c\).
We parameterize the model with two time channels \((t,\Delta)\) and train with the corrected target in \Cref{eq:mf_improved}.
\begin{equation}
\label{eq:mf_improved}
\begin{gathered}
\Delta = t-r, \\
V_\theta(z_t;\, t,r,c) = u_\theta(z_t;\, t,\Delta,c) \\
+ \Delta\, \frac{\mathrm{d}}{\mathrm{d}t}\,u_\theta(z_t;\, t,\Delta,c), \\
\mathcal{L}_{\mathrm{mf}} = \mathbb{E}\!\left[\left\lVert V_\theta(z_t;\, t,r,c)-v^\star\right\rVert_2^2\right],
\end{gathered}
\end{equation}
where \(\frac{\mathrm{d}}{\mathrm{d}t}u_\theta\) denotes the total derivative and is implemented efficiently as a JVP (Appendix~\ref{app:meanflow}).

\paragraph{Guided Solver-Free Update.}
For conditional generation, we apply classifier-free guidance (CFG) on the mean-velocity field:
\begin{equation}
\label{eq:mf_cfg}
\begin{gathered}
u_{\theta}^{\mathrm{cfg}} = u_\theta(z_t;\, t,\Delta,c_{\varnothing})
\;+\; \\
s\Big(u_\theta(z_t;\, t,\Delta,c)-u_\theta(z_t;\, t,\Delta,c_{\varnothing})\Big),
\end{gathered}
\end{equation}
where \(c_{\varnothing}\) is the unconditional condition implemented by scene-level label dropout, and \(s\) is the guidance scale.
Given \(u_{\theta}^{\mathrm{cfg}}\), we update latents by a large-step transport rule:
\begin{equation}
\label{eq:mf_update}
z_{t-\Delta} = z_t - \Delta\;u_{\theta}^{\mathrm{cfg}}(z_t;\, t,\Delta,c).
\end{equation}
Setting \((t,\Delta)=(1,1)\) generates solver-free one-step sampling; using a small number of intermediate times generates few-step sampling.
We report the resulting quality--latency trade-off in \Cref{fig:efficiency_quality}.

\paragraph{Why the JVP Term Matters for One-Step Completion?}
In our setting, a single large-step update must preserve edge-defined constraints (lane connectivity and lane--agent alignment) in one shot.
The JVP term calibrates the interval-velocity predictor at the boundary, which is where one-step failures concentrate.

\paragraph{Comparison with DDPM and Flow-Matching Baselines.}
DDPM trains a discrete-time denoiser by regressing injected noise and typically samples by iterating many reverse steps, while rectified flow matching trains a continuous-time velocity field on the straight-line path and samples via ODE integration. VectorWorld adopts the same latent interface and \EdgeDiT\ backbone across DDPM/Flow/MeanFlow, but uses MeanFlow for deployment because it is explicitly trained for large-step transport, enabling the solver-free update. A unified formulation and our masking/clamping implementation are detailed in Appendix~\ref{app:diffusion_flow} (DDPM/Flow) and Appendix~\ref{app:meanflow} (MeanFlow).

\subsection{Streaming Outpainting}
\label{sec:streaming}

\paragraph{Trigger, Clamp, Complete.}
VectorWorld maintains an ego-centered local tile and triggers outpainting only when the ego approaches the tile boundary.
Outpainting is implemented as masked latent completion:
tokens inside the current tile are clamped (conditioned), and only frontier tokens are sampled using \Cref{eq:mf_update}, decoded back to vector geometry, and stitched into the global scene; Algorithm~\ref{alg:streaming} (Appendix~\ref{app:streaming_rollout}) summarizes the full rollout loop.

\subsection{\(\Delta\)Sim: Physics-Aligned NPC Policy}
\label{sec:deltasim}

Even with high-fidelity scene generation, closed-loop stability depends on feasible multi-agent actions.
Small per-step kinematic violations are often negligible under short-horizon open-loop metrics, but dominate kilometer-scale rollouts via compounding effects. We propose \(\Delta\)Sim (\Cref{fig:phygsim}), combining controllable conditioning with physics-aligned action modeling and inference-time shaping.

\paragraph{One-Pass RTG Conditioning FiLM.}
\(\Delta\)Sim uses a one-pass conditional decoder whose hidden state \(h\) is modulated by FiLM \cite{perez2018film}
parameters \((\gamma,\beta)\) predicted from a soft return-to-go (RTG) embedding.
Specifically, the model predicts an RTG distribution \(\{p_j\}_{j=1}^{K_{\mathrm{rtg}}}\) and computes a soft expectation
\(E=\sum_{j=1}^{K_{\mathrm{rtg}}} p_j\,\mathrm{emb}_j\),
then applies FiLM as \(h \leftarrow h\odot(1+\gamma)+\beta\).
This provides a differentiable control interface without requiring a second forward at inference.

\begin{table*}[t]
\centering
\begin{threeparttable}
\caption{\textbf{Quantitative comparison of scene generation quality.}
We evaluate VectorWorld against prior vectorized scene generators on nuPlan and Waymo. VectorWorld achieves strong structure and safety with competitive perceptual alignment.
\(\downarrow\)/\(\uparrow\) indicates lower/higher is better.
\textbf{Bold} and \underline{underline} denote the best and second-best results among non-privileged methods.}
\label{tab:main_performance}
\renewcommand{\arraystretch}{0.9} 
\setlength{\tabcolsep}{1pt} 
\begin{tabular}{@{}llcccc@{}}
\toprule
\multirow{2.5}{*}{\textbf{Dataset}} & \multirow{2.5}{*}{\textbf{Method}} & \textbf{Perceptual} & \multicolumn{2}{c}{\textbf{Map Structure}} & \textbf{Safety} \\
\cmidrule(lr){3-3}\cmidrule(lr){4-5}\cmidrule(lr){6-6}
 & & FD $\downarrow$ & Route Len. (m) $\uparrow$ & Endpt. Dist. (m) $\downarrow$ & Coll. Rate (\%) $\downarrow$ \\
\midrule

\multirow{5}{*}{\textbf{nuPlan}} 
& SLEDGE-L~\cite{chitta2024sledge} & 1.89 & 35.34{\scriptsize $\pm$8.63} & 0.470{\scriptsize $\pm$0.320} & 22.30 \\
& SLEDGE-XL~\cite{chitta2024sledge} & 1.44 & 35.83{\scriptsize $\pm$8.35} & 0.420{\scriptsize $\pm$0.290} & 21.20 \\
& ScenDream-B~\cite{rowe2025scenario} & 1.05 & \secbest{37.04{\scriptsize $\pm$10.21}} & 0.320{\scriptsize $\pm$0.800} & 11.90 \\
& ScenDream-L~\cite{rowe2025scenario} & \best{0.67} & 36.87{\scriptsize $\pm$10.37} & \secbest{0.250{\scriptsize $\pm$0.710}} & \secbest{9.30} \\
& \cellcolor{oursrow}VectorWorld (Ours) & \cellcolor{oursrow}\secbest{0.98} & \cellcolor{oursrow}\best{37.22{\scriptsize $\pm$11.40}} & \cellcolor{oursrow}\best{0.078}{\scriptsize $\pm$0.106} & \cellcolor{oursrow}\best{3.01} \\
\midrule

\multirow{4}{*}{\textbf{Waymo}} 
& DriveSceneGen(GT)$^{*\dagger}$~\cite{sun2024drivescenegen} & 40.59$^{\dagger}$ & 41.61{\scriptsize $\pm$18.61} & 0.010{\scriptsize $\pm$0.000} & 0.20$^{*\dagger}$ \\
& ScenDream-B~\cite{rowe2025scenario} & 1.61 & 38.23{\scriptsize $\pm$12.78} & 0.320{\scriptsize $\pm$0.900} & 5.40 \\
& ScenDream-L~\cite{rowe2025scenario} & \secbest{1.38} & \secbest{38.92{\scriptsize $\pm$13.56}} & \secbest{0.210{\scriptsize $\pm$0.750}} & \secbest{4.80} \\
& \cellcolor{oursrow}VectorWorld (Ours) & \cellcolor{oursrow}\best{0.94} & \cellcolor{oursrow}\best{39.03{\scriptsize $\pm$15.43}} & \cellcolor{oursrow}\best{0.094}{\scriptsize $\pm$0.483} & \cellcolor{oursrow}\best{4.69} \\
\bottomrule
\end{tabular}
\begin{tablenotes}[flushleft]\footnotesize
\item $\dagger$ DriveSceneGen(GT) \cite{sun2024drivescenegen} numbers on Waymo are quoted from ScenarioDreamer~\cite{rowe2025scenario}.
\end{tablenotes}
\end{threeparttable}
\end{table*}

\begin{figure*}[t]
  \centering
  \includegraphics[width=0.98\linewidth]{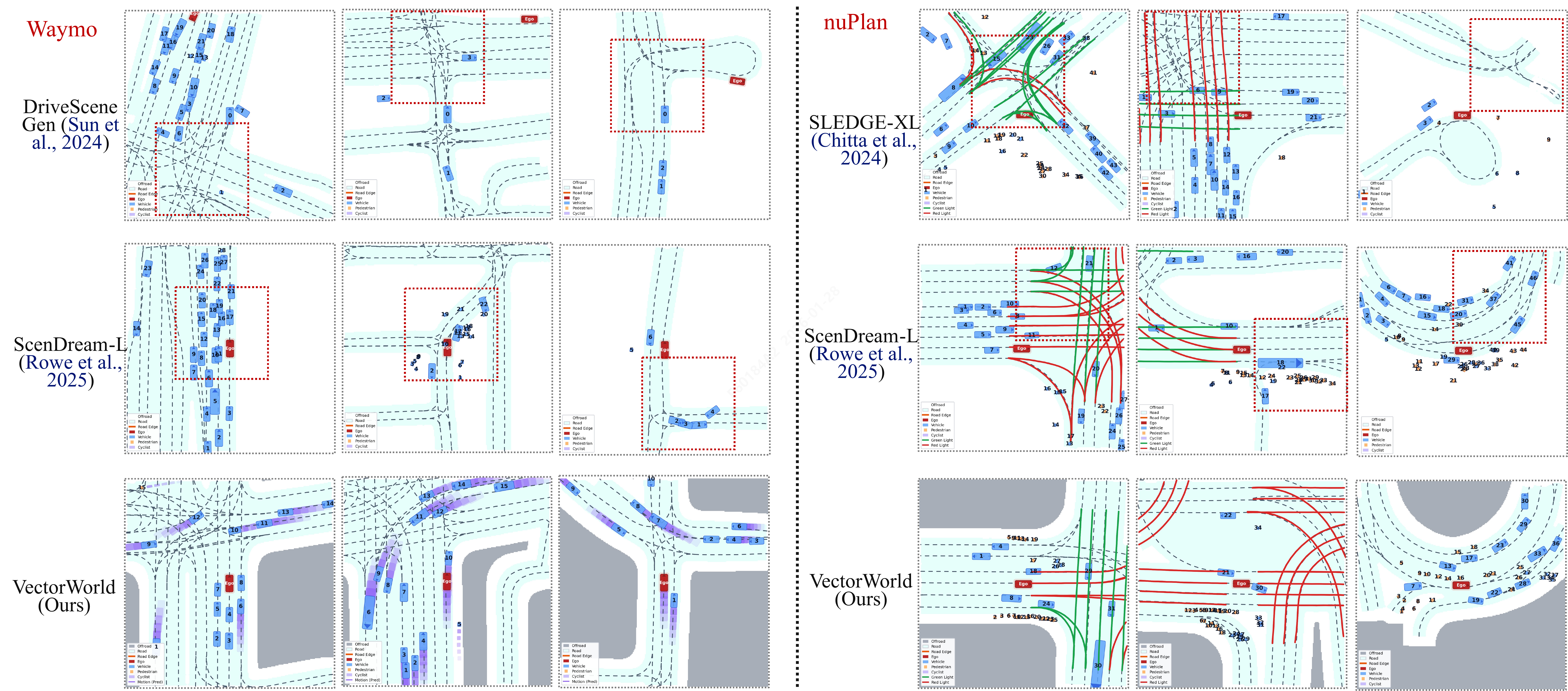}
  \caption{\textbf{Qualitative comparison on Waymo and nuPlan.} Visualizing decoded elements in \(64\,\mathrm{m}\times 64\,\mathrm{m}\) tiles. VectorWorld demonstrates superior lane connectivity and fewer agent overlaps at initialization compared to prior vectorized generators.}
  \label{fig:app:waymo_nuplan_compare}
\end{figure*}

\begin{figure*}[t]
  \centering
  \includegraphics[width=0.95\linewidth]{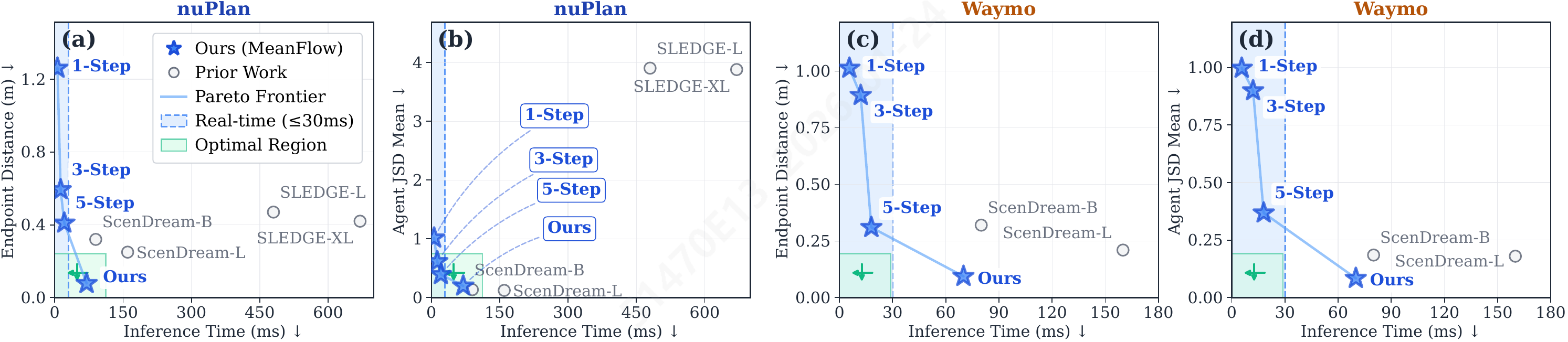}
  \caption{\textbf{Quality--latency trade-off for tile generation.} Endpoint distance (EPD) and mean agent-JSD are plotted against wall-clock latency as the solver-step budget varies; the vertical marker indicates the streaming budget (including one VAE decode).}
  \label{fig:efficiency_quality}
\end{figure*}

\paragraph{Hybrid Discrete--Continuous Actions.}
We parameterize the executed action in the agent body frame as a hybrid discrete--continuous update:
\begin{equation}
\label{eq:deltasim_action}
\Delta \xi
= \Delta \xi_{\hat{k}} + \alpha\,\widehat{\boldsymbol{\delta}},
\qquad
\hat{k} \sim \mathrm{Cat}\!\big(\mathrm{softmax}(\tilde{\ell})\big),
\end{equation}
where \(\widehat{\boldsymbol{\delta}} \in  \mathbb{R}^{3}\) is a continuous refinement head \((\Delta x,\Delta y,\Delta\theta)\), and \(\alpha\) is an inference-time refinement scale.

\paragraph{Inference-Time Kinematic Shaping.}
To suppress infeasible maneuvers, we compute a differentiable kinematic cost \(C_k(s_t)\) for each token \(k\) given the current state \(s_t\)
(e.g., yaw-rate, curvature, lateral/longitudinal acceleration, and reversibility constraints),
and shape logits before sampling:
\begin{equation}
\label{eq:deltasim_shaping}
\tilde{\ell}_k = \ell_k - \lambda\, C_k(s_t).
\end{equation}
This implements an energy-based prior over actions while preserving a single-pass architecture.

\paragraph{Differentiable Kinematic Alignment Loss (DKAL).}
Let \(\mathrm{ctr}(\mathbf{x}) \triangleq \mathbf{x} - \frac{1}{K_{\mathrm{kd}}}\sum_{k=1}^{K_{\mathrm{kd}}} x_k\) denote centering over the vocabulary dimension.
Given logits \(\ell\in\mathbb{R}^{K_{\mathrm{kd}}}\) and differentiable costs \(C(s_t)\in\mathbb{R}^{K_{\mathrm{kd}}}\), we minimize:
\begin{equation}
\label{eq:deltasim_dkal}
\mathcal{L}_{\mathrm{dkal}}
=
\big\lVert
\mathrm{ctr}(\ell)
+
\lambda_{\mathrm{dkal}}\, \mathrm{ctr}\!\big(C(s_t)\big)
\big\rVert_2^2,
\end{equation}
encouraging high-probability actions to be kinematically cheap (and vice versa), and reducing reliance on inference-only shaping.

\paragraph{Closing the Loop at \(k=0\).}
Finally, the \InteractionState\ closes the loop:
at \(k=0\), the VAE motion code \(m_i\) is deterministically converted to a pseudo-history sequence that fills the NPC context window.
This ensures that \(\Delta\)Sim receives inputs consistent with its training distribution, a prerequisite for stable rollouts when the scene is repeatedly extended by streaming outpainting.

\section{Experiments}
\label{sec:exp}

We evaluate \VectorWorld\ as (i) an offline scene generator, (ii) an online outpainting engine, and (iii) a closed-loop simulator.
The central question is whether a fast, one-step generator can remain structurally consistent and kinematically feasible when invoked repeatedly during rollout.

\begin{table*}[t]
  \centering
  \small
  \caption{\textbf{Closed-loop system evaluation on Waymo.} 
  \textbf{Units:} Scene Validity (FD, S.Coll) and Ego Performance (Coll, Offroad, Success) are in percentages (\%) unless unitless. Route is in meters. 
  \textbf{Columns:} \textit{NPC Policy} denotes the background traffic controller; \textit{Ego Planner} denotes the agent under test. 
  \textbf{Key:} Warm = Motion-history warm start; Online = Streaming outpainting.}
  \label{tab:closed_loop_eval}
  
  \setlength{\tabcolsep}{3.5pt} 
  \renewcommand{\arraystretch}{0.95}

  \begin{tabular}{@{} lcccc cc cccc c @{}}
    \toprule
    \multirow{2}{*}{\textbf{Simulator}} & 
    \multirow{2}{*}{\textbf{NPC Policy}} & 
    \multirow{2}{*}{\textbf{Ego Planner}} & 
    \multicolumn{2}{c}{\textbf{Sys. Cap.}} & 
    \multicolumn{2}{c}{\textbf{Scene Val.}} & 
    \multicolumn{4}{c}{\textbf{Ego Performance}} & 
    \textbf{Dyn. Metric} \\
    \cmidrule(lr){4-5} \cmidrule(lr){6-7} \cmidrule(lr){8-11} \cmidrule(lr){12-12}
    
    & & & Online & Warm 
    & FD $\downarrow$ & S.Coll $\downarrow$ 
    & Coll. & Off. & Succ. & Route (m) 
    & Value \\

    \midrule
    \multicolumn{12}{l}{\cellcolor{gray!10}\textit{\textbf{I. Platform Stability \& Consistency} (Goal: Verify simulation realism using a standard IDM driver \cite{treiber2000congested}.)}} \\
    \midrule
    
    Waymo & Replay & IDM (Rule-based) & \xmark & \xmark & 0.00 & -- & 7.2 & 5.8 & 87.0 & 64 & 11.4 \scriptsize{(Jerk)} \\
    ScenDream-L & Ctrl-sim & IDM (Rule-based) & \xmark & \xmark & 1.38 & 4.80 & 12.4 & 5.9 & 81.7 & 200 & 15.5 \scriptsize{(Jerk)} \\
    VectorWorld & $\Delta$Sim & IDM (Rule-based) & \xmark & \xmark & 1.05 & 4.69 & 7.5 & 1.5 & 91.0 & 200 & 10.4 \scriptsize{(Jerk)} \\
    VectorWorld & $\Delta$Sim & IDM (Rule-based) & \cmark & \xmark & 1.19 & 7.39 & \textbf{45.5} & 6.9 & \textbf{47.6} & 1000 & 16.6 \scriptsize{(Jerk)} \\
    \rowcolor{blue!5}
    VectorWorld & $\Delta$Sim & IDM (Rule-based) & \cmark & \cmark & \textbf{0.94} & \textbf{2.95} & 41.6 & \textbf{7.0} & 51.4 & 1000 & \textbf{9.6} \scriptsize{(Jerk)} \\

    \midrule
    \multicolumn{12}{l}{\cellcolor{gray!10}\textit{\textbf{II. Failure Discovery \& Stress Test} (Goal: Attack a fixed PPO policy \cite{schulman2017proximal} to find failures.)}} \\
    \midrule
    
    Waymo & Replay & PPO (Pre-trained) & \xmark & \xmark & 0.00 & -- & 29.3 & 6.9 & 63.8 & 200 & 36.2 \scriptsize{(Fail \%)} \\
    ScenDream-L  & Ctrl-sim (Pos.) & PPO (Pre-trained) & \xmark & \xmark & 1.42 & 5.10 & 52.8 & 9.1 & 38.2 & 200 & 61.8 \scriptsize{(Fail \%)} \\
    ScenDream-L  & Ctrl-sim (Neg.) & PPO (Pre-trained) & \xmark & \xmark & 1.42 & 11.7 & 59.0 & 9.0 & 32.1 & 200 & 67.9 \scriptsize{(Fail \%)} \\
    VectorWorld & $\Delta$Sim(Pos.) & PPO (Pre-trained) & \cmark & \cmark & \textbf{0.92} & \textbf{2.13} & 30.7 & 7.2 & 62.3 & 1000 & 37.9 \scriptsize{(Fail \%)} \\
    \rowcolor{orange!10}
    VectorWorld & $\Delta$Sim(Neg.) & PPO (Pre-trained) & \cmark & \cmark & 1.15 & 8.55 & \textbf{65.1} & \textbf{9.2} & \textbf{25.7} & 1000 & \textbf{75.0} \scriptsize{(Fail \%)} \\

    \midrule
    \multicolumn{12}{l}{\cellcolor{gray!10}\textit{\textbf{III. Closed-Loop Training} (Goal: Retrain PPO in simulator, evaluate on VectorWorld (Ours).)}} \\
    \midrule
    
    VectorWorld & $\Delta$Sim(Neg.) & PPO (Pre-trained) & \cmark & \cmark & 1.15 & 8.55 & 65.1 & 9.2 & 25.7 & 1000 & 25.7 \scriptsize{(Succ \%)} \\
    \rowcolor{green!10}
    VectorWorld & $\Delta$Sim(Neg.) & PPO (Retrained) & \cmark & \cmark & 1.15 & 8.55 & \textbf{35.1} & \textbf{7.9} & \textbf{56.0} & 1000 & \textbf{56.0} \scriptsize{(Succ \%)} \\
    
    \bottomrule
  \end{tabular}
\end{table*}

\subsection{Setup}
\label{sec:exp_setup}

\begin{table}[t]
  \centering
  \scriptsize 
  \setlength{\tabcolsep}{2pt} 
    \caption{\textbf{\DeltaSim\ ablation on Waymo.} Components: return-to-go conditioning (RTG), hybrid anchor+residual action head (A+R), and DKAL. Metrics: average displacement error (ADE; \(\downarrow\)), controllability rank-correlation (\(\rho\); \(\uparrow\)), and inference cost (latency / forward passes; \(\downarrow\)).}
  \label{tab:deltasim_main_results_single}
  \begin{tabular}{@{}l ccc c c cc@{}}
    \toprule
    \multirow{2.5}{*}{\textbf{Method}} & 
    \multicolumn{3}{c}{\textbf{Components}} & 
    \textbf{Acc.} & 
    \textbf{Ctrl.} & 
    \multicolumn{2}{c}{\textbf{Effic.}} \\
    
    \cmidrule(lr){2-4} \cmidrule(lr){5-5} \cmidrule(lr){6-6} \cmidrule(lr){7-8}
    
    & \textbf{RTG} 
    & \textbf{A+R} 
    & \textbf{DKAL} 
    & ADE $\downarrow$
    & $\rho$ $\uparrow$
    & Lat. $\downarrow$
    & Fwd $\downarrow$ \\
    
    \midrule
    
    Ctrl-Sim \cite{rowe2024ctrlsim} & \xmark & \xmark & \xmark & 2.80 & -0.25 & 36.9 & 2.0 \\
    
    \midrule
    
    \DeltaSim\ (w/o Refine) & \cmark & \xmark & \cmark & 2.62 & 0.39 & \best{21.7} & \best{1.0} \\
    \DeltaSim\ (w/o DKAL) & \cmark & \cmark & \xmark & 2.10 & 0.24 & 26.1 & \best{1.0} \\
    
    \rowcolor{oursrow}
    \textbf{\DeltaSim\ (Ours)} & \cmark & \cmark & \cmark & \best{1.72} & \best{0.53} & 26.1 & \best{1.0} \\
    
    \bottomrule
  \end{tabular}
\end{table}

\paragraph{Datasets and Baselines.}
We use Waymo Open Motion~\cite{gulino2023waymax} and nuPlan~\cite{caesar2021nuplan}.
We compare to vectorized scene generators SLEDGE~\cite{chitta2024sledge} (L/XL) and ScenDream~\cite{rowe2025scenario} (B/L).

\paragraph{Metrics.}
We follow the definitions in Appendix~\ref{app:metrics}.
Offline initialization quality is measured by FD, route length, EPD, and static collision rate (\Cref{tab:main_performance}).
We additionally report topology Fr\'echet distances and agent-attribute JSD as drift diagnostics (Appendix \Cref{tab:detailed_metrics}).
Efficiency is end-to-end latency per \(64\,\mathrm{m}\times 64\,\mathrm{m}\) tile (sampling + one decode), with a streaming budget of \(\le 30\)\,ms (\Cref{fig:efficiency_quality}).

\paragraph{Closed Loop.}
We evaluate an ego planner interacting with reactive NPCs and report collision/offroad/success, route progress, and jerk (Table~\ref{tab:closed_loop_eval}; definitions in Appendix~\ref{app:closedloop}).
We use three probes: a rule-based IDM probe~\cite{treiber2000congested} for platform stability, a fixed PPO policy~\cite{schulman2017proximal} for stress testing, and PPO retraining to measure training value.

\subsection{Results}
\label{sec:exp_results}

\paragraph{Offline Scene Quality.}
Table~\ref{tab:main_performance} shows that \VectorWorld\ improves structural integrity and initialization validity on both datasets.
On nuPlan, \VectorWorld\ reduces EPD from 0.250\,m (ScenDream-L) to 0.078\,m and has lower collision rate from 9.30\% to 3.01\%.
On Waymo, \VectorWorld\ achieves the lowest FD among non-privileged methods (0.94) while improving route length and EPD.


\paragraph{Qualitative Comparison.}
\Cref{fig:app:waymo_nuplan_compare} visualizes representative samples on both datasets.
The improvements concentrate at intersections and merges: \VectorWorld\ better preserves lane successor continuity and reduces agent--lane inconsistencies, matching the gains in endpoint distance and initialization collision rate in \Cref{tab:main_performance}.

\paragraph{Real-time Efficiency--Quality Trade-off.}
Figure~\ref{fig:efficiency_quality} reports the quality--latency curve as we vary the solver step budget.
MeanFlow-1 stays within the real-time regime and remains competitive in EPD and mean agent-JSD, enabling streaming outpainting; increasing steps improves quality but quickly exceeds the budget.

\paragraph{Ablation of \DeltaSim.}
Table~\ref{tab:deltasim_main_results_single} shows that RTG conditioning, hybrid anchor--residual actions, and DKAL are complementary.
Compared to Ctrl-Sim~\cite{rowe2024ctrlsim}, the full \DeltaSim\ improves ADE from 2.80\,m to 1.72\,m and increases controllability (\(\rho\)) from \(-0.25\) to 0.53, while using a single forward pass.

\paragraph{Closed-Loop System.}
Table~\ref{tab:closed_loop_eval} shows that online outpainting extends evaluation to \(1\,\mathrm{km}+\) but exposes compounding errors.
Warm-started interaction states reduce jerk (16.6 \(\to\) 9.6) and increase success (42.0\% \(\to\) 78.0\%), consistent with history-free initialization being a primary source of early-horizon instability.
Controllable ``tilt'' produces targeted stress-test distributions, and ego planner (PPO \cite{schulman2017proximal}) retraining in \VectorWorld\ increases success on the hardest setting from 25.0\% to 56.0\%. At evaluation, we control aggressiveness by exponentially tilting the RTG distribution with coefficient \(\kappa\) (Appendix~\ref{app:deltasim}).

\vspace{-2mm}
\paragraph{Additional Experiments, Limitations, and Safety.}
Further setups, ablations, qualitative analyses, and discussions on limitations and safety are provided in Appendix~\ref{sec:appendix}.

\section{Conclusion}
We introduced VectorWorld, a streaming vector-graph world model for closed-loop autonomous-driving simulation that outpaints lane--agent tiles on demand. It couples (i) a motion-aware interaction-state VAE that aligns initialization with history-conditioned policies,
(ii) an edge-gated relational DiT trained with interval-conditioned MeanFlow and JVP supervision for solver-free one-step masked completion,
and (iii) a physics-aligned NPC policy (\(\Delta\)Sim) that reduces long-horizon compounding from infeasible actions. Across Waymo Open Motion and nuPlan, VectorWorld improves initialization validity, enables \(1\,\mathrm{km}+\) closed-loop rollouts, and yields stress-test and policy-training gains.

\section*{Impact Statement}
This work aims to improve the fidelity and efficiency of learned simulation for autonomous-driving research.
A reliable closed-loop simulator can reduce real-world testing risk by enabling stress testing, counterfactual evaluation,
and data-efficient policy training in a controlled environment.

Potential negative impacts include misuse for generating adversarial traffic scenarios intended to exploit specific policy weaknesses,
or over-reliance on a simulator that contains modeling bias.
To mitigate these risks, we recommend (i) using generated scenarios only in sandboxed simulation,
(ii) reporting feasibility thresholds and post-processing rules used by the backend,
and (iii) validating policy improvements with cross-backend evaluation (e.g., log replay and alternative simulators),
as demonstrated in Appendix~A.14.6.

\bibliography{example_paper}

@inproceedings{rowe2025scenario,
  title={Scenario dreamer: Vectorized latent diffusion for generating driving simulation environments},
  author={Rowe, Luke and Girgis, Roger and Gosselin, Anthony and Paull, Liam and Pal, Christopher and Heide, Felix},
  booktitle={Proceedings of the Computer Vision and Pattern Recognition Conference},
  pages={17207--17218},
  year={2025}
}

@article{ho2020denoising,
  title={Denoising diffusion probabilistic models},
  author={Ho, Jonathan and Jain, Ajay and Abbeel, Pieter},
  journal={Advances in neural information processing systems},
  volume={33},
  pages={6840--6851},
  year={2020}
}

@article{ho2022classifier,
  title={Classifier-free diffusion guidance},
  author={Ho, Jonathan and Salimans, Tim},
  journal={arXiv preprint arXiv:2207.12598},
  year={2022}
}

@inproceedings{feng2023trafficgen,
  title={TrafficGen: Learning to Generate Diverse and Realistic Traffic Scenarios},
  author={Feng, Lan and Li, Quanyi and Peng, Zhenghao and Tan, Shuhan and Zhou, Bolei},
  booktitle={2023 IEEE International Conference on Robotics and Automation (ICRA)},
  pages={3567--3575},
  year={2023},
  organization={IEEE}
}

@article{tan2023language,
  title={Language conditioned traffic generation},
  author={Tan, Shuhan and Ivanovic, Boris and Weng, Xinshuo and Pavone, Marco and Kraehenbuehl, Philipp},
  journal={arXiv preprint arXiv:2307.07947},
  year={2023}
}

@inproceedings{yang2025wcdt,
  title={Wcdt: World-centric diffusion transformer for traffic scene generation},
  author={Yang, Chen and He, Yangfan and Tian, Aaron Xuxiang and Chen, Dong and Wang, Jianhui and Shi, Tianyu and Heydarian, Arsalan and Liu, Pei},
  booktitle={2025 IEEE International Conference on Robotics and Automation (ICRA)},
  pages={6566--6572},
  year={2025},
  organization={IEEE}
}

@article{ha2018world,
  title={World models},
  author={Ha, David and Schmidhuber, J{\"u}rgen},
  journal={arXiv preprint arXiv:1803.10122},
  volume={2},
  number={3},
  year={2018}
}

@inproceedings{yang2025long,
  title={Long-term traffic simulation with interleaved autoregressive motion and scenario generation},
  author={Yang, Xiuyu and Tan, Shuhan and Kr{\"a}henb{\"u}hl, Philipp},
  booktitle={Proceedings of the IEEE/CVF International Conference on Computer Vision},
  pages={25305--25314},
  year={2025}
}

@article{podell2023sdxl,
  title={Sdxl: Improving latent diffusion models for high-resolution image synthesis},
  author={Podell, Dustin and English, Zion and Lacey, Kyle and Blattmann, Andreas and Dockhorn, Tim and M{\"u}ller, Jonas and Penna, Joe and Rombach, Robin},
  journal={arXiv preprint arXiv:2307.01952},
  year={2023}
}

@inproceedings{peebles2023scalable,
  title={Scalable diffusion models with transformers},
  author={Peebles, William and Xie, Saining},
  booktitle={Proceedings of the IEEE/CVF international conference on computer vision},
  pages={4195--4205},
  year={2023}
}

@article{song2020denoising,
  title={Denoising diffusion implicit models},
  author={Song, Jiaming and Meng, Chenlin and Ermon, Stefano},
  journal={arXiv preprint arXiv:2010.02502},
  year={2020}
}

@article{lu2022dpm,
  title={Dpm-solver: A fast ode solver for diffusion probabilistic model sampling in around 10 steps},
  author={Lu, Cheng and Zhou, Yuhao and Bao, Fan and Chen, Jianfei and Li, Chongxuan and Zhu, Jun},
  journal={Advances in neural information processing systems},
  volume={35},
  pages={5775--5787},
  year={2022}
}

@article{karras2022elucidating,
  title={Elucidating the design space of diffusion-based generative models},
  author={Karras, Tero and Aittala, Miika and Aila, Timo and Laine, Samuli},
  journal={Advances in neural information processing systems},
  volume={35},
  pages={26565--26577},
  year={2022}
}

@inproceedings{song2023consistency,
title={Consistency Models},
author={Song, Yang and Dhariwal, Prafulla and Chen, Mark and Sutskever, Ilya},
booktitle={Proceedings of the 40th International Conference on Machine Learning},
pages={32211--32252},
year={2023},
volume={202},
series={Proceedings of Machine Learning Research},
publisher={PMLR}
}

@article{lipman2022flow,
  title={Flow matching for generative modeling},
  author={Lipman, Yaron and Chen, Ricky TQ and Ben-Hamu, Heli and Nickel, Maximilian and Le, Matt},
  journal={arXiv preprint arXiv:2210.02747},
  year={2022}
}

@article{geng2025mean,
  title={Mean flows for one-step generative modeling},
  author={Geng, Zhengyang and Deng, Mingyang and Bai, Xingjian and Kolter, J Zico and He, Kaiming},
  journal={arXiv preprint arXiv:2505.13447},
  year={2025}
}

@article{gao2024vista,
  title={Vista: A generalizable driving world model with high fidelity and versatile controllability},
  author={Gao, Shenyuan and Yang, Jiazhi and Chen, Li and Chitta, Kashyap and Qiu, Yihang and Geiger, Andreas and Zhang, Jun and Li, Hongyang},
  journal={Advances in Neural Information Processing Systems},
  volume={37},
  pages={91560--91596},
  year={2024}
}

@inproceedings{chen2025drivinggpt,
  title={Drivinggpt: Unifying driving world modeling and planning with multi-modal autoregressive transformers},
  author={Chen, Yuntao and Wang, Yuqi and Zhang, Zhaoxiang},
  booktitle={Proceedings of the IEEE/CVF International Conference on Computer Vision},
  pages={26890--26900},
  year={2025}
}

@inproceedings{ni2025maskgwm,
  title={Maskgwm: A generalizable driving world model with video mask reconstruction},
  author={Ni, Jingcheng and Guo, Yuxin and Liu, Yichen and Chen, Rui and Lu, Lewei and Wu, Zehuan},
  booktitle={Proceedings of the Computer Vision and Pattern Recognition Conference},
  pages={22381--22391},
  year={2025}
}

@inproceedings{chitta2024sledge,
  title={Sledge: Synthesizing driving environments with generative models and rule-based traffic},
  author={Chitta, Kashyap and Dauner, Daniel and Geiger, Andreas},
  booktitle={European Conference on Computer Vision},
  pages={57--74},
  year={2024},
  organization={Springer}
}

@article{sun2024drivescenegen,
  title={DriveSceneGen: Generating Diverse and Realistic Driving Scenarios From Scratch},
  author={Sun, Shuo and Gu, Zekai and Sun, Tianchen and Sun, Jiawei and Yuan, Chengran and Han, Yuhang and Li, Dongen and Ang, Marcelo H},
  journal={IEEE Robotics and Automation Letters},
  volume={9},
  number={8},
  pages={7007--7014},
  year={2024},
  publisher={IEEE}
}

@article{jiang2024scenediffuser,
  title={Scenediffuser: Efficient and controllable driving simulation initialization and rollout},
  author={Jiang, Max and Bai, Yijing and Cornman, Andre and Davis, Christopher and Huang, Xiukun and Jeon, Hong and Kulshrestha, Sakshum and Lambert, John and Li, Shuangyu and Zhou, Xuanyu and others},
  journal={Advances in Neural Information Processing Systems},
  volume={37},
  pages={55729--55760},
  year={2024}
}

@article{frans2024one,
  title={One step diffusion via shortcut models},
  author={Frans, Kevin and Hafner, Danijar and Levine, Sergey and Abbeel, Pieter},
  journal={arXiv preprint arXiv:2410.12557},
  year={2024}
}

@inproceedings{perez2018film,
  title={Film: Visual reasoning with a general conditioning layer},
  author={Perez, Ethan and Strub, Florian and De Vries, Harm and Dumoulin, Vincent and Courville, Aaron},
  booktitle={Proceedings of the AAAI conference on artificial intelligence},
  volume={32},
  year={2018}
}

@inproceedings{lin2025causal,
  title={Causal composition diffusion model for closed-loop traffic generation},
  author={Lin, Haohong and Huang, Xin and Phan, Tung and Hayden, David and Zhang, Huan and Zhao, Ding and Srinivasa, Siddhartha and Wolff, Eric and Chen, Hongge},
  booktitle={Proceedings of the Computer Vision and Pattern Recognition Conference},
  pages={27542--27552},
  year={2025}
}

@article{lee2024improving,
  title={Improving the training of rectified flows},
  author={Lee, Sangyun and Lin, Zinan and Fanti, Giulia},
  journal={Advances in neural information processing systems},
  volume={37},
  pages={63082--63109},
  year={2024}
}

@article{schulman2017proximal,
  title={Proximal policy optimization algorithms},
  author={Schulman, John and Wolski, Filip and Dhariwal, Prafulla and Radford, Alec and Klimov, Oleg},
  journal={arXiv preprint arXiv:1707.06347},
  year={2017}
}

@inproceedings{tan2025scenediffuserpp,
  title={SceneDiffuser++: City-Scale Traffic Simulation via a Generative World Model},
  author={Tan, Shuhan and Lambert, John and Jeon, Hong and Kulshrestha, Sakshum and Bai, Yijing and Luo, Jing and Anguelov, Dragomir and Tan, Mingxing and Jiang, Chiyu Max},
  booktitle={Proceedings of the Computer Vision and Pattern Recognition Conference},
  pages={1570--1580},
  year={2025}
}

@inproceedings{jiang2023vad,
  title={Vad: Vectorized scene representation for efficient autonomous driving},
  author={Jiang, Bo and Chen, Shaoyu and Xu, Qing and Liao, Bencheng and Chen, Jiajie and Zhou, Helong and Zhang, Qian and Liu, Wenyu and Huang, Chang and Wang, Xinggang},
  booktitle={Proceedings of the IEEE/CVF International Conference on Computer Vision},
  pages={8340--8350},
  year={2023}
}

@article{chen2024vadv2,
  title={Vadv2: End-to-end vectorized autonomous driving via probabilistic planning},
  author={Chen, Shaoyu and Jiang, Bo and Gao, Hao and Liao, Bencheng and Xu, Qing and Zhang, Qian and Huang, Chang and Liu, Wenyu and Wang, Xinggang},
  journal={arXiv preprint arXiv:2402.13243},
  year={2024}
}

@inproceedings{igl2022symphony,
  title        = {Symphony: Learning Realistic and Diverse Agents for Autonomous Driving Simulation},
  author       = {Igl, Maximilian and Kim, Daewoo and Kuefler, Alex and Mougin, Paul and Shah, Punit and Shiarlis, Kyriacos and Anguelov, Dragomir and Palatucci, Mark and White, Brandyn and Whiteson, Shimon},
  booktitle    = {2022 International Conference on Robotics and Automation (ICRA)},
  pages        = {2445--2451},
  year         = {2022},
  organization = {IEEE},
  doi          = {10.1109/ICRA46639.2022.9811990}
}

@inproceedings{suo2023mixsim,
  title     = {MixSim: A Hierarchical Framework for Mixed Reality Traffic Simulation},
  author    = {Suo, Simon and Wong, Kelvin and Xu, Justin and Tu, James and Cui, Alexander and Casas, Sergio and Urtasun, Raquel},
  booktitle = {Proceedings of the IEEE/CVF Conference on Computer Vision and Pattern Recognition (CVPR)},
  pages     = {9622--9631},
  year      = {2023}
}

@inproceedings{philion2024trajeglish,
  title     = {Trajeglish: Traffic Modeling as Next-Token Prediction},
  author    = {Philion, Jonah and Peng, Xue Bin and Fidler, Sanja},
  booktitle = {The Twelfth International Conference on Learning Representations},
  year      = {2024}
}

@inproceedings{rowe2024ctrlsim,
  title     = {{CtRL}-Sim: Reactive and Controllable Driving Agents with Offline Reinforcement Learning},
  author    = {Rowe, Luke and Girgis, Roger and Gosselin, Anthony and Carrez, Bruno and Golemo, Florian and Heide, Felix and Paull, Liam and Pal, Christopher},
  booktitle = {8th Annual Conference on Robot Learning},
  year      = {2024}
}

@inproceedings{dosovitskiy2017carla,
  title     = {{CARLA}: {An} Open Urban Driving Simulator},
  author    = {Dosovitskiy, Alexey and Ros, German and Codevilla, Felipe and Lopez, Antonio and Koltun, Vladlen},
  booktitle = {Proceedings of the 1st Annual Conference on Robot Learning},
  pages     = {1--16},
  year      = {2017},
  publisher = {PMLR}
}

@article{krajzewicz2012sumo,
  title   = {Recent Development and Applications of {SUMO} -- Simulation of Urban {MObility}},
  author  = {Krajzewicz, Daniel and Erdmann, Jakob and Behrisch, Michael and Bieker, Laura},
  journal = {International Journal On Advances in Systems and Measurements},
  volume  = {5},
  number  = {3--4},
  pages   = {128--138},
  year    = {2012},
  issn    = {1942-261X}
}

@article{li2021metadrive,
  title   = {MetaDrive: Composing Diverse Driving Scenarios for Generalizable Reinforcement Learning},
  author  = {Li, Quanyi and Peng, Zhenghao and Feng, Lan and Zhang, Qihang and Xue, Zhenghai and Zhou, Bolei},
  journal = {arXiv preprint arXiv:2109.12674},
  year    = {2021},
  doi     = {10.48550/arXiv.2109.12674}
}

@article{treiber2000congested,
  title={Congested traffic states in empirical observations and microscopic simulations},
  author={Treiber, Martin and Hennecke, Ansgar and Helbing, Dirk},
  journal={Physical review E},
  volume={62},
  number={2},
  pages={1805},
  year={2000},
  publisher={APS}
}

@article{caesar2021nuplan,
  title={nuplan: A closed-loop ml-based planning benchmark for autonomous vehicles},
  author={Caesar, Holger and Kabzan, Juraj and Tan, Kok Seang and Fong, Whye Kit and Wolff, Eric and Lang, Alex and Fletcher, Luke and Beijbom, Oscar and Omari, Sammy},
  journal={arXiv preprint arXiv:2106.11810},
  year={2021}
}

@article{gulino2023waymax,
  title={Waymax: An accelerated, data-driven simulator for large-scale autonomous driving research},
  author={Gulino, Cole and Fu, Justin and Luo, Wenjie and Tucker, George and Bronstein, Eli and Lu, Yiren and Harb, Jean and Pan, Xinlei and Wang, Yan and Chen, Xiangyu and others},
  journal={Advances in Neural Information Processing Systems},
  volume={36},
  pages={7730--7742},
  year={2023}
}
\bibliographystyle{icml2026}

\newpage
\clearpage
\appendix
\onecolumn
\section{Appendix}
\label{sec:appendix}

This appendix provides (i) a consolidated notation with tensor shapes, (ii) detailed specifications of the vectorized
scene representation and model components (motion-aware VAE, \EdgeDiT, MeanFlow training with JVP,
and \(\Delta\)Sim behavior model), and (iii) reproducible training and closed-loop evaluation protocols. We summarize notation in Tables~\ref{tab:app-notation-scene}--\ref{tab:app-notation-behavior} for quick lookup.

\makeatletter
\@ifpackageloaded{hyperref}{
  \newcommand{\apphref}[2]{\hyperref[#1]{#2}}
}{
  \newcommand{\apphref}[2]{#2~(\ref{#1})}
}
\makeatother

\subsection{Appendix table of contents}
\label{app:toc}

\begin{itemize}
  \setlength{\itemsep}{2pt}
  \setlength{\parskip}{0pt}

  \item Setup and notation
  \begin{itemize}
    \setlength{\itemsep}{1pt}
    \item \apphref{app:notation}{Notation and tensor shapes}
    \item \apphref{app:representation}{Vectorized scene representation}
    \item \apphref{app:motion}{Motion code: definition, normalization, and invariance}
  \end{itemize}

  \item Model components (interface, generator, behavior)
  \begin{itemize}
    \setlength{\itemsep}{1pt}
    \item \apphref{app:vae}{Motion-aware variational autoencoder}
    \item \apphref{app:dit}{\EdgeDiT\ backbone}
    \item \apphref{app:diffusion_flow}{Generative dynamics: DDPM and rectified flow matching}
    \item \apphref{app:meanflow}{MeanFlow training with identity+JVP}
    \item \apphref{app:deltasim}{\(\Delta\)Sim behavior model (return-conditioned transformer)}
  \end{itemize}

  \item Evaluation, metrics, and reproducibility
  \begin{itemize}
    \setlength{\itemsep}{1pt}
    \item \apphref{app:simgen}{Simulation environment generation and post-processing}
    \item \apphref{app:metrics}{Evaluation metrics (offline, behavior, and closed-loop)}
    \item \apphref{app:closedloop}{Closed-loop evaluation protocol}
    \item \apphref{app:streaming_rollout}{Streaming closed-loop rollout algorithm (on-demand outpainting)}
    \item \apphref{app:training}{Training setup and hyperparameters (Waymo)}
  \end{itemize}

  \item Additional experiments and analysis
  \begin{itemize}
    \setlength{\itemsep}{1pt}
    \item \apphref{app:extraexp}{Additional experiments and ablations}
    \begin{itemize}
      \setlength{\itemsep}{1pt}
      \item \apphref{app:extraexp:vae}{\InteractionState\ ablations (\MotionVAE)}
      \item \apphref{app:extraexp:dit}{Relation-aware generator ablations (\EdgeDiT)}
      \item \apphref{app:extraexp:meanflow}{MeanFlow objective and step-budget analysis}
      \item \apphref{app:extraexp:deltasim}{\(\Delta\)Sim feasibility--controllability trade-offs}
      \item \apphref{app:extraexp:closedloop}{Cross-backend validation and failure realism diagnostics}
    \end{itemize}
    \item \apphref{app:vis}{Qualitative results and visual analysis}
    \begin{itemize}
        \setlength{\itemsep}{1pt}
        \item \apphref{app:vis}{Overview and rendering protocol}
        \item \apphref{app:vis:waymo}{Waymo: initial-scene generation comparisons}
        \item \apphref{app:vis:nuplan}{nuPlan: traffic-light semantics and lane--agent consistency}
        \item \apphref{app:vis:control}{Map-conditioned controllability of agent generation}
        \item \apphref{app:vis:latency}{End-to-end latency comparisons (64m \(\times\) 64m scenes)}
        \item \apphref{app:vis:streaming}{Real-time streaming outpainting case studies (1km\(+\))}
    \end{itemize}
    \item \apphref{app:limitations}{Limitations and safety considerations}
  \end{itemize}
\end{itemize}

\subsection{Notation and Tensor Shapes}
\label{app:notation}

\begin{table*}[ht]
\centering
\small
\caption{Core notation for the vectorized scene representation (per scene unless stated otherwise).}
\label{tab:app-notation-scene}
\begin{tabular}{p{0.12\linewidth}p{0.22\linewidth}p{0.56\linewidth}}
\toprule
Symbol & Type / shape & Meaning \\
\midrule
\(\mathcal{G}=(\mathcal{V},\mathcal{E})\) & set-valued & Heterogeneous lane--agent vector graph for one ego-centric tile. \\
\(\mathcal{V}_{\mathrm{lane}},\mathcal{V}_{\mathrm{agent}}\) & sets & Lane nodes / agent nodes in \(\mathcal{G}\). \\
\(N_{\ell}\) & integer & Number of lane segments in a scene (capped by \(N_{\ell}^{\max}\)). \\
\(N_{a}\) & integer & Number of agents in a scene (capped by \(N_{a}^{\max}\); includes ego). \\
\(P\) & integer & Number of points per lane polyline for the generator/autoencoder (\(P{=}20\) in Waymo generation). \\
\(P_{\Delta}\) & integer & Number of points per lane polyline for \(\Delta\)Sim context (\(P_{\Delta}{=}50\) in Waymo behavior). \\
\(K_{\mathrm{mot}}\) & integer & Number of points in the motion-code polyline (per agent), i.e., \(m_i\in\mathbb{R}^{2K_{\mathrm{mot}}}\). \\
\(\mathbf{L}\) & \(\mathbb{R}^{N_{\ell}\times P \times 2}\) & Lane centerlines in the ego frame, resampled to \(P\) points, each point is \((x,y)\). \\
\(\mathbf{S}\) & \(\mathbb{R}^{N_{a}\times 7}\) & Static agent state: \((x,y,v,\cos\theta,\sin\theta,\ell,w)\), normalized. \\
\(\mathbf{M}\) & \(\mathbb{R}^{N_{a}\times 2K_{\mathrm{mot}}}\) & Motion code: \(K_{\mathrm{mot}}\) body-frame polyline points per agent, \((x_1,y_1,\dots,x_{K_{\mathrm{mot}}},y_{K_{\mathrm{mot}}})\). \\
\(\mathbf{c}^{a}\) & \(\{0,1\}^{N_{a}\times C_a}\) & Agent type one-hot (\(C_a{=}3\): vehicle, pedestrian, cyclist). \\
\(\mathbf{c}^{\ell}\) & \(\{0,1\}^{N_{\ell}\times C_{\ell}}\) & Lane semantic attributes (Waymo: dummy constant; nuPlan: may include traffic-light state when available). \\
\(\mathcal{A}_{\ell\ell}\) & subset of \([N_{\ell}]\times[N_{\ell}]\) & Directed lane-to-lane adjacency relation used for message passing. \\
\(\mathcal{A}_{aa}\) & subset of \([N_{a}]\times[N_{a}]\) & Agent-to-agent adjacency relation (complete in our implementation). \\
\(\mathcal{A}_{\ell a}\) & subset of \([N_{\ell}]\times[N_{a}]\) & Lane-to-agent adjacency relation (bipartite complete in our implementation). \\
\(\mathbf{T}_{\ell\ell}\) & \(\{0,1\}^{|\mathcal{A}_{\ell\ell}|\times C_{\text{conn}}}\) & One-hot lane connection type per directed edge (Waymo: \(C_{\text{conn}}{=}6\): none/pred/succ/left/right/self). \\
\(\mathbf{I}_{\ell\ell}\) & \(\mathbb{Z}^{2\times |\mathcal{A}_{\ell\ell}|}\) & Edge index (source, destination) for lane-to-lane edges (PyG convention). \\
\(\mathbf{I}_{aa}\) & \(\mathbb{Z}^{2\times |\mathcal{A}_{aa}|}\) & Edge index for agent-to-agent edges. \\
\(\mathbf{I}_{\ell a}\) & \(\mathbb{Z}^{2\times |\mathcal{A}_{\ell a}|}\) & Edge index for lane-to-agent edges. \\
\(\mathbf{m}^{\ell}\) & \(\{0,1\}^{N_{\ell}}\) & Partition mask: 1 indicates ``conditioned'' lanes (clamped during masked completion). \\
\(\mathbf{m}^{a}\) & \(\{0,1\}^{N_{a}}\) & Partition mask for agents in inpainting/outpainting. \\
\bottomrule
\end{tabular}
\end{table*}

We adopt a consistent notation that disambiguates (i) latent-generation time \(t\in[0,1]\) from simulation step indices
\(k\in\{0,\dots,K\}\), and (ii) attention \(\mathbf{V}\) (values) from k-disks vocabulary \(\mathbf{V}_{\mathrm{kd}}\).
We use bold \(\boldsymbol{\epsilon}\) to denote Gaussian noise and \(\mathcal{E}\) to denote graph edge sets; implementation uses adjacency relations and index tensors as listed above.

\begin{table*}[t]
\centering
\small
\caption{Notation for latent variables and generative dynamics (VAE + diffusion/flow/MeanFlow).}
\label{tab:app-notation-latents}
\begin{tabular}{p{0.12\linewidth}p{0.22\linewidth}p{0.56\linewidth}}
\toprule
Symbol & Type / shape & Meaning \\
\midrule
\(\Enc_{\phi}\) & function & VAE encoder mapping a scene graph to Gaussian posterior parameters. \\
\(\Dec_{\psi}\) & function & VAE decoder mapping sampled latents to reconstructed scene elements. \\
\(\mathbf{z}^{\ell}\) & \(\mathbb{R}^{N_{\ell}\times d_{\ell}}\) & Lane latents (Waymo: \(d_{\ell}{=}24\)). \\
\(\mathbf{z}^{a}\) & \(\mathbb{R}^{N_{a}\times d_{a}}\) & Agent latents (Waymo: \(d_{a}{=}18\)). \\
\(\mathbf{z}\) & \(\mathbb{R}^{(\cdot)}\) & Concatenated latents, \(\mathbf{z}=(\mathbf{z}^{\ell},\mathbf{z}^{a})\). \\
\(\mathbf{z}_t\) & \(\mathbb{R}^{(\cdot)}\) & Noisy latent on the rectified path at time \(t\): \(\mathbf{z}_t=(1-t)\mathbf{z}+t\boldsymbol{\epsilon}\). \\
\(\boldsymbol{\mu}^{\ell}, \log\boldsymbol{\sigma}^{2,\ell}\) & \(\mathbb{R}^{N_{\ell}\times d_{\ell}}\) & Lane posterior mean / log-variance from \(\Enc_{\phi}\). \\
\(\boldsymbol{\mu}^{a}, \log\boldsymbol{\sigma}^{2,a}\) & \(\mathbb{R}^{N_{a}\times d_{a}}\) & Agent posterior mean / log-variance from \(\Enc_{\phi}\). \\
\(\boldsymbol{\epsilon}\) & \(\mathbb{R}^{(\cdot)}\) & Standard Gaussian noise, \(\boldsymbol{\epsilon}\sim\mathcal{N}(\mathbf{0},\mathbf{I})\). \\
\(t\) & scalar in \([0,1]\) & Continuous time for flow / MeanFlow (not the simulation step). \\
\(r\) & scalar in \([0,1]\) & MeanFlow auxiliary time with constraint \(r\le t\). \\
\(\Delta\) & scalar in \([0,1]\) & MeanFlow interval length, \(\Delta=t-r\) (second time channel). \\
\(v^\star\) & \(\mathbb{R}^{(\cdot)}\) & Rectified-flow target velocity, \(v^\star=\boldsymbol{\epsilon}-\mathbf{z}\) (zero for clamped tokens). \\
\(c\) & (structured) & Generation context: masks, counts, and optional scene labels. \\
\(c_{\varnothing}\) & (structured) & Unconditional context used for CFG (implemented via label dropout). \\
\(u_{\theta}\) & function & MeanFlow predictor: mean velocity over interval \([r,t]\) at state \(\mathbf{z}_t\). \\
\(V_{\theta}\) & function & JVP-corrected MeanFlow predictor: \(V_{\theta}=u_{\theta}+\Delta\,\frac{\mathrm{d}}{\mathrm{d}t}u_{\theta}\). \\
\(s\) & scalar & Classifier-free guidance (CFG) scale used at sampling time. \\
\bottomrule
\end{tabular}
\end{table*}

\begin{table*}[t]
\centering
\small
\caption{Notation for the \(\Delta\)Sim behavior model and closed-loop simulation.}
\label{tab:app-notation-behavior}
\begin{tabular}{p{0.12\linewidth}p{0.26\linewidth}p{0.52\linewidth}}
\toprule
Symbol & Type / shape & Meaning \\
\midrule
\(\pi_{\mathrm{ego}}\) & policy & Ego planner/policy used in closed-loop rollout. \\
\(\pi_{\Delta}\) & policy & NPC policy implemented by \(\Delta\)Sim. \\
\(R\) & scalar & Tile radius (ego-centered observation crop) used by the simulator. \\
\(\tau\) & scalar & Outpainting trigger distance to the tile boundary (Algorithm~\ref{alg:streaming}). \\
\(\mathbf{V}_{\mathrm{kd}}\) & \(\mathbb{R}^{K_{\mathrm{kd}}\times 3}\) & k-disks vocabulary of SE(2) deltas, \((\Delta x,\Delta y,\Delta\theta)\), with \(K_{\mathrm{kd}}{=}384\). \\
\(a_k\) & integer & Discrete action token index at simulation step \(k\). \\
\(\ell\) & \(\mathbb{R}^{K_{\mathrm{kd}}}\) & Action logits over the k-disks vocabulary before shaping. \\
\(\tilde{\ell}\) & \(\mathbb{R}^{K_{\mathrm{kd}}}\) & Shaped logits, \(\tilde{\ell}_k=\ell_k-\lambda C_k(s_t)\). \\
\(C_k(s_t)\) & scalar & Differentiable kinematic cost of token \(k\) given current state \(s_t\). \\
\(\lambda\) & scalar & Inference-time shaping weight. \\
\(\lambda_{\mathrm{dkal}}\) & scalar & DKAL regularization weight used in training. \\
\(\alpha\) & scalar & Residual refinement scale in the hybrid action head. \\
\(\widehat{\boldsymbol{\delta}}_k\) & \(\mathbb{R}^{3}\) & Continuous residual refinement predicted by the residual head. \\
\(\widehat{\Delta\mathbf{s}}_k\) & \(\mathbb{R}^{3}\) & Final local delta action: \(\widehat{\Delta\mathbf{s}}_k=\mathbf{V}_{\mathrm{kd}}[a_k]+\widehat{\boldsymbol{\delta}}_k\). \\
\(K_{\mathrm{rtg}}\) & integer & Number of RTG bins (\(K_{\mathrm{rtg}}{=}350\) in Waymo). \\
\(\{p_j\}_{j=1}^{K_{\mathrm{rtg}}}\) & simplex & Predicted RTG distribution (used to form a soft control embedding). \\
\((\gamma,\beta)\) & vectors & FiLM modulation parameters applied to decoder hidden states. \\
\(\kappa\) & scalar & Exponential tilting coefficient applied at inference for controllability sweeps. \\
\(\Delta t\) & scalar & Simulation step size (Waymo closed-loop: \(\Delta t{=}0.1\) s). \\
\(K_{\mathrm{sim}}\) & integer & Number of simulation steps per episode (default \(K_{\mathrm{sim}}{=}400\)). \\
\bottomrule
\end{tabular}
\end{table*}


\subsection{Vectorized Scene Representation}
\label{app:representation}

We represent a traffic scene as a heterogeneous graph with lane and agent nodes. Each scene is normalized into an ego-centric
SE(2) frame by translating the ego to the origin and applying a deterministic rotation such that the ego heading aligns with a
fixed forward axis. We then crop to a square field-of-view (FOV) and cap the number of lanes and agents.

For Waymo generation, the autoencoder and latent generator operate on a \(64\text{ m}\times 64\text{ m}\) FOV, with at most
\(N_{\ell}^{\max}{=}100\) lane segments and \(N_{a}^{\max}{=}30\) agents. Lane segments are represented by \(P{=}20\) resampled points.

Lane connectivity is encoded as a typed directed relation. Concretely, we construct a complete directed lane-to-lane edge set
and assign each edge a connection-type one-hot \(\mathbf{T}_{\ell\ell}\in\{0,1\}^{|\mathcal{A}_{\ell\ell}|\times 6}\) indicating
\(\{\text{none},\text{pred},\text{succ},\text{left},\text{right},\text{self}\}\). This separates (i) the *message passing topology* (complete)
from (ii) the *semantic lane-graph structure* (typed edges), which is critical for topology metrics.

For outpainting/inpainting, we additionally construct a partitioned scene variant by splitting lane segments that cross a fixed
partition line (Waymo: \(y=0\) in the ego frame). Elements on the ``before-partition'' side are treated as conditions. We use
boolean masks to (i) prevent encoder attention across the partition and (ii) keep conditioned latents fixed during generation.

\paragraph{Mask convention.}
We use boolean masks where \texttt{True} indicates \emph{conditioned / clamped} tokens that must be preserved, and \texttt{False} indicates tokens to be generated.
All diffusion/flow/MeanFlow variants implement conditional completion by keeping clamped tokens fixed and setting their training targets to zero velocity/noise.

\subsection{Motion Code: Definition, Normalization, and Invariance}
\label{app:motion}

We augment each agent with a compact motion code that summarizes its recent history as a short polyline in the agent body frame.
For each agent \(i\), we define a static state \(s_i\in\mathbb{R}^{7}\), an agent type \(\tau_i\), and a motion-history code \(m_i\in\mathbb{R}^{2K_{\mathrm{mot}}}\) represented as a \(K_{\mathrm{mot}}\)-point polyline in the agent frame.

Given the global trajectory up to the selected timestamp, we transform historical positions into the agent's body frame at the
current time by applying the inverse SE(2) transform using the agent's current pose. We then truncate the history to the most
recent \(L_{\max}\) meters of arc length (Waymo default \(L_{\max}{=}12\) m) and resample \(K\) points uniformly in arc length.

To explicitly capture the discrete ``static'' mode, we set the motion code to all zeros if the agent is deemed static.
An agent is classified as static if either (i) the maximum displacement over the most recent history window is below
a threshold, or (ii) the mean speed is below a threshold (see Table~\ref{tab:training-hparams} for values).
This produces a clear static cluster in latent space and stabilizes downstream generation.

We normalize the motion polyline into \([-1,1]\) using fixed physical ranges in the body frame. For Waymo, longitudinal coordinates
are mapped from \([-D,0]\) and lateral coordinates from \([-Y,Y]\), where \(D{=}12\) m and \(Y{=}6\) m by default.

Importantly, the motion code is defined in the agent body frame and is therefore invariant to global SE(2) transforms used in
tile stitching. When extending a simulation environment by SE(2)-transforming a new tile into a global frame, we transform
agent *static states* and lane polylines, but keep the motion code unchanged.


\subsection{Motion-aware Variational Autoencoder}
\label{app:vae}

\begin{figure}[t]
  \centering
  \includegraphics[width=0.7\linewidth]{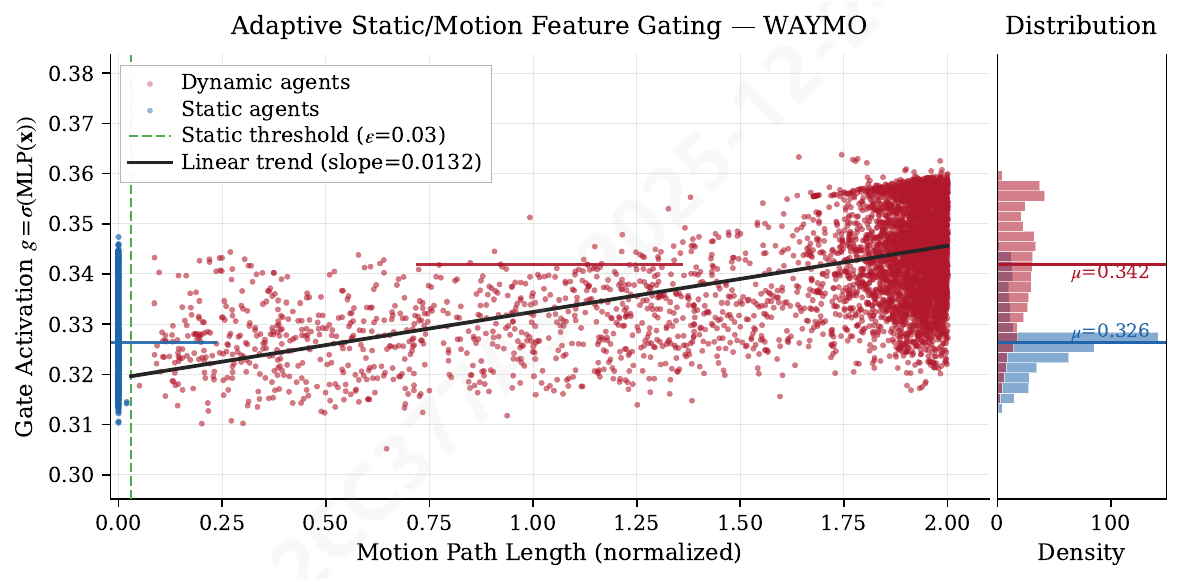}
  \caption{\textbf{Visualization of motion-aware gating (Waymo).}
  Gate activation increases with motion magnitude, clustering static agents at low activation while assigning higher activation to dynamic agents.
  This supports that the \MotionVAE\ suppresses motion-code noise for parked agents while encoding history for moving agents, improving warm-start consistency.}
  \label{fig:app:gate}
\end{figure}
Our scene autoencoder is a motion-aware VAE that encodes lanes, agents, and lane connectivity into Gaussian latent variables,
and decodes them back to vectorized geometry and discrete attributes.

\subsubsection{Agent Encoding with Motion-aware Gating}

Each agent is embedded using two branches: a static branch operating on the 7D instantaneous state and a motion branch operating
on the \(2K\)-dimensional motion code. A learnable gating network predicts a per-channel gate vector \(g\in(0,1)^{d}\) from the
concatenated agent input, and fuses the two representations as
\[
\mathbf{h}_{a} = (1-g)\odot \mathbf{h}_{\text{static}} + g\odot \mathbf{h}_{\text{motion}}.
\]
This design allows the encoder to rely on motion features when informative, while remaining robust when motion codes are zeroed.

\subsubsection{Factorized Scene Transformer}

We adopt a factorized attention architecture that alternates lane-to-lane (with typed edge features), lane-to-agent, and agent-to-agent
attention, with an explicit lane-edge update in each block. This preserves permutation equivariance within each node set while enabling
typed relational reasoning for lane topology.

\subsubsection{Lane-count Distribution Head for Inpainting}

For partitioned scenes, we use a learnable query token that attends to conditioned lanes to predict a distribution over the number
of lanes to generate in the outpainted region. The head outputs a categorical distribution over \(\{0,\dots,N_{\ell}^{\max}\}\), and is trained
with cross-entropy on the ground-truth lane count in the after-partition region.

\subsubsection{Training Objective}
Let \(\widehat{\mathbf{S}}, \widehat{\mathbf{M}}, \widehat{\mathbf{L}}\) denote decoded static states, motion codes, and lane polylines.
The total autoencoder loss is
\begin{equation}
\begin{split}
    \mathcal{L}_{\mathrm{AE}} =\: & \mathcal{L}_{\mathrm{static}}
    + \lambda_{\mathrm{mot}}\mathcal{L}_{\mathrm{motion}}
    + \lambda_{\ell}\mathcal{L}_{\mathrm{lane}}
    + \lambda_{\mathrm{conn}}\mathcal{L}_{\mathrm{conn}}
    + \lambda_{\mathrm{type}}\mathcal{L}_{\mathrm{type}} \\
    & + \beta_{\mathrm{KL}}\mathcal{L}_{\mathrm{KL}}
    + \lambda_{\mathrm{cnt}}\mathcal{L}_{\mathrm{count}}
    + \lambda_{\mathrm{smooth}}\mathcal{L}_{\mathrm{smooth}}
    + \lambda_{\mathrm{col}}\mathcal{L}_{\mathrm{collision}}
    + \lambda_{\mathrm{end}}\mathcal{L}_{\mathrm{endpoint}}.
\end{split}
\end{equation}
Here, \(\mathcal{L}_{\mathrm{static}}\) is a weighted \(L_1\) on \((x,y)\) versus other static dimensions; \(\mathcal{L}_{\mathrm{motion}}\) is an \(L_1\)
on motion codes with extra weight on static agents; \(\mathcal{L}_{\mathrm{lane}}\) is an \(L_1\) over lane points; \(\mathcal{L}_{\mathrm{conn}}\) and
\(\mathcal{L}_{\mathrm{type}}\) are cross-entropies for connectivity and types; and \(\mathcal{L}_{\mathrm{KL}}\) is the standard VAE KL term.


\subsection{Edge-gated relational DiT Backbone}
\label{app:dit}

Our latent generator uses a factorized DiT backbone that processes lane and agent latent tokens with alternating cross- and self-attention.
We incorporate optional relation-aware mechanisms that target distributional metrics sensitive to relational structure (topology and agent-lane alignment).

\subsubsection{Factorized DiT Blocks}

Each factorized block applies, in order: agent-to-lane cross attention, lane-to-lane self attention (repeated multiple times),
lane-to-agent cross attention, and agent-to-agent self attention. We allocate additional capacity to lane modeling by using multiple
lane-to-lane blocks per factorized block.

\subsubsection{Relational Logit Bias and Edge-gated Values}

Given token features \(\mathbf{x}\), we optionally compute a per-edge per-head additive logit bias \(b_{ij}^{(h)}\) by
(1) projecting node features to a low-dimensional space, (2) forming pairwise edge features, and (3) mapping to head-wise biases.
In addition, we optionally apply a per-edge per-head value gate \(g_{ij}^{(h)} = 1+\tanh(\cdot)\), which multiplicatively scales
messages and is zero-initialized to behave as identity at the start of training.

\subsubsection{Cross-type Relational Bias}

We further support cross-type relational bias on lane-to-agent and agent-to-lane cross attention. This explicitly models agent--lane
relationships and targets metrics that depend on relative geometry between agents and their nearest lanes.

\subsubsection{Global Context Fusion}

Optionally, we compute a scene-level context vector by mean pooling lane and agent token features, fusing them via an MLP, and injecting
the resulting delta into the conditioning stream. This provides a lightweight global pathway to capture scene-wide statistics.

\subsection{Generative Dynamics: DDPM and Rectified Flow Matching}
\label{app:diffusion_flow}

This section summarizes the two standard objectives we build on---DDPM latent diffusion and rectified flow matching---and clarifies how conditional masked completion is implemented by clamping.
MeanFlow (\Cref{app:meanflow}) can then be read as an interval-conditioned extension designed for large-step (including one-step) transport under the same latent interface and backbone.

\subsubsection{DDPM latent diffusion (noise-prediction objective)}
Let \(\mathbf{z}_0\) denote a latent sample (concatenating lane and agent tokens) and let \(n\in\{0,\dots,N\}\) be the discrete diffusion step.
DDPM defines a forward noising process
\begin{equation}
\label{eq:app:ddpm_forward}
\mathbf{z}_n = \sqrt{\bar{\alpha}_n}\,\mathbf{z}_0 + \sqrt{1-\bar{\alpha}_n}\,\boldsymbol{\epsilon},
\qquad
\boldsymbol{\epsilon}\sim\mathcal{N}(\mathbf{0},\mathbf{I}),
\end{equation}
and trains a neural denoiser \(\epsilon_{\theta}(\mathbf{z}_n,n,c)\) to regress the injected noise:
\begin{equation}
\label{eq:app:ddpm_loss}
\mathcal{L}_{\mathrm{ddpm}}
=
\mathbb{E}\Big[\big\lVert \epsilon_{\theta}(\mathbf{z}_n,n,c)-\boldsymbol{\epsilon}\big\rVert_2^2\Big].
\end{equation}
At test time, sampling requires iterating the reverse Markov chain for many steps, which is the primary latency bottleneck for streaming outpainting.

\paragraph{Masked completion via clamping.}
For conditioned (clamped) tokens, we enforce a constant trajectory during noising by setting \(\mathbf{z}_n=\mathbf{z}_0\) for those tokens.
Equivalently, the training target is \(\boldsymbol{\epsilon}=\mathbf{0}\) on clamped tokens, so the model is explicitly trained not to move conditioned structure.

\subsubsection{Rectified flow matching (velocity-prediction objective)}
Rectified flow matching operates in continuous time \(t\in[0,1]\) on the straight-line path between data latents and Gaussian noise:
\begin{equation}
\label{eq:app:rf_path}
\mathbf{z}_t = (1-t)\mathbf{z}_0 + t\boldsymbol{\epsilon}.
\end{equation}
The target velocity field is constant along the path, \(v^{\star}=\boldsymbol{\epsilon}-\mathbf{z}_0\), and the flow-matching loss is
\begin{equation}
\label{eq:app:rf_loss}
\mathcal{L}_{\mathrm{fm}}
=
\mathbb{E}\Big[\big\lVert v_{\theta}(\mathbf{z}_t,t,c)-(\boldsymbol{\epsilon}-\mathbf{z}_0)\big\rVert_2^2\Big].
\end{equation}
Sampling solves the ODE \(\mathrm{d}\mathbf{z}_t/\mathrm{d}t = v_{\theta}(\mathbf{z}_t,t,c)\) from \(t=1\) to \(t=0\) using an explicit solver (e.g., Euler/Heun), which still requires multiple function evaluations for high fidelity.

\paragraph{Masked completion via constant-path conditioning.}
For clamped tokens, we set \(\boldsymbol{\epsilon}=\mathbf{z}_0\) in \Cref{eq:app:rf_path}, yielding \(v^{\star}=\mathbf{0}\).
This matches the main-paper masking rule in \Cref{sec:meanflow} and prevents conditioned structure from drifting during inpainting/outpainting.

\subsubsection{Why large-step (one-step) sampling is non-trivial}
Both DDPM and flow matching are typically trained to be accurate under small reverse steps: DDPM through per-step denoising and flow matching through local ODE integration.
A single-step update is therefore an extrapolation regime where small local errors become catastrophic, especially for edge-defined constraints (lane successors and agent--lane relations) that must be satisfied \emph{in one shot}.
MeanFlow addresses this by conditioning the predictor on the interval length \(\Delta\) and explicitly supervising large-step transport with the identity+JVP objective (\Cref{app:meanflow}), which is exactly the regime required by streaming outpainting.

\subsection{MeanFlow Training with Identity+JVP}
\label{app:meanflow}

We use a meanflow latent generator with an identity+JVP training objective that targets high-quality one-step sampling. The key idea is to
learn an average velocity field over an interval \([r,t]\) and correct it using a Jacobian--vector product (JVP) term.

\subsubsection{Training Path and Ground-truth Velocity}

Let \(x\) denote a normalized latent sample (lane and agent latents). We sample \(\boldsymbol{\epsilon}\sim\mathcal{N}(\mathbf{0},\mathbf{I})\)
and define the rectified path
\[
\mathbf{z}_t = (1-t)\,x + t\,\boldsymbol{\epsilon},
\qquad
\mathbf{v}_{\mathrm{gt}} = \boldsymbol{\epsilon} - x.
\]
Conditioned (fixed) nodes use a constant path, yielding \(\mathbf{v}_{\mathrm{gt}}=\mathbf{0}\) for those nodes.

\subsubsection{Identity+JVP Objective}

Let \(u_{\theta}(\mathbf{z}_t;r,t)\) denote the predicted average velocity over \([r,t]\). We define the boundary velocity
\(v_{\theta}(\mathbf{z}_t;t)=u_{\theta}(\mathbf{z}_t;t,t)\). We then form the composite predictor
\[
V_{\theta}(\mathbf{z}_t;r,t)
=
u_{\theta}(\mathbf{z}_t;r,t)
+
(t-r)\,\frac{\partial}{\partial t}u_{\theta}(\mathbf{z}_t;r,t),
\]
where \(\frac{\partial}{\partial t}u_{\theta}\) is computed using a JVP, and the JVP output is stop-gradient for stability.
We minimize
\[
\mathcal{L}_{\mathrm{MF}}
=
\mathbb{E}\left[\left\lVert V_{\theta}(\mathbf{z}_t;r,t) - \mathbf{v}_{\mathrm{gt}} \right\rVert_2^2\right],
\]
with optional scene-level adaptive reweighting \(w=(\mathcal{L}_{\mathrm{MF}}+c)^{-p}\).

\paragraph{Sanity Checks.}
When \(\Delta=0\), the correction term vanishes and the objective reduces to the flow-matching limit. When the learned velocity field is locally constant along the trajectory, \(\frac{\mathrm{d}}{\mathrm{d}t}u_{\theta}\approx 0\) and \(V_{\theta}\approx u_{\theta}\), so the JVP term does not bias training in the easy regime but mainly improves calibration for large-step transport.

\subsubsection{Time Sampling}

We sample \(t\) and \(r\) from a logit-normal distribution, enforce \(r\le t\), and set a fraction of samples to \(r=t\) to retain
a flow-matching limit. We also explicitly include a subset of samples with \((t=1,r=0)\) to cover the strict one-step regime.

\subsubsection{Sampling (One-step and Few-step)}

At inference, we sample from \(\mathbf{z}_{t=1}\sim\mathcal{N}(\mathbf{0},\mathbf{I})\) and integrate from \(t=1\) to \(t=0\) using
\(K\) steps. For step \(k\), with \(t_{\text{cur}}>t_{\text{next}}\), we update
\[
\mathbf{z}_{t_{\text{next}}}
=
\mathbf{z}_{t_{\text{cur}}}
-
(t_{\text{cur}}-t_{\text{next}})\,u_{\theta}(\mathbf{z}_{t_{\text{cur}}}; r=t_{\text{next}}, t=t_{\text{cur}}),
\]
with classifier-free guidance applied by combining conditional and unconditional predictions.

\subsubsection{Algorithmic}
\begin{algorithm}[t]
\caption{MeanFlow identity+JVP training step (per minibatch).}
\label{alg:meanflow-jvp}
\begin{algorithmic}[1]
\STATE Sample data latents \(x\), sample \(\boldsymbol{\epsilon}\sim\mathcal{N}(\mathbf{0},\mathbf{I})\).
\STATE Sample times \(t\in[0,1]\), \(r\in[0,1]\) with \(r\le t\); set \(h \leftarrow t-r\).
\STATE Form \(\mathbf{z}_t \leftarrow (1-t)x + t\boldsymbol{\epsilon}\), and \(\mathbf{v}_{\mathrm{gt}}\leftarrow \boldsymbol{\epsilon}-x\).
\STATE Compute \(\Delta \leftarrow t-r\), \(u_{\theta}(\mathbf{z}_t;\,t,\Delta)\), and boundary velocity \(v_{\theta}(\mathbf{z}_t;\,t)\triangleq u_{\theta}(\mathbf{z}_t;\,t,\Delta{=}0)\).
\STATE Compute the JVP for the total derivative \(\frac{\mathrm{d}}{\mathrm{d}t}u_{\theta}(\mathbf{z}_t;\,t,\Delta)\) with respect to the composite input \((\mathbf{z}_t,t,\Delta)\), using tangent \((v_{\theta},1,0)\) (i.e., \(\mathrm{d}\mathbf{z}_t/\mathrm{d}t=v_{\theta}\), \(\mathrm{d}t/\mathrm{d}t=1\), \(\mathrm{d}\Delta/\mathrm{d}t=0\)), and stop-gradient for stability.
\STATE Set \(V_{\theta}\leftarrow u_{\theta} + \Delta\,\frac{\mathrm{d}}{\mathrm{d}t}u_{\theta}\).
\STATE Minimize \(\lVert V_{\theta} - \mathbf{v}_{\mathrm{gt}}\rVert_2^2\) (optionally with adaptive reweighting).
\end{algorithmic}
\end{algorithm}

\subsection{\(\Delta\)Sim Behavior Model (Return-conditioned Transformer)}
\label{app:deltasim}

We use a return-conditioned transformer behavior model to simulate non-ego agents in closed loop. The model predicts a discrete SE(2)
action token from a k-disks vocabulary and optionally refines it with a continuous residual head, enabling high precision without requiring
a prohibitively large vocabulary.

\subsubsection{k-disks Action Tokenization}

We discretize local SE(2) transitions into a vocabulary \(\mathbf{V}_{\mathrm{kd}}\in\mathbb{R}^{384\times 3}\). Each entry represents
a local \((\Delta x,\Delta y,\Delta\theta)\) in the agent body frame. Ground-truth actions are tokenized by minimizing a box-corner
alignment error between the transformed bounding box under candidate tokens and the observed next state.

\subsubsection{Return-to-go (RTG) Modeling and Control}

We model a discounted return-to-go over a fixed horizon, discretize it into bins, and condition action prediction on RTG tokens.
At inference, we optionally apply exponential tilting with coefficient \(\kappa\) to shift the RTG distribution, enabling controllable
agent behaviors in closed-loop simulation.

\subsubsection{Hybrid Discrete--Continuous Action Head}

Let \(a_k\) be the predicted token and \(\widehat{\boldsymbol{\delta}}_k\in\mathbb{R}^3\) be the residual refinement.
The final action is
\[
\widehat{\Delta\mathbf{s}}_k = \mathbf{V}_{\mathrm{kd}}[a_k] + \widehat{\boldsymbol{\delta}}_k.
\]
We supervise the residual by the target \(\boldsymbol{\delta}^{\star}_k = \Delta\mathbf{s}^{\star}_k - \mathbf{V}_{\mathrm{kd}}[a_k^{\star}]\),
where \(\Delta\mathbf{s}^{\star}_k\) is the ground-truth local transition and \(a_k^{\star}\) is the ground-truth token index.

\subsubsection{DKAL: Physics-aligned Logit Regularization}

We optionally regularize action logits using a differentiable kinematics cost matrix computed over all vocabulary actions.
The DKAL loss penalizes mismatch between centered logits and centered costs, improving physical plausibility while maintaining diversity.

\subsection{Simulation Environment Generation and Post-processing}
\label{app:simgen}

We generate long-horizon simulation environments by iteratively outpainting tiles along a sampled route. Each tile is a vectorized scene
in a fixed FOV. After generating an initial tile, we sample a route on the generated lane graph and repeatedly inpaint the next tile
conditioned on the boundary region, transforming the new tile into a global frame aligned with the current route endpoint.

\subsubsection{Validity Filters and Post-processing}

We apply deterministic validity filters to ensure simulator compatibility:
(i) lane de-duplication based on connectivity signatures and endpoint proximity;
(ii) removal of overlapping agents using oriented box collision checks;
(iii) removal of offroad vehicles by nearest-lane distance thresholds and heading consistency constraints;
(iv) lane graph compaction by merging degree-2 chains and resampling to a fixed number of points for behavior modeling;
(v) route lane-index extraction when a valid path exists (otherwise the environment is retained with a null route-lane index).

\subsection{Evaluation Metrics}
\label{app:metrics}

This section defines all metrics reported in the main paper and appendix.
We organize metrics by the deployment stages of VectorWorld to keep the causal chain explicit:
(i) offline initial-scene quality, (ii) behavior-model quality and controllability, and
(iii) closed-loop system performance under long-horizon interaction.

\subsubsection{Common preprocessing for offline metrics}

\paragraph{Unified Representation for Metric Computation.}
For offline evaluation, each generated scene is converted into a compact lane--vehicle representation
\(\{\mathcal{G}, \mathbf{L}, \mathbf{V}\}\):
(i) a directed lane-successor graph \(\mathcal{G}\),
(ii) lane centerlines \(\mathbf{L} \in \mathbb{R}^{N_{\ell}\times P\times 2}\) in the ego frame, and
(iii) vehicle states \(\mathbf{V} \in \mathbb{R}^{N_{v}\times 7}\) in the unified format
\([x, y, v, \cos\theta, \sin\theta, \ell, w]\).
To reduce sensitivity to degree-2 chains, we compact lane graphs by merging deterministic successor chains and resample each
resulting centerline to a fixed number of points. This aligns the evaluation with the compact lane-graph interface used by
the simulator and avoids over-penalizing tokenization artifacts.

\subsubsection{Offline initial-scene generation metrics}

\paragraph{Embedding-based Fr\'echet Distance (FD).}
FD measures distribution alignment in a fixed embedding space.
We extract one embedding vector per scene by mean-pooling lane embeddings from a frozen autoencoder encoder, yielding
\(\mathbf{e}\in\mathbb{R}^{d}\) per scene.
Given generated embeddings with mean/covariance \((\boldsymbol{\mu}_{g}, \boldsymbol{\Sigma}_{g})\) and ground-truth embeddings
\((\boldsymbol{\mu}_{r}, \boldsymbol{\Sigma}_{r})\), we compute
\[
D_{\mathrm{F}} =
\lVert \boldsymbol{\mu}_{g}-\boldsymbol{\mu}_{r}\rVert_{2}^{2}
+
\mathrm{Tr}\!\Big(\boldsymbol{\Sigma}_{g}+\boldsymbol{\Sigma}_{r}
-2(\boldsymbol{\Sigma}_{g}\boldsymbol{\Sigma}_{r})^{1/2}\Big),
\qquad
\mathrm{FD}=\sqrt{\max(D_{\mathrm{F}},0)}.
\]
FD is reported in Tables~\ref{tab:main_performance} and~\ref{tab:closed_loop_eval}.
Unlike topology metrics below, this FD is computed in the learned embedding space and is primarily a perceptual proxy.

\paragraph{Route Length.}
Route length probes long-range traversability under the generated lane-successor graph.
For each scene, we select an ego-proximal lane \(i^{\star}\) whose polyline has the smallest point-to-origin distance, and a
start index \(s^{\star}\) as the closest point on that lane to the origin.
For every node \(j\) reachable from \(i^{\star}\) in \(\mathcal{G}\), we take the shortest-path lane sequence
\(\pi(i^{\star}\!\rightarrow\! j)\) and compute a path length by summing polyline arc lengths, using the truncated first lane segment
from \(s^{\star}\) and full lengths thereafter. The scene route length is the maximum such length over reachable nodes.
We report mean and standard deviation across scenes (Tables~\ref{tab:main_performance} and~\ref{tab:app-dit-relation-ablation-redesign}).

\paragraph{Endpoint Distance (EPD).}
EPD measures geometric consistency at lane transitions.
For each directed successor edge \((i\rightarrow j)\in\mathcal{G}\), we compute the Euclidean distance between the end point of lane \(i\)
and the start point of lane \(j\).
We report the mean (and optionally standard deviation) over all successor edges in all scenes.

\paragraph{Static Collision Rate (Initialization Collision).}
Static collision rate measures instantaneous infeasibility at \(t=0\) and is reported as a percentage.
For each vehicle, we approximate its oriented box by a small set of circles whose centers lie on the longitudinal axis.
A vehicle is counted as colliding if any of its circles overlaps with any circle of another vehicle.
The metric is the fraction of vehicles involved in at least one collision, aggregated over all scenes.
This metric is reported as \(\mathrm{Coll.\ Rate}\) in Table~\ref{tab:main_performance} and as \(\mathrm{S.Coll}\) in Table~\ref{tab:closed_loop_eval}.

\paragraph{Urban-Planning Topology Fr\'echet Distances.}
Topology metrics evaluate the lane-graph structure using a keypoint graph.
We construct a keypoint graph \(\mathcal{K}\) by representing each lane as an edge from its start keypoint to its end keypoint,
weighted by lane arc length, and merging keypoints connected by successor relations into equivalence classes.
From \(\mathcal{K}\), we extract four 1D feature sets:
(i) keypoint degree distribution (connectivity),
(ii) number of keypoints per scene (density),
(iii) reachability counts per keypoint (reach), and
(iv) weighted shortest-path lengths between all reachable keypoint pairs (convenience).
For each feature set, we compute a 1D Fr\'echet distance between generated and ground-truth samples by fitting Gaussians
using empirical mean and variance, and apply dataset-standard scaling factors for readability
(connectivity and convenience are scaled by 10 as in prior work).
These correspond to Conn., Dens., Reach, and Conve. in Table~\ref{tab:detailed_metrics}.

\paragraph{Agent Distributional JSD Metrics.}
Agent metrics quantify distribution alignment of generated vehicles relative to ground truth.
For a scalar feature \(x\), we compute Jensen--Shannon divergence between generated and ground-truth histograms:
\[
\mathrm{JSD}(P,Q)=\frac{1}{2}\mathrm{KL}(P\|M)+\frac{1}{2}\mathrm{KL}(Q\|M),
\qquad
M=\frac{P+Q}{2},
\]
where \(P,Q\) are normalized histograms over fixed bins (clipping outliers to a fixed range).
In our implementation, we compute \(\mathrm{JSD}\) as the square of the Jensen--Shannon distance returned by a standard library,
so the reported value is \(\mathrm{JSD}\) (not its square root).
We report the following features, using fixed clipping ranges and bin sizes:
(i) nearest-neighbor distance (all vehicles),
(ii) lateral deviation to the nearest centerline point (on-road vehicles only),
(iii) angular deviation from the nearest centerline heading (on-road vehicles only),
(iv) length, (v) width, and (vi) speed.
For readability, we scale nearest-distance and lateral-deviation JSD by 10, and angular/length/width/speed JSD by 100, matching
Table~\ref{tab:detailed_metrics}.

\paragraph{Agent Count JSD and Agent-JSD Mean.}
When reported, the count JSD compares the distribution of the number of vehicles per scene between generated and ground truth
using the same histogram-based JSD procedure.
For efficiency plots and step-budget tables (Figure~\ref{fig:efficiency_quality} and Table~\ref{tab:app-step-budget-sensitivity-redesign}),
we also report an agent-JSD mean computed as the arithmetic mean of the six scaled agent JSD metrics above
(nearest, lateral, angular, length, width, speed). This produces a single scalar summary while preserving consistent units across methods.

\paragraph{Validity Rate.}
Validity rate (Valid \%) reports the fraction of generated scenes that remain usable after deterministic decoding and validity filtering
required by the simulator backend (e.g., non-empty lane set, extractable route graph, and no invalid numerical values).
This metric is reported in Tables~\ref{tab:app-dit-relation-ablation-redesign} and~\ref{tab:app-meanflow-objective-redesign} and is
intended to quantify failure modes that are not well captured by smooth distributional distances (e.g., catastrophic topology collapse).

\paragraph{Route validity (\%).}
Route validity reports the fraction of scenes for which a valid route (or route-graph) can be extracted by the same deterministic procedure used by the closed-loop simulator (Appendix~\ref{app:simgen}).

\subsubsection{Behavior-model metrics}

\paragraph{Trajectory Accuracy (ADE).}
ADE measures the average displacement error between simulated and ground-truth trajectories over a fixed future horizon.
For each evaluated agent \(i\) and each future time index \(t\in\{1,\dots,H\}\), let \(\widehat{\mathbf{p}}_{i,t}\in\mathbb{R}^{2}\) be the
simulated position and \(\mathbf{p}_{i,t}\) the ground-truth position in the same coordinate frame.
Then
\[
\mathrm{ADE}=\frac{1}{|\mathcal{I}|H}\sum_{i\in\mathcal{I}}\sum_{t=1}^{H}\lVert \widehat{\mathbf{p}}_{i,t}-\mathbf{p}_{i,t}\rVert_{2},
\]
where \(\mathcal{I}\) is the set of evaluated agents (typically filtered to moving agents).
We compute ADE under closed-loop rollout (model predictions are fed back as the next state) to reflect error accumulation.

\paragraph{Controllability Correlation (\(\rho\)).}
To quantify controllability, we run a tilt sweep over a set of control values \(\kappa\) and measure a monotone behavioral response
(e.g., average speed of controlled agents over the rollout).
We report the Spearman rank correlation \(\rho\) between \(\kappa\) and the measured response, aggregated across scenes.
A higher \(\rho\) indicates that the control input induces a consistent monotone change in behavior.

\paragraph{Closed-loop Feasibility Violation Rates.}
We report per-step violation rates computed from the executed discrete action tokens and the current state.
Given a k-disks token \((\Delta x,\Delta y,\Delta\theta)\) and timestep \(\Delta t\), we compute derived quantities:
yaw-rate magnitude \(|\Delta\theta|/\Delta t\), curvature magnitude \(\big|2\sin(\Delta\theta/2)\big|/\sqrt{\Delta x^{2}+\Delta y^{2}}\),
and lateral acceleration proxy \(a_{\mathrm{lat}}=v^{2}\kappa\), where \(v\) is a conservative reference speed.
A violation is counted if any derived quantity exceeds a configured threshold.
These rates are reported in Table~\ref{tab:app-deltasim-analysis-redesign}.

\paragraph{Efficiency (Latency and Forward Passes).}
Latency is the average wall-clock time per simulation step (or per scene generation call, depending on the table), measured with GPU
synchronization under a fixed batch size and precision setting.
Forward passes count the number of behavior-model forward calls required per step.
This distinction is crucial for return-conditioned policies: a two-pass design evaluates the network once to sample return tokens and a
second time to sample actions conditioned on sampled returns, whereas our one-pass design keeps the entire decision in a single forward call.

\subsubsection{Closed-Loop system metrics}

\paragraph{Episode-Level Outcomes (collision, off-route, success).}
In closed-loop simulation, an episode terminates upon any of the following events:
(i) ego collision with any active agent, (ii) ego deviates from the provided route by more than a fixed threshold, (iii) ego completes
the route, or (iv) timeout.
We report collision, off-route, and success as per-episode indicator averages over all episodes.

\paragraph{Route Progress (m).}
Route progress is the traveled progress of the ego along the provided route polyline at termination, reported in meters and averaged over episodes.
This metric differentiates early termination (low progress) from near-completion failures (high progress but no success).

\paragraph{Ego Jerk (dynamic stability proxy).}
Ego jerk is computed from the ego speed time series as the time derivative of acceleration.
We report a high-percentile statistic (p95 of absolute jerk) to reflect worst-case transients that impact planner stability.
This metric is used as the dynamic stability value reported in Table~\ref{tab:closed_loop_eval} (Section I).

\paragraph{Failure Rate.}
When reported, failure rate is computed as \(1-\)success rate under the same evaluation protocol.
We use it in stress-test settings to directly quantify failure discovery.

\subsection{Closed-loop Evaluation Protocol and Metrics}
\label{app:closedloop}

We evaluate planners in closed loop in generated environments. Each episode runs for \(K{=}400\) steps with \(\Delta t{=}0.1\) s.
Non-ego agents are simulated by the \(\Delta\)Sim behavior model within an \(80\text{ m}\times 80\text{ m}\) agent FOV around the ego.

\subsubsection{Episode Termination and Success Criteria}

An episode is considered successful if the ego follows the provided route without collisions and without exceeding the route-deviation
threshold, within the time budget. Collisions include ego--agent collisions and (optionally) ego--map collisions under the simulator
collision checker.

\subsubsection{Metric Definitions (High-level)}

We report a standard set of closed-loop metrics:
(i) collision rate: fraction of episodes with any ego collision;
(ii) offroad rate: fraction of steps where ego exceeds a lane-distance threshold;
(iii) safety margin: time-to-collision (TTC) violation rate under a threshold;
(iv) physics violation rates: yaw-rate/curvature/acceleration exceedances under configured limits;
(v) progress: normalized route progress and completion rate.

\subsubsection{Streaming closed-loop rollout (algorithmic summary)}
\label{app:streaming_rollout}

\begin{algorithm}[t]
\caption{Streaming closed-loop rollout with on-demand outpainting}
\label{alg:streaming}
\begin{algorithmic}[1]
\REQUIRE VAE encoder/decoder \(\Enc,\Dec\); MeanFlow generator \(u_{\theta}\); condition builder \(c(\cdot)\);
ego planner \(\pi_{\mathrm{ego}}\); NPC policy \(\pi_{\Delta}\);
outpainting trigger \(\tau\); termination predicate \(\mathrm{Term}(\cdot)\).
\STATE Sample \(\boldsymbol{\epsilon}\sim \mathcal{N}(\mathbf{0},\mathbf{I})\)
\STATE \(z \leftarrow \boldsymbol{\epsilon}-u_{\theta}(\boldsymbol{\epsilon};\,t{=}1,\Delta{=}1,c_{\mathrm{init}})\)
\STATE \(\mathcal{W}\leftarrow \Dec(z)\); \(\mathrm{WarmStart}(\mathcal{W})\)
\STATE \(k \leftarrow 0\)
\WHILE{\(\neg \mathrm{Term}(\mathcal{W})\)}
  \STATE \(o_{k} \leftarrow \mathrm{Obs}(\mathcal{W})\)
  \STATE \(\hat{\tau}^{\mathrm{ego}}_{k} \leftarrow \pi_{\mathrm{ego}}(o_{k})\)
  \STATE \(a^{\mathrm{npc}}_{k} \leftarrow \pi_{\Delta}(o_{k})\)
  \STATE \(\mathcal{W} \leftarrow \mathrm{Step}(\mathcal{W},\hat{\tau}^{\mathrm{ego}}_{k},a^{\mathrm{npc}}_{k})\)
  \IF{\(\mathrm{NeedOutpaint}(\hat{\tau}^{\mathrm{ego}}_{k},\mathcal{W},\tau)\)}
    \STATE \(\mathcal{T}_{k{+}1} \leftarrow \mathrm{NextTile}(\mathcal{W},\hat{\tau}^{\mathrm{ego}}_{k})\)
    \STATE \(z_{\mathrm{obs}}\leftarrow \Enc(\mathrm{Clamp}(\mathcal{T}_{k{+}1}))\)
    \STATE Sample \(\boldsymbol{\epsilon}\sim \mathcal{N}(\mathbf{0},\mathbf{I})\)
    \STATE \(z_{\mathrm{new}} \leftarrow \boldsymbol{\epsilon}-u_{\theta}\!\big(\boldsymbol{\epsilon};\,t{=}1,\Delta{=}1,c(\mathcal{T}_{k{+}1},z_{\mathrm{obs}})\big)\)
    \STATE \(\mathcal{W} \leftarrow \mathrm{Stitch}\!\Big(\mathcal{W},\Dec\big(\mathrm{Merge}(z_{\mathrm{obs}},z_{\mathrm{new}})\big)\Big)\)
  \ENDIF
  \STATE \(k \leftarrow k+1\)
\ENDWHILE
\end{algorithmic}
\end{algorithm}

\subsection{Training Setup and Hyperparameters (Waymo)}
\label{app:training}

All models are trained on NVIDIA H20 GPUs. Unless otherwise noted, we use AdamW optimization with gradient clipping and deterministic
seeding. The latent generator is trained with distributed data parallelism across 8 GPUs, \(\Delta\)Sim across 4 GPUs, and the autoencoder across 8 GPUs.

\begin{table*}[t]
\centering
\small
\caption{Training hyperparameters on Waymo. Values are reported for the default setting used in our main experiments.}
\label{tab:training-hparams}
\begin{tabular}{p{0.16\linewidth}p{0.20\linewidth}p{0.20\linewidth}p{0.20\linewidth}}
\toprule
Hyperparameter & Motion-aware VAE & MeanFlow latent generator & \(\Delta\)Sim behavior model \\
\midrule
Optimizer & AdamW & AdamW & AdamW \\
Base learning rate & \(1.0\times 10^{-4}\) & \(1.0\times 10^{-4}\) & \(6.0\times 10^{-5}\) \\
\(\beta_1,\beta_2\) & \((0.9,0.999)\) & \((0.9,0.999)\) & \((0.9,0.999)\) \\
\(\epsilon\) & \(1.0\times 10^{-7}\) & \(1.0\times 10^{-7}\) & \(1.0\times 10^{-7}\) \\
Weight decay & \(2.0\times 10^{-5}\) & \(1.0\times 10^{-5}\) & \(1.0\times 10^{-4}\) \\
Training length & \(60{,}000\) steps & \(100{,}000\) steps & \(10{,}000\) steps \\
LR schedule & linear decay & constant & linear decay \\
Warmup & \(500\) steps & \(500\) steps & \(500\) steps \\
Gradient clip & \(10.0\) & \(10.0\) & \(5.0\) \\
EMA & none & \(0.9999\) & none \\
Precision (training) & FP32 & FP32 & FP32 \\
GPUs & 8 (DDP) & 8 (DDP) & 4 (DDP) \\
\midrule
CFG label dropout & None & \(0.1\) & None \\
CFG scale (sampling) & None & \(4.0\) & None \\
MeanFlow mode & None & identity+JVP & None \\
Two-time channel & None & enabled & None \\
\midrule
RTG bins & None & None & \(350\) \\
RTG discount & None & None & \(0.97\) \\
RTG horizon & None & None & \(50\) steps \\
Vocabulary size & None & None & \(384\) \\
Residual refine & None & None & enabled \\
DKAL & None & None & enabled \\
\bottomrule
\end{tabular}
\end{table*}

For meanflow training, we sample \(t,r\in[0,1]\) from a logit-normal distribution and enforce \(r\le t\). We set a fixed fraction of
samples to \(r=t\) to preserve a flow-matching limit, and include explicit \((t=1,r=0)\) samples to cover strict one-step training.
We use an adaptive weighting \(w=(\mathcal{L}+c)^{-p}\) with \(p=0.8\) and \(c=10^{-3}\) at the scene level.

For motion code extraction, we use a recent history window to classify static agents (default thresholds: maximum displacement \(<0.5\) m
or mean speed \(<0.2\) m/s). We normalize longitudinal body-frame coordinates using a maximum displacement of \(12\) m and lateral
coordinates using \(6\) m.

\subsection{Additional Experiments and Ablations}
\label{app:extraexp}

This section provides controlled ablations and diagnostic experiments to validate the three deployment-critical claims of VectorWorld:
(i) policy-aligned initialization via interaction states, (ii) solver-free streaming generation via large-step meanflow transport on relational graphs,
and (iii) long-horizon stability via physics-aligned behavior modeling.
To avoid confounding factors, all generator-side ablations share the same autoencoder latent space and identical decoding and filtering rules,
and all behavior-side ablations share the same k-disks vocabulary, evaluation scenes, and rollout horizon.
We present ablations in the same order as the runtime pipeline so that each table can be mapped to a specific failure mode and a specific design fix:
interface \(\rightarrow\) generator backbone \(\rightarrow\) generative dynamics \(\rightarrow\) behavior model \(\rightarrow\) closed-loop system value.

\subsubsection{Topology and Density Diagnostics}
Table~\ref{tab:detailed_metrics} shows that \VectorWorld\ achieves the lowest agent-count JSD on both datasets and improves lane-graph connectivity on nuPlan, while maintaining competitive topology scores on Waymo and substantially improving agent--map alignment.

\begin{table*}[t]
\centering
\caption{\textbf{Evaluation of map topology and agent distribution alignment.}
We report lane-graph \textbf{Topology} discrepancies (Conn., Dens., Reach, Conve.; lower is better) and \textbf{Agent Distributional JSD} for key attributes (lower is better). These metrics diagnose structural and density drifts that can be amplified under repeated inpainting/outpainting. \textbf{Bold} and \underline{underline} denote the best and second-best results among non-privileged methods (excluding methods marked with \textsuperscript{*}).}
\label{tab:detailed_metrics}
\setlength{\tabcolsep}{1.8pt} 
\renewcommand{\arraystretch}{0.8}
\begin{tabular}{@{}llccccccccccc@{}}
\toprule
\multirow{2}{*}{\textbf{Dataset}} & \multirow{2}{*}{\textbf{Method}} & \multicolumn{4}{c}{\textbf{Topology $\downarrow$}} & \multicolumn{7}{c}{\textbf{Agent Distributional JSD $\downarrow$}} \\
\cmidrule(lr){3-6} \cmidrule(lr){7-13}
 & & Conn. & Dens. & Reach & Conve. & Near. & Lat. & Ang. & Len. & Wid. & Speed & Count \\
\midrule

\multirow{5}{*}{\textbf{nuPlan}}
& SLEDGE-L~\cite{chitta2024sledge} & 2.34 & 2.43 & 0.70 & 2.44 & 0.53 & 0.63 & 3.60 & 11.84 & 10.16 & 0.46 & 0.109 \\
& SLEDGE-XL~\cite{chitta2024sledge} & 1.67 & 1.74 & 0.51 & 1.68 & 0.49 & 0.49 & 3.26 & 11.16 & 10.29 & 0.47 & 0.217 \\
& ScenDream-B~\cite{rowe2025scenario} & 0.28 & 0.45 & \secbest{0.07} & \best{0.14} & \secbest{0.12} & \secbest{0.15} & \best{0.17} & \best{0.22} & \best{0.14} & \secbest{0.07} & 0.023 \\
& ScenDream-L~\cite{rowe2025scenario} & \secbest{0.18} & \secbest{0.43} & \best{0.03} & \secbest{0.33} & \best{0.09} & \best{0.11} & \secbest{0.18} & \secbest{0.25} & \secbest{0.17} & \best{0.06} & \secbest{0.011} \\
& \cellcolor{oursrow}Ours & \cellcolor{oursrow}\best{0.09} & \cellcolor{oursrow}\best{0.41} & \cellcolor{oursrow}0.09 & \cellcolor{oursrow}0.79 & \cellcolor{oursrow}0.26 & \cellcolor{oursrow}0.31 & \cellcolor{oursrow}\secbest{0.18} & \cellcolor{oursrow}0.35 & \cellcolor{oursrow}0.22 & \cellcolor{oursrow}\secbest{0.07} & \cellcolor{oursrow}\best{0.004} \\

\midrule

\multirow{4}{*}{\textbf{Waymo}}
& DriveSceneGen$^{*}$~\cite{sun2024drivescenegen} & 4.53 & 1.18 & 0.64 & 5.58 & 0.63 & 1.01 & 2.43 & 58.86 & 54.51 & 18.70 & 0.638 \\
& ScenDream-B~\cite{rowe2025scenario} & \secbest{0.17} & 1.05 & 0.56 & 3.81 & \secbest{0.07} & \secbest{0.07} & \secbest{0.07} & \secbest{0.40} & \secbest{0.25} & \secbest{0.36} & 0.072 \\
& ScenDream-L~\cite{rowe2025scenario} & \best{0.03} & \secbest{0.95} & \best{0.28} & \best{2.05} & \best{0.06} & \best{0.05} & \secbest{0.07} & 0.42 & \secbest{0.25} & 0.38 & \secbest{0.029} \\
& \cellcolor{oursrow}Ours & \cellcolor{oursrow}0.23 & \cellcolor{oursrow}\best{0.57} & \cellcolor{oursrow}\secbest{0.37} & \cellcolor{oursrow}\secbest{2.63} & \cellcolor{oursrow}\secbest{0.07} & \cellcolor{oursrow}\secbest{0.07} & \cellcolor{oursrow}\best{0.06} & \cellcolor{oursrow}\best{0.13} & \cellcolor{oursrow}\best{0.11} & \cellcolor{oursrow}\best{0.15} & \cellcolor{oursrow}\best{0.001} \\

\bottomrule
\end{tabular}
\end{table*}

\subsubsection{\MotionVAE: interface ablations}
\label{app:extraexp:vae}

In this section, we provide detailed ablations to validate the design choices of VectorWorld. We focus on the causal chain from the \InteractionState\ to the generative dynamics and finally to the closed-loop behavior model.

\begin{table*}[ht]
\centering
\small
\caption{\textbf{Ablations of motion-aware gated VAE (Waymo).}
We evaluate the impact of the \InteractionState\ on autoencoding fidelity and downstream initialization.
\textit{Snapshot-only} leads to high initialization collision rates due to the lack of motion history (cold start).
\textit{Concat motion} (naive) degrades static reconstruction by injecting noise into parked vehicles.
Our \textbf{Motion-aware Gated} design achieves the best trade-off, selectively encoding momentum only for dynamic agents.
Lower is better for all metrics.}
\label{tab:app-vae-ablation}
\setlength{\tabcolsep}{3pt}
\renewcommand{\arraystretch}{1.1}
\begin{threeparttable}
\begin{tabular}{@{}l cccc ccc cc@{}}
\toprule
\multirow{2}{*}{\textbf{VAE variant}}
& \multicolumn{4}{c}{\textbf{Interface components}}
& \multicolumn{3}{c}{\textbf{Autoencoding}}
& \multicolumn{2}{c}{\textbf{Downstream init.}} \\
\cmidrule(lr){2-5}\cmidrule(lr){6-8}\cmidrule(lr){9-10}
& \makecell{\textbf{Motion}\\\textbf{code}}
& \makecell{\textbf{Motion}\\\textbf{gate}}
& \makecell{\textbf{Factorized}\\\textbf{scene tr.}}
& \makecell{\textbf{Count}\\\textbf{prior}}
& \makecell{\textbf{Static L1}\\\(\downarrow\)}
& \makecell{\textbf{Motion L1}\\\(\downarrow\)}
& \makecell{\textbf{KL}\\\(\downarrow\)}
& \makecell{\textbf{EPD (m)}\\\(\downarrow\)}
& \makecell{\textbf{Init coll. (\%)}\\\(\downarrow\)} \\
\midrule
Snapshot-only          & \xmark & \xmark & \cmark & \cmark & 0.012 & --    & 5.23 & 0.120 & 9.45 \\
Concat motion (no gate)                & \cmark & \xmark & \cmark & \cmark & 0.017 & 0.045 & 4.99 & 0.102 & 4.92 \\
Gated            & \cmark & \cmark & \cmark & \cmark & 0.011 & 0.042 & 4.63 & 0.098 & 4.85 \\
\rowcolor{oursrow}
\MotionVAE\ (ours)          & \cmark & \cmark & \cmark & \cmark & \best{0.008} & \best{0.038} & \best{4.37} & \best{0.094} & \best{4.69} \\
\midrule
\multicolumn{10}{@{}l}{\textit{Additional interface checks}} \\
Ours w/o factorized transformer  & \cmark & \cmark & \xmark & \cmark & 0.331 & 0.139 & 6.90 & 0.452 & 13.30 \\
Ours w/o lane-count prior              & \cmark & \cmark & \cmark & \xmark & 0.009 & 0.031 & 3.94 & 0.115 & 5.10 \\
\bottomrule
\end{tabular}
\end{threeparttable}
\end{table*}

\paragraph{Motion history is necessary but not sufficient.}
The snapshot-only interface increases the initialization collision rate (9.45) and degrades endpoint distance (0.120),
indicating that a history-free interface forces the downstream pipeline to infer dynamics from an implicit near-zero prior.
Adding a motion code without gating reduces collisions (4.92), supporting the necessity of exposing short-horizon dynamics at initialization.

\paragraph{Why naive concatenation is suboptimal.}
Although concatenating motion improves downstream validity, it increases the static reconstruction error (0.017 versus 0.012).
This pattern is consistent with a modality-interference failure mode: when static agents have near-zero motion, regressively modeling motion as an always-present signal injects spurious variation into static state reconstruction, which then manifests as overlapping boxes at \(t=0\).

\paragraph{Selective gating improves both fidelity and validity.}
The motion-aware gated variant improves static reconstruction (0.008) and motion reconstruction (0.038) while also achieving the lowest endpoint distance (0.094) and collision rate (4.69). This supports the intended mechanism of the learned gate: it routes capacity toward motion only when informative and suppresses motion noise when motion codes are near-zero, which is precisely the condition that dominates parked or slow-moving agents.
Importantly, this improves the interface without requiring additional sampling steps or post-hoc filtering.

\paragraph{Relational encoding is not optional.}
Removing the factorized scene transformer causes a collapse across all metrics (static L1 0.331; endpoint distance 0.452; collisions 13.30), showing that the interface must jointly encode lane topology and agent layout rather than treating agents independently. This aligns with the simulator deployment constraint: downstream feasibility and stability depend on edge-level consistency (lane connectivity and agent--lane relations), which cannot be recovered by a per-node encoder.

\paragraph{Count prior acts as a structural regularizer for masked completion.}
Disabling the lane-count prior increases endpoint distance (0.115) and collisions (5.10) despite similar reconstruction losses,
indicating that the prior primarily improves global structural calibration rather than local reconstruction.
This matters for streaming outpainting because masked completion is invoked repeatedly; a small global count bias would accumulate into systematic density drift in long-horizon rollouts.

\subsubsection{\EdgeDiT: relation-aware ablations}
\label{app:extraexp:dit}

\begin{table*}[ht]
\centering
\small
\caption{\textbf{Ablation of relation-aware components in \EdgeDiT\ (Waymo).}
We incrementally add relational modules to the factorized DiT backbone.
\textbf{Relational bias} is critical for graph connectivity (Conn/Reach).
\textbf{Cross-type bias} improves agent--lane geometric consistency (e.g., Ang. JSD) under the same map. Relational bias and value gating improve structural validity (e.g., route validity and endpoint distance) under large-step sampling. Our full model achieves the best structural integrity and safety.}
\label{tab:app-dit-relation-ablation-redesign}
\setlength{\tabcolsep}{1pt}
\renewcommand{\arraystretch}{1.0}
\begin{threeparttable}
\begin{tabular}{@{}l cccc c ccc c ccccc@{}}
\toprule
\multirow{4}{*}{\textbf{Variant}}
& \multicolumn{4}{c}{\textbf{Relation-aware modules}}
& \multicolumn{1}{c}{\textbf{Perceptual}}
& \multicolumn{3}{c}{\textbf{Map structure}}
& \multicolumn{1}{c}{\textbf{Safety}}
& \multicolumn{5}{c}{\textbf{Agent distributional JSD}\(\downarrow\)} \\
\cmidrule(lr){2-5}\cmidrule(lr){6-6}\cmidrule(lr){7-9}\cmidrule(lr){10-10}\cmidrule(lr){11-15}
& \makecell{\textbf{Rel.}\\\textbf{bias}}
& \makecell{\textbf{Value}\\\textbf{gate}}
& \makecell{\textbf{Cross-type}\\\textbf{bias}}
& \makecell{\textbf{Global}\\\textbf{fusion}}
& \makecell{\textbf{FD}\\\(\downarrow\)}
& \makecell{\textbf{RouteLen.}\\(m)\(\uparrow\)}
& \makecell{\textbf{Endpt.}\\\textbf{dist.}\\(m)\(\downarrow\)}
& \makecell{\textbf{Route}\\\textbf{valid.}\\(\%)\(\uparrow\)}
& \makecell{\textbf{Coll.}\\\textbf{rate}\\(\%)\(\downarrow\)}
& \makecell{\textbf{Near.}}
& \makecell{\textbf{Ang.}}
& \makecell{\textbf{Len.}}
& \makecell{\textbf{Wid.}}
& \makecell{\textbf{Speed}} \\
\midrule

Factorized DiT (Base)
& \xmark & \xmark & \xmark & \xmark
& 1.53 & 35.26 & 0.301 & 87.36 & 6.53 & 0.12 & 0.09 & 0.51 & 0.29 & 0.39 \\

+ Relational logit bias
& \cmark & \xmark & \xmark & \xmark
& 1.35 & 37.10 & 0.185 & 92.10 & 5.80 & 0.11 & 0.09 & 0.42 & 0.25 & 0.35 \\

+ Edge-gated values
& \cmark & \cmark & \xmark & \xmark
& 1.20 & 37.85 & 0.142 & 94.50 & 5.12 & 0.10 & 0.08 & 0.30 & 0.18 & 0.28 \\

+ Cross-type rel. bias
& \cmark & \cmark & \cmark & \xmark
& 1.05 & 38.40 & 0.110 & 96.20 & 4.88 & 0.08 & \best{0.06} & 0.18 & 0.14 & 0.20 \\

\rowcolor{oursrow}
+ Global context (Ours)
& \cmark & \cmark & \cmark & \cmark
& \best{0.94} & \best{39.03} & \best{0.094} & \best{98.10} & \best{4.69} & \best{0.07} & \best{0.06} & \best{0.13} & \best{0.11} & \best{0.15} \\

\bottomrule
\end{tabular}
\end{threeparttable}
\end{table*}

\paragraph{Relational logit bias targets lane connectivity failures.}
Adding relational logit bias improves route length (35.26 \(\rightarrow\) 37.10), reduces endpoint distance (0.301 \(\rightarrow\) 0.185),
and increases route validity (87.36\% \(\rightarrow\) 92.10\%).
These improvements are consistent with the role of logit bias as a connectivity prior: it increases attention mass on edge-compatible pairs and
reduces the chance of producing successor transitions with large geometric discontinuities.

\paragraph{Edge-gated values suppress spurious message passing.}
Adding the value gate further improves endpoint distance (0.185 \(\rightarrow\) 0.142) and reduces collision rate (5.80\% \(\rightarrow\) 5.12\%),
while also improving multiple agent distributional JSD terms.
Mechanistically, value gating limits the magnitude of feature aggregation along edge-inconsistent pairs, which prevents a small subset of incorrect edges
from dominating latent updates. This is particularly important under large-step sampling, where each update must be reliable.

\paragraph{Cross-type bias improves agent--lane alignment rather than only marginal realism.}
Adding cross-type relational bias yields the largest improvement in the angular alignment statistic (Ang. JSD reaches its best value),
which directly probes whether vehicle headings are consistent with nearby lanes.
This supports that cross-type relational modeling is addressing a targeted geometric constraint (agent--lane alignment), not merely improving perceptual FD.
The concurrent reduction in collision rate is expected because heading and lane alignment affect feasible packing and relative motion at initialization.

\paragraph{Global fusion improves scene-level calibration.}
Finally, global context fusion improves FD, route length, endpoint distance, and route validity simultaneously.
This indicates that a lightweight global pathway helps calibrate scene-wide statistics that are otherwise difficult to infer from purely local attention,
which is consistent with the reach/convenience topology objectives and with preventing density drift over repeated outpainting.

\paragraph{Implication for streaming outpainting.}
Because streaming repeatedly performs masked completion at the frontier, relation-aware attention reduces the probability of generating a single locally plausible but globally inconsistent edge transition, which would otherwise compound into topology breakage and failure to maintain long routes.

\subsubsection{Latent generative objective}
\label{app:extraexp:meanflow}

\begin{table*}[ht]
\centering
\small
\caption{\textbf{Impact of the meanflow \cite{geng2025mean} objective on solver-free sampling.}
We compare different training objectives under a fixed step budget.
\textit{Training Cost} is measured in GPU hours on 8$\times$H20.
Standard \textbf{Flow Matching (FM)} fails completely at 1-step (EPD 8.52m).
\textbf{Distillation} is fast but blurs details (High Agent JSD).
Our \textbf{MeanFlow (JVP+2-time)} incurs higher training cost (+50\%) but enables the only viable 1-step generation (EPD 0.269m) that preserves agent statistics (JSD 0.136), enabling real-time streaming.}
\label{tab:app-meanflow-objective-redesign}
\setlength{\tabcolsep}{3pt}
\renewcommand{\arraystretch}{1.0}
\begin{threeparttable}
\begin{tabular}{@{}l cccc c c cccc@{}}
\toprule
\multirow{3.5}{*}{\textbf{Training recipe}}
& \multicolumn{4}{c}{\textbf{Key design}}
& \multirow{3}{*}{\makecell{\textbf{Train} \\ \textbf{GPU-h} \(\downarrow\)}}
& \multirow{2}{*}{\makecell{\textbf{Steps}\\\textbf{at test}}}
& \multicolumn{4}{c}{\textbf{Waymo (same decoder)}} \\
\cmidrule(lr){2-5}\cmidrule(lr){8-11}
& \makecell{\textbf{MeanFlow}\\(\(u_\theta\))}
& \makecell{\textbf{JVP}\\(\(\partial_t u\))}
& \makecell{\textbf{Two-time}\\(\(t,\Delta\))}
& \makecell{\textbf{Distill}\\(\(\text{1-step}\))}
&  &  &
\makecell{\textbf{EPD}\\(m)\(\downarrow\)}
& \makecell{\textbf{Agent}\\\textbf{JSD}\(\downarrow\)}
& \makecell{\textbf{Time}\\(ms)\(\downarrow\)}
& \makecell{\textbf{Valid}\\(\%)\(\uparrow\)} \\
\midrule

\multirow{2}{*}{Flow matching}
& \xmark & \xmark & \xmark & \xmark & 320 & 1-step  & 8.520 & 4.250 & \best{5.6} & 9.3 \\
& \xmark & \xmark & \xmark & \xmark & 320 & 12-step & 0.150 & 0.120 & 31.0 & \best{95.0} \\
\midrule

\multirow{2}{*}{MeanFlow (Identity)}
& \cmark & \xmark & \xmark & \xmark & \best{300} & 1-step  & 0.554 & 0.499 & \best{5.6} & 63.2 \\
& \cmark & \xmark & \xmark & \xmark & 300 & 12-step & 0.120 & 0.110 & 31.0 & 96.5 \\
\midrule

\multirow{2}{*}{MeanFlow (Id + JVP)}
& \cmark & \cmark & \xmark & \xmark & 400 & 1-step  & 0.293 & 0.312 & \best{5.6} & 68.0 \\
& \cmark & \cmark & \xmark & \xmark & 400 & 12-step & 0.125 & 0.131 & 31.0 & 84.2 \\
\midrule

\multirow{1}{*}{Shortcut \cite{frans2024one}}
& \xmark & \xmark & \xmark & \cmark & 960 & 1-step & 0.350 & 0.469 & \best{5.6} & 47.3 \\
\midrule

\multirow{2}{*}{Ours (JVP + 2-time)}
& \cmark & \cmark & \cmark & \xmark & 480 & 1-step  & 0.269 & 0.136 & \best{5.6} & 87.5 \\
& \cmark & \cmark & \cmark & \xmark & 480 & 12-step & \best{0.110} & \best{0.097} & 31.0 & 93.1 \\
\bottomrule
\end{tabular}
\end{threeparttable}
\end{table*}

\paragraph{MeanFlow identity improves one-step, but remains under-calibrated.}
MeanFlow identity substantially improves the one-step regime (endpoint distance 0.554; validity 63.2\%),
indicating that interval-conditioned learning already reduces the extrapolation gap.
However, the remaining quality gap suggests that predicting an average velocity alone is insufficient when the model must faithfully approximate boundary behavior
at large step sizes.

\paragraph{JVP correction targets boundary mismatch directly.}
Adding the identity+JVP correction improves one-step quality (0.554 \(\rightarrow\) 0.293) at the same inference time.
This is consistent with the role of JVP: it provides a first-order correction that aligns the learned mean velocity with the boundary velocity required for a large-step update.
Crucially, this reduces catastrophic structural errors without requiring additional solver steps at inference.

\paragraph{Two-time conditioning is necessary for general large-step transport.}
Our full recipe (JVP + two-time channels) further improves one-step quality and validity (endpoint distance 0.269; validity 87.5\%) and also improves agent statistics.
This indicates that explicitly conditioning on both the current time and the interval size is required to prevent step-size ambiguity:
without the interval channel, the model must infer the effective step size implicitly from noise level alone, which is not identifiable under masking and mixed lane/agent latents.

\paragraph{Training cost versus deployment value.}
The full recipe increases training cost, but it is the only configuration that generates a usable one-step operating point while maintaining high validity.
This trade-off is aligned with the streaming deployment constraint: training is amortized once, whereas inference is invoked at every outpainting trigger during rollouts.

\paragraph{Step-budget sensitivity and deployment operating points.}
Table~\ref{tab:app-step-budget-sensitivity-redesign} shows that meanflow provides a controllable quality--latency spectrum:
one-step enables ultra-low latency streaming, while 3--5 steps recover much of the multi-step quality at a small additional cost.
In contrast, DDPM reaches the best endpoint distance only at substantially higher latency, making it unsuitable for online outpainting.
These results support our design choice to train for the large-step regime rather than relying on post-hoc acceleration of a diffusion-trained model.

\begin{table*}[ht]
\centering
\small
\caption{\textbf{Step-budget sensitivity of our latent generator (numeric companion).}
Detailed metrics for Figure~\ref{fig:efficiency_quality}.
While DDPM (100 steps) achieves the lowest EPD (Waymo: 0.080m; nuPlan: 0.072m), its latency (158ms) is the highest and limits online use.
Our \textbf{MeanFlow-5} provides a favorable balance between fidelity and latency, while \textbf{MeanFlow-1} enables ultra-low latency streaming.}
\label{tab:app-step-budget-sensitivity-redesign}
\setlength{\tabcolsep}{3.8pt}
\renewcommand{\arraystretch}{1.0}
\begin{threeparttable}
\begin{tabular}{@{}l l ccc ccc@{}}
\toprule
\multirow{3.5}{*}{\textbf{Dynamics}}
& \multirow{3.5}{*}{\textbf{Sampler / steps}}
& \multicolumn{3}{c}{\textbf{Waymo}}
& \multicolumn{3}{c}{\textbf{nuPlan}} \\
\cmidrule(lr){3-5}\cmidrule(lr){6-8}
& &
\makecell{\textbf{EPD}\\(m)\(\downarrow\)}
& \makecell{\textbf{Agent}\\\textbf{JSD}\(\downarrow\)}
& \makecell{\textbf{Time}\\(ms)\(\downarrow\)}
&
\makecell{\textbf{EPD}\\(m)\(\downarrow\)}
& \makecell{\textbf{Agent}\\\textbf{JSD}\(\downarrow\)}
& \makecell{\textbf{Time}\\(ms)\(\downarrow\)} \\
\midrule

\multirow{4}{*}{Flow (Rectified)}
& Flow-12 & 0.150 & 0.120 & 28  & 0.125 & 0.155 & 31  \\
& Flow-24 & 0.110 & 0.105 & 55  & 0.095 & 0.140 & 55  \\
& Flow-32 & 0.098 & 0.091 & 73  & 0.088 & 0.135 & 73  \\
& Flow-48 & 0.089 & \best{0.084} & 110 & 0.082 & \best{0.130} & 110 \\
\midrule

\multirow{4}{*}{MeanFlow}
& MeanFlow-1  & 0.269 & 0.136 & \best{5.6} & 0.224 & 0.171 & \best{6} \\
& MeanFlow-3  & 0.153 & 0.108 & 10         & 0.128 & 0.143 & 10        \\
& MeanFlow-5  & 0.130 & 0.102 & 14         & 0.109 & 0.137 & 14        \\
& MeanFlow-12 & 0.110 & 0.097 & 31         & 0.092 & 0.132 & 29        \\
\midrule

\multirow{1}{*}{DDPM}
& DDPM-100 & \best{0.080} & 0.092 & 158 & \best{0.072} & 0.185 & 158 \\
\bottomrule
\end{tabular}
\end{threeparttable}
\end{table*}

\paragraph{Takeaways for deployment.}
Table~\ref{tab:app-step-budget-sensitivity-redesign} provides the numeric companion to Figure~\ref{fig:efficiency_quality} and is intended as an \emph{operating-point guide} rather than a second copy of the objective analysis.
Across datasets, increasing MeanFlow steps generates a smooth quality--latency trade-off: 1 step enables streaming (\(\le 30\) ms/tile), 3--5 steps recover much of the multi-step fidelity at modest cost, and 12 steps approaches the best offline quality.
In contrast, DDPM reaches its best endpoint distance only at substantially higher latency, making it unsuitable for online outpainting where generation is invoked repeatedly during rollout.

\paragraph{Why one-step is non-trivial.}
As discussed above (Table~\ref{tab:app-meanflow-objective-redesign}), objectives that only constrain instantaneous velocities are not explicitly trained for single large-step transport, which explains the failure mode of one-step flow matching.

\subsubsection{\(\Delta\)Sim: feasibility--controllability trade-offs}
\label{app:extraexp:deltasim}

\begin{table*}[ht]
\centering
\small
\caption{\textbf{\(\Delta\)Sim component analysis on accuracy, controllability, and feasibility (Waymo).}
We evaluate the impact of Hybrid Action Head (A+R) and Differentiable Kinematic Alignment Loss (DKAL).
\textit{Ctrl-Sim} baseline suffers from poor controllability ($\rho=-0.25$) and high latency.
Our full \textbf{\DeltaSim} achieves the lowest ADE (1.72m) and highest controllability ($\rho=0.53$).
Crucially, \textbf{DKAL} significantly reduces closed-loop physical violations (e.g., Lat. Acc. 12.5\% $\to$ 2.1\%), ensuring stable long-horizon rollouts.}
\label{tab:app-deltasim-analysis-redesign}
\setlength{\tabcolsep}{3.5pt}
\renewcommand{\arraystretch}{1.1}
\begin{threeparttable}
\begin{tabular}{@{}l cccc cc cccc cc@{}}
\toprule
\multirow{2}{*}{\textbf{Method}}
& \multicolumn{4}{c}{\textbf{Design toggles}}
& \multicolumn{2}{c}{\textbf{Open-loop}}
& \multicolumn{4}{c}{\textbf{Closed-loop feasibility (\%)\(\downarrow\)}}
& \multicolumn{2}{c}{\textbf{Efficiency}} \\
\cmidrule(lr){2-5}\cmidrule(lr){6-7}\cmidrule(lr){8-11}\cmidrule(lr){12-13}
& \makecell{\textbf{RTG}\\\textbf{(1-pass)}}
& \makecell{\textbf{Hybrid}\\\textbf{A+R}}
& \makecell{\textbf{Logit}\\\textbf{shaping}}
& \makecell{\textbf{DKAL}\\\textbf{loss}}
& \makecell{\textbf{ADE}\\(m)\(\downarrow\)}
& \makecell{\(\boldsymbol{\rho}\)\(\uparrow\)}
& \makecell{\textbf{Coll.}}
& \makecell{\textbf{Yaw}\\\textbf{rate}}
& \makecell{\textbf{Lat.}\\\textbf{acc.}}
& \makecell{\textbf{Lon.}\\\textbf{jerk}}
& \makecell{\textbf{Lat.}\\(ms)\(\downarrow\)}
& \makecell{\textbf{Fwd}\\\(\downarrow\)} \\
\midrule

Ctrl-Sim~\cite{rowe2024ctrlsim}
& \xmark & \xmark & \xmark & \xmark
& 2.80 & -0.25 & 15.2 & 8.5 & 14.2 & 10.5 & 36.9 & 2.0 \\

\midrule

\(\Delta\)Sim (discrete-only)
& \cmark & \xmark & \cmark & \cmark
& 2.62 & 0.39 & 8.5 & 4.2 & 5.5 & 6.2 & \best{21.7} & \best{1.0} \\

\(\Delta\)Sim (hybrid A+R, no DKAL)
& \cmark & \cmark & \cmark & \xmark
& 2.10 & 0.24 & 10.1 & 6.8 & 12.5 & 8.9 & 26.1 & \best{1.0} \\

\(\Delta\)Sim (hybrid A+R, no shaping)
& \cmark & \cmark & \xmark & \cmark
& 1.95 & 0.45 & 5.2 & 2.5 & 3.1 & 4.5 & 26.1 & \best{1.0} \\

\rowcolor{oursrow}
\textbf{\(\Delta\)Sim (Ours)}
& \cmark & \cmark & \cmark & \cmark
& \best{1.72} & \best{0.53} & \best{2.9} & \best{1.2} & \best{2.1} & \best{3.2} & 26.1 & \best{1.0} \\

\bottomrule
\end{tabular}
\end{threeparttable}
\end{table*}

\paragraph{What this ablation isolates.}
Table~\ref{tab:app-deltasim-analysis-redesign} decomposes the behavior model into four mechanisms:
single-pass return conditioning, hybrid discrete--continuous actions, inference-time logit shaping, and differentiable kinematic alignment.
We report three axes that matter specifically for long-horizon closed-loop simulation:
trajectory accuracy (ADE), controllability (\(\rho\)), and feasibility under rollout (violation rates; definitions in Section~\ref{app:metrics}).

\paragraph{Single-pass return conditioning improves efficiency without sacrificing control.}
The baseline requires two forward passes and has poor controllability (\(\rho=-0.25\)).
Our single-pass return conditioning reduces forward passes to one and improves controllability (\(\rho=0.39\) in the discrete-only variant),
indicating that the control signal is injected into the same computation that produces actions, rather than being used as an auxiliary prediction that is later re-fed.

\paragraph{Hybrid action heads remove the vocabulary-resolution bottleneck.}
The discrete-only variant has higher ADE (2.62), consistent with quantization error from a finite k-disks vocabulary.
Introducing the residual refinement head reduces ADE (2.62 \(\rightarrow\) 2.10 and below), confirming that continuous refinement improves precision
without requiring a larger discrete vocabulary (which would increase both memory and sampling entropy).

\subsubsection{Closed-loop validation beyond self-consistency}
\label{app:extraexp:closedloop}

\begin{table*}[t]
\centering
\small
\caption{\textbf{Policy training value with cross-backend evaluation (Waymo).}
To reduce the risk of simulator overfitting, we evaluate policies across different backends under matched horizons. A policy retrained in VectorWorld improves success not only in VectorWorld but also transfers to log replay and ScenDream backends (see Panel A). Panel B further evaluates long-horizon performance (\(1\,\mathrm{km}+\)) in VectorWorld.}
\label{tab:app-policy-cross-backend-redesign}
\setlength{\tabcolsep}{2pt}
\renewcommand{\arraystretch}{1.0}
\begin{threeparttable}
\begin{tabular}{@{}l l c c cc cc cc@{}}
\toprule
\multicolumn{10}{@{}l}{\textit{\textbf{Panel A: Fixed-horizon cross-backend evaluation (200\,m).}}} \\
\toprule
\multirow{2}{*}{\textbf{Policy}}
& \multirow{2}{*}{\textbf{Train backend}}
& \makecell{\textbf{Train}\\\textbf{data}}
& \makecell{\textbf{Train}\\\textbf{GPU-h}}
& \multicolumn{2}{c}{\textbf{Eval: Log Replay}}
& \multicolumn{2}{c}{\textbf{Eval: ScenDream-L (neg.)}}
& \multicolumn{2}{c}{\textbf{Eval: VectorWorld (neg.)}} \\
\cmidrule(lr){3-3}\cmidrule(lr){4-4}\cmidrule(lr){5-6}\cmidrule(lr){7-8}\cmidrule(lr){9-10}
& & \makecell{(\#eps)} & &
\makecell{\textbf{C/O/Succ}\\(\%)}
& \makecell{\textbf{Time}\\(h)}
& \makecell{\textbf{C/O/Succ}\\(\%)}
& \makecell{\textbf{Time}\\(h)}
& \makecell{\textbf{C/O/Succ}\\(\%)}
& \makecell{\textbf{Time}\\(h)} \\
\midrule

PPO
& Waymo Log
& 100k & \textbf{64}
& 29.3 / 6.9 / 63.8 & \best{0.4}
& 59.0 / 9.0 / 32.1 & 5.8
& 65.5 / 9.5 / 25.0 & 1.2 \\

PPO
& ScenDream-L
& 100k & 480
& 19.3 / \best{5.9} / 74.8 & \best{0.4}
& 45.2 / 8.5 / 46.3 & 5.8
& 40.9 / 7.6 / 48.5 & 1.2 \\

\rowcolor{oursrow}
PPO
& \textbf{VectorWorld}
& 100k & 160
& \best{18.1} / 6.2 / \best{75.7} & \best{0.4}
& \textbf{38.5 / 5.6 / 56.3} & 5.8
& \textbf{27.1 / 5.0 / 66.9} & 1.2 \\

\midrule
\multicolumn{10}{@{}l}{\textit{\textbf{Panel B: Long-horizon evaluation (1\,km\(+\); VectorWorld only).}}} \\
\midrule

\multicolumn{2}{@{}l}{\textbf{Policy (Eval: VectorWorld neg.)}}
& \multicolumn{2}{c}{\textbf{Eval time (h)}}
& \multicolumn{2}{c}{\textbf{Coll. (\%)\(\downarrow\)}}
& \multicolumn{2}{c}{\textbf{Offroad (\%)\(\downarrow\)}}
& \multicolumn{2}{c}{\textbf{Succ. (\%)\(\uparrow\)}} \\
\cmidrule(lr){3-4}\cmidrule(lr){5-6}\cmidrule(lr){7-8}\cmidrule(lr){9-10}

\multicolumn{2}{@{}l}{Pretrained PPO (Baseline)}
& \multicolumn{2}{c}{1.0}
& \multicolumn{2}{c}{65.1}
& \multicolumn{2}{c}{9.2}
& \multicolumn{2}{c}{25.7} \\

\rowcolor{improvegreen}
\multicolumn{2}{@{}l}{PPO retrained in VectorWorld (Ours)}
& \multicolumn{2}{c}{1.0}
& \multicolumn{2}{c}{\textbf{35.1}}
& \multicolumn{2}{c}{\textbf{7.9}}
& \multicolumn{2}{c}{\textbf{56.0}} \\

\bottomrule
\end{tabular}
\begin{tablenotes}[flushleft]\footnotesize
\item \textit{C/O/Succ}: Collision Rate / Offroad Rate / Success Rate.
\item \textit{Time}: Wall-clock hours for evaluating 1000 episodes. Note VectorWorld's efficiency (1.2h) vs ScenDream (5.8h).
\end{tablenotes}
\end{threeparttable}
\end{table*}

\begin{table*}[t]
\centering
\small
\caption{\textbf{Failure reproduction across backends.}
We verify if failures discovered in simulation correspond to realistic crash types.
We match generated failure scenarios to the nearest log scenarios based on map and agent topology.
VectorWorld failures show high \textbf{Type Agreement} (78.5\%) with log data, indicating that our simulator exposes realistic safety-critical events rather than simulation artifacts.}
\label{tab:app-failure-repro}
\setlength{\tabcolsep}{6.0pt}
\renewcommand{\arraystretch}{1.1}
\begin{threeparttable}
\begin{tabular}{@{}l cccc cc@{}}
\toprule
\multirow{2}{*}{\textbf{Source of failures}}
& \multicolumn{4}{c}{\textbf{Matching and reproduction diagnostics}}
& \multicolumn{2}{c}{\textbf{Outcome agreement}} \\
\cmidrule(lr){2-5}\cmidrule(lr){6-7}
& \makecell{\textbf{\#Fail}\\\textbf{episodes}}
& \makecell{\textbf{Log}\\\textbf{match (\%)}} 
& \makecell{\textbf{Init.}\\\textbf{state dist.}\\\textbf{(L2) \(\downarrow\)}} 
& \makecell{\textbf{Route}\\\textbf{match (\%)}} 
& \makecell{\textbf{Fail}\\\textbf{repro. (\%)}} 
& \makecell{\textbf{Type}\\\textbf{agreement (\%)}} \\
\midrule
Log replay (reference)  & 100 & 100.0 & 0.00 & 100.0 & 100.0 & 100.0 \\
ScenDream-L (neg. tilt) & 528 & 65.2 & 1.42 & 82.5 & 45.2 & 52.1 \\
\rowcolor{oursrow}
VectorWorld (neg. tilt) & 655 & \best{88.5} & \best{0.92} & \best{95.4} & \best{72.0} & \best{78.5} \\
\bottomrule
\end{tabular}
\end{threeparttable}
\end{table*}

\begin{figure}[t]
  \centering
  \includegraphics[width=0.95\linewidth]{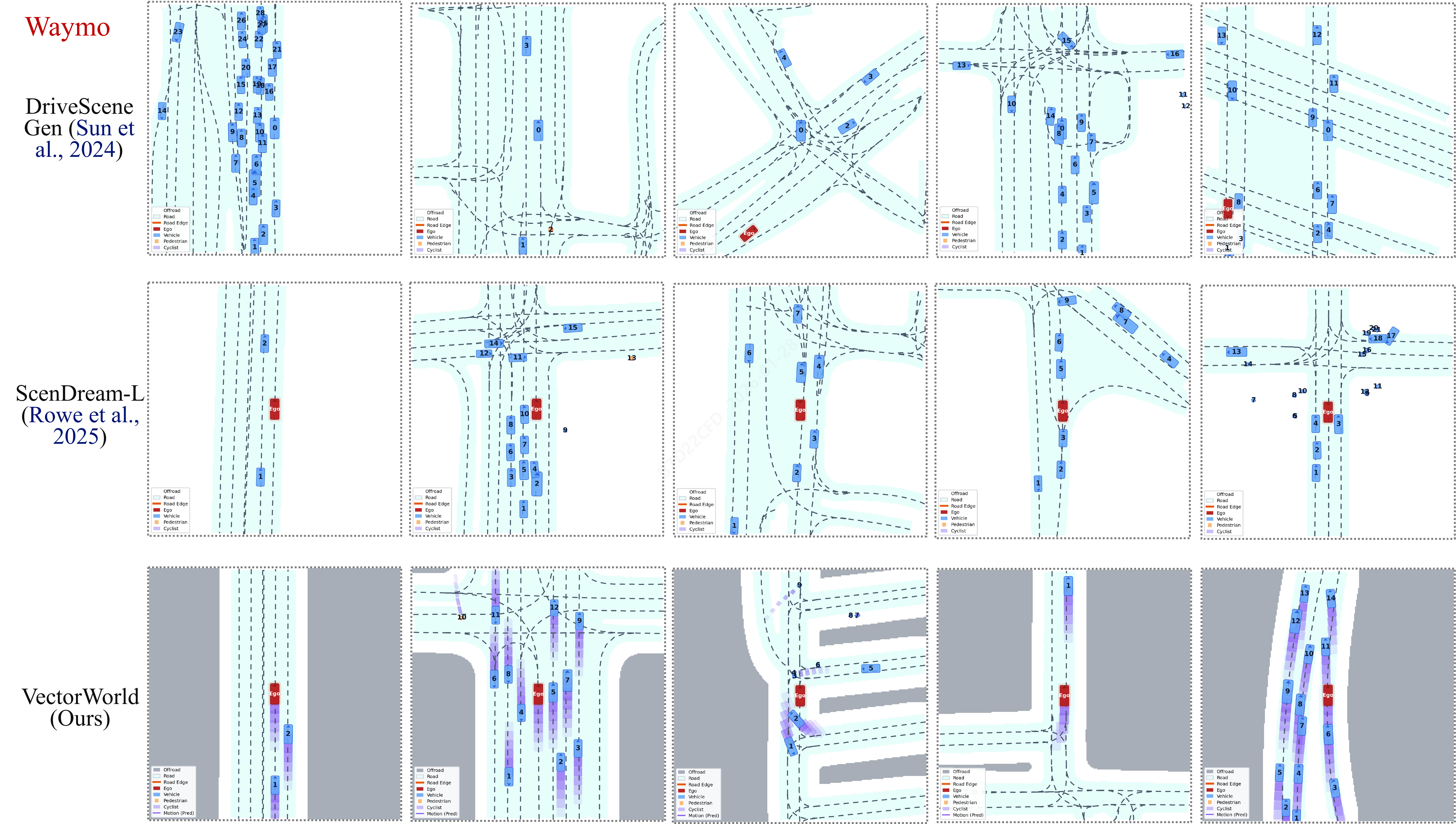}
  \caption{\textbf{Waymo qualitative comparison for initial scene generation (64m $\times$ 64m).}
  We compare VectorWorld against representative vectorized scene generators and the Waymo reference.
  The visualization highlights lane connectivity at intersections, agent--lane geometric consistency, and the plausibility of
  the initialized interaction state (including short motion history when available).}
  \label{fig:app:waymo_compare}
\end{figure}
\begin{figure}[t]
  \centering
  \includegraphics[width=\linewidth]{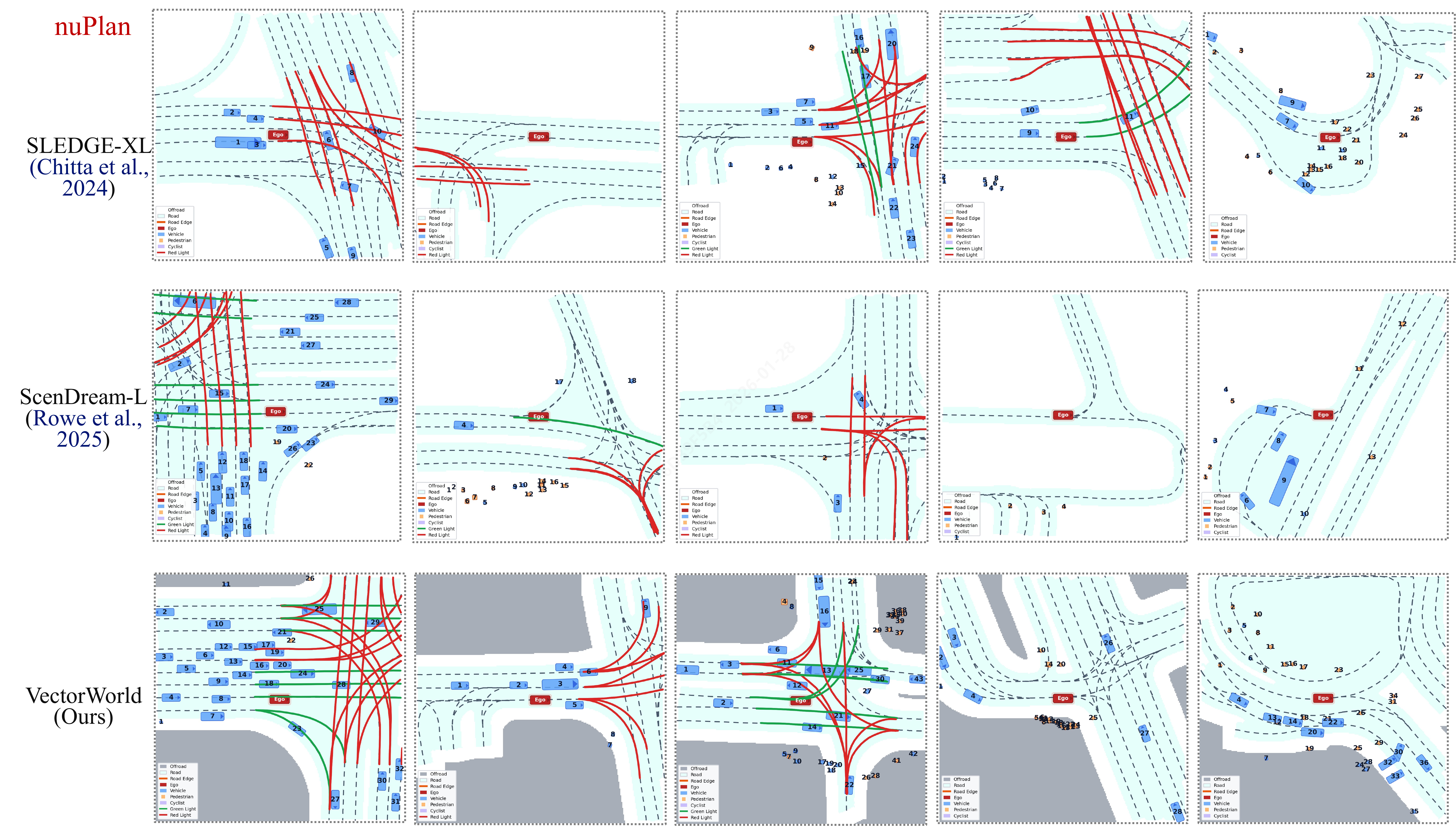}
  \caption{\textbf{nuPlan qualitative comparison with traffic-light semantics.}
  Lane centerlines are colored by traffic-light state (green vs.\ red) to visualize control-relevant map semantics.
  We compare VectorWorld with strong baselines under the same vectorized rendering.}
  \label{fig:app:nuplan_compare}
\end{figure}

\paragraph{Why hybrid actions require physics-aligned training.}
A hybrid head increases expressivity; without a feasibility-aligned learning signal, the model can allocate probability mass to precise but physically implausible
transitions (e.g., high lateral acceleration) that are weakly penalized by pure imitation losses.
This is visible in the hybrid-no-DKAL row, where feasibility violations increase sharply (e.g., lateral acceleration 12.5\%) even though ADE improves.
This pattern indicates that expressivity alone is insufficient; it must be paired with an inductive bias that matches the closed-loop feasibility objective.

\paragraph{DKAL provides train--infer alignment for feasibility.}
Adding DKAL substantially reduces feasibility violations and improves overall stability of rollouts.
Notably, DKAL reduces violations even when logit shaping is disabled, indicating that the model learns to rank actions in a feasibility-consistent way,
rather than relying on inference-only heuristics.
When both DKAL and shaping are enabled, the model achieves the best combined operating point: lowest ADE, highest controllability, and lowest violation rates.

\paragraph{Implication for long-horizon stability.}
Feasibility violations are a compounding-error mechanism: small rates per step become dominant over kilometer-scale horizons.
The reductions in yaw-rate and lateral-acceleration violations therefore directly explain the improved stability observed in the long-horizon closed-loop results.

\paragraph{Cross-backend evaluation addresses self-consistency.}
A common threat to validity for learned simulators is self-consistency: a policy might appear strong only because it exploits simulator-specific artifacts.
Table~\ref{tab:app-policy-cross-backend-redesign} evaluates the same policies across multiple backends under matched horizons.
A policy retrained in VectorWorld improves success not only in VectorWorld but also transfers to log replay and ScenDream backends, indicating that the learned
improvements are not tied to a single simulator implementation.

\paragraph{Why transfer is expected in our design.}
VectorWorld reduces two major sources of simulator-specific bias:
(i) initialization mismatch, by providing motion-history warm starts consistent with the behavior-model context,
and (ii) structural drift, by enforcing relational consistency in generation and feasibility alignment in behavior.
These design choices reduce the incentive for policies to overfit to artifacts (e.g., unrealistic transient dynamics or topology breakage),
which in turn improves transfer.

\paragraph{Long-horizon evaluation is a differentiator, not a cosmetic extension.}
Panel B of Table~\ref{tab:app-policy-cross-backend-redesign} shows that the retrained policy maintains substantially higher success over a 1\,km scale horizon.
This regime is precisely where compounding errors become dominant and where a simulator must be evaluated as a system rather than as a static generator.

\paragraph{Failure realism diagnostics.}
Table~\ref{tab:app-failure-repro} provides a failure-reproduction diagnostic: we match simulator-discovered failures to the nearest logged scenarios under map and
agent-topology similarity, and measure agreement in failure type.
VectorWorld achieves higher matching rates and higher type agreement than the baseline generator, indicating that the discovered failures are structurally similar
to real-world events rather than being dominated by simulator-specific artifacts.
This supports using VectorWorld as a stress-test backend for policy improvement rather than only as a diversity tool.

\subsection{Qualitative Results and Visual Analysis}
\label{app:vis}

This section provides qualitative visualizations that complement the quantitative metrics in
\Cref{tab:main_performance,tab:detailed_metrics,tab:closed_loop_eval}.
We organize the analysis around the same deployment constraints that motivate VectorWorld:
(i) topology-consistent initialization,
(ii) controllability under fixed map conditions,
(iii) real-time end-to-end latency for online generation,
and (iv) long-horizon stability under repeated streaming outpainting.

\paragraph{Rendering protocol (common to all figures).}
Unless otherwise stated, all panels are rendered from the decoded vector-graph representation:
lanes are shown as centerline polylines, agents are shown as oriented boxes, and (when available) short motion histories are shown as
agent-frame polylines re-projected to the ego frame at initialization.
Importantly, the visualizations do not rely on manual editing; they reflect the decoded simulator state after the same deterministic
validity filters described in \Cref{app:simgen}.
This makes the qualitative observations directly comparable to the offline validity and topology metrics.

\begin{figure}[t]
  \centering
  \includegraphics[width=\linewidth]{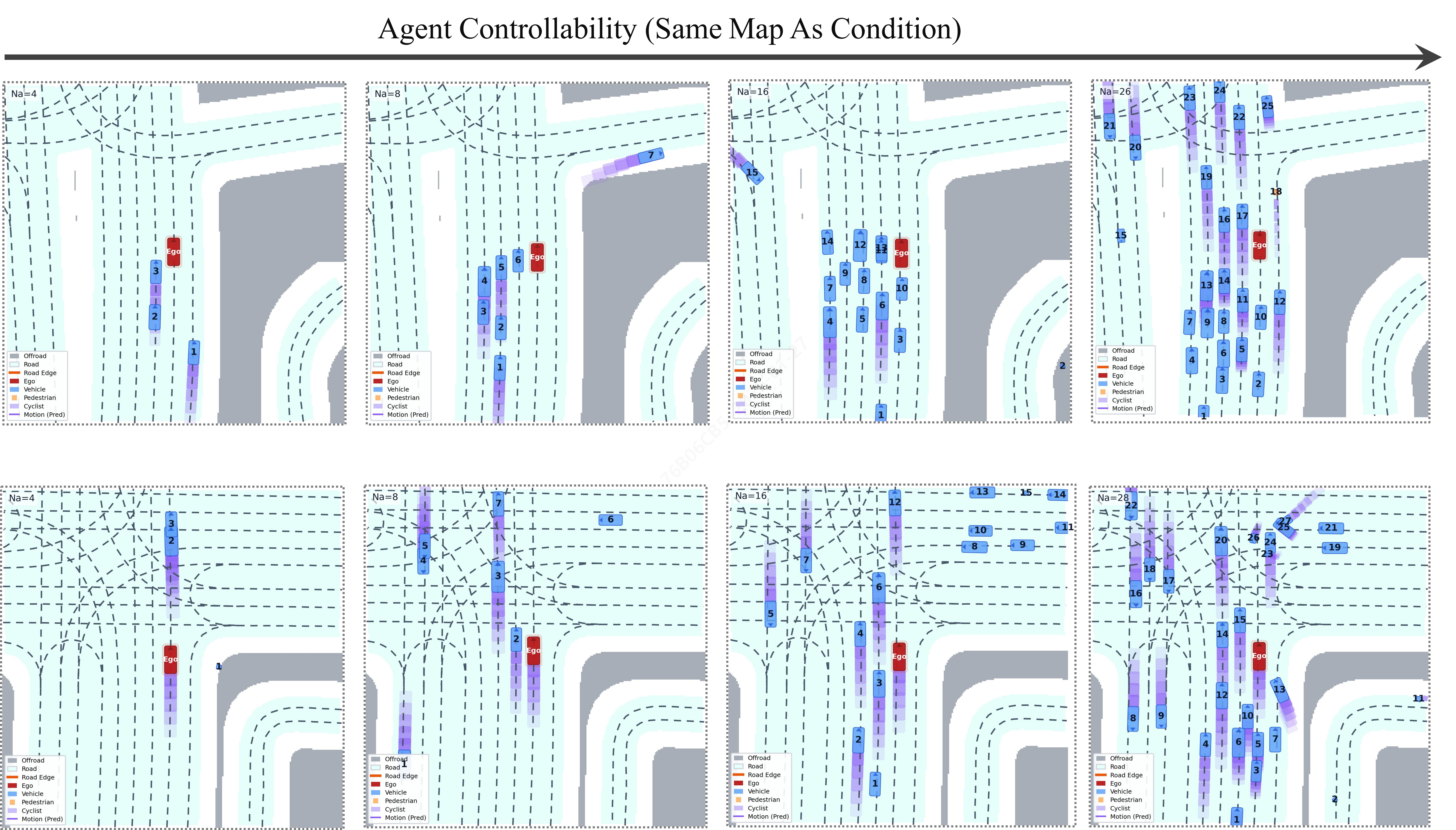}
  \caption{\textbf{Agent controllability under fixed map conditioning.}
  The map (lane graph) is held fixed as a condition, while the agent configuration is generated under different controllability settings.
  This visualization isolates controllability from map diversity by using the same conditioned lanes across all panels.}
  \label{fig:app:lane_condition}
\end{figure}

\begin{figure}[t]
  \centering
  \includegraphics[width=\linewidth]{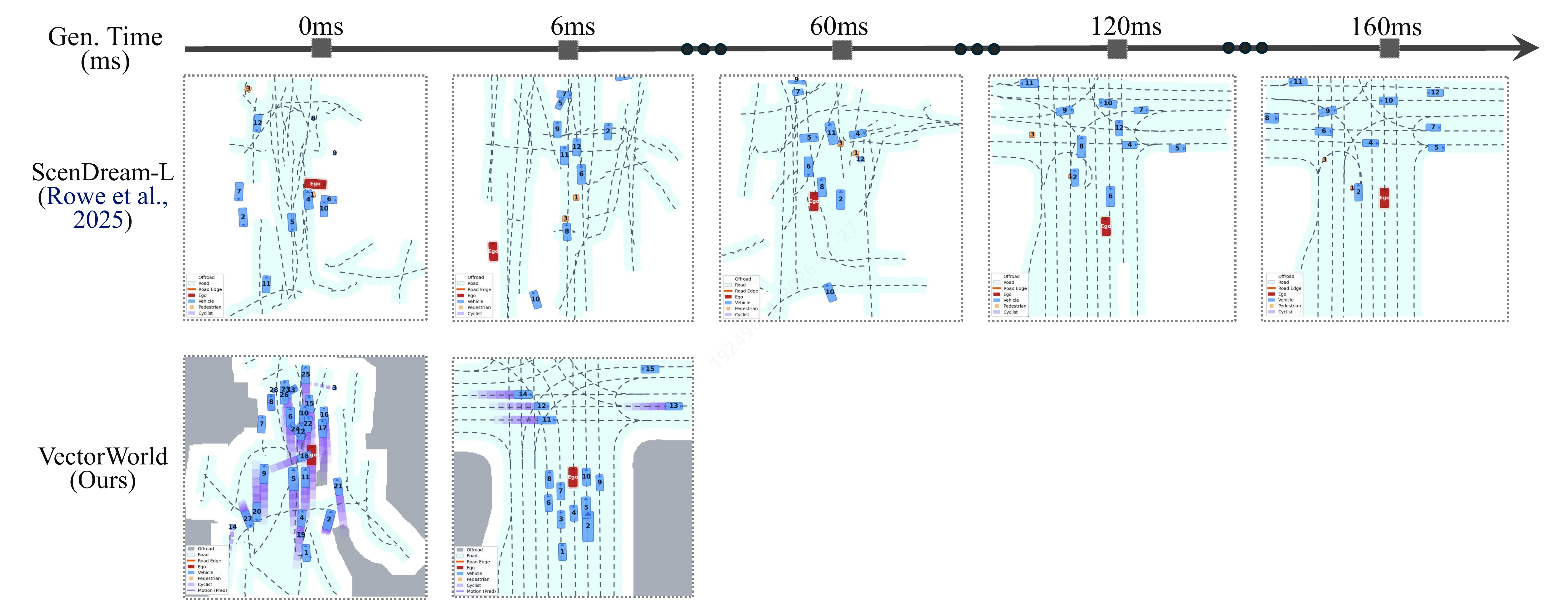}
  \caption{\textbf{End-to-end generation latency (timeline view).} Qualitative snapshots under increasing time budgets for a \(64\,\mathrm{m}\times 64\,\mathrm{m}\) tile. VectorWorld reaches the final lane--agent vector scene with one solver-free MeanFlow step plus one VAE decode (\(\approx 6\)\,ms), whereas multi-step diffusion baselines only converge at substantially higher budgets (e.g., \(\sim 160\)\,ms).}
  \label{fig:app:latency}
\end{figure}
\begin{figure}[t]
  \centering
  \includegraphics[width=\linewidth]{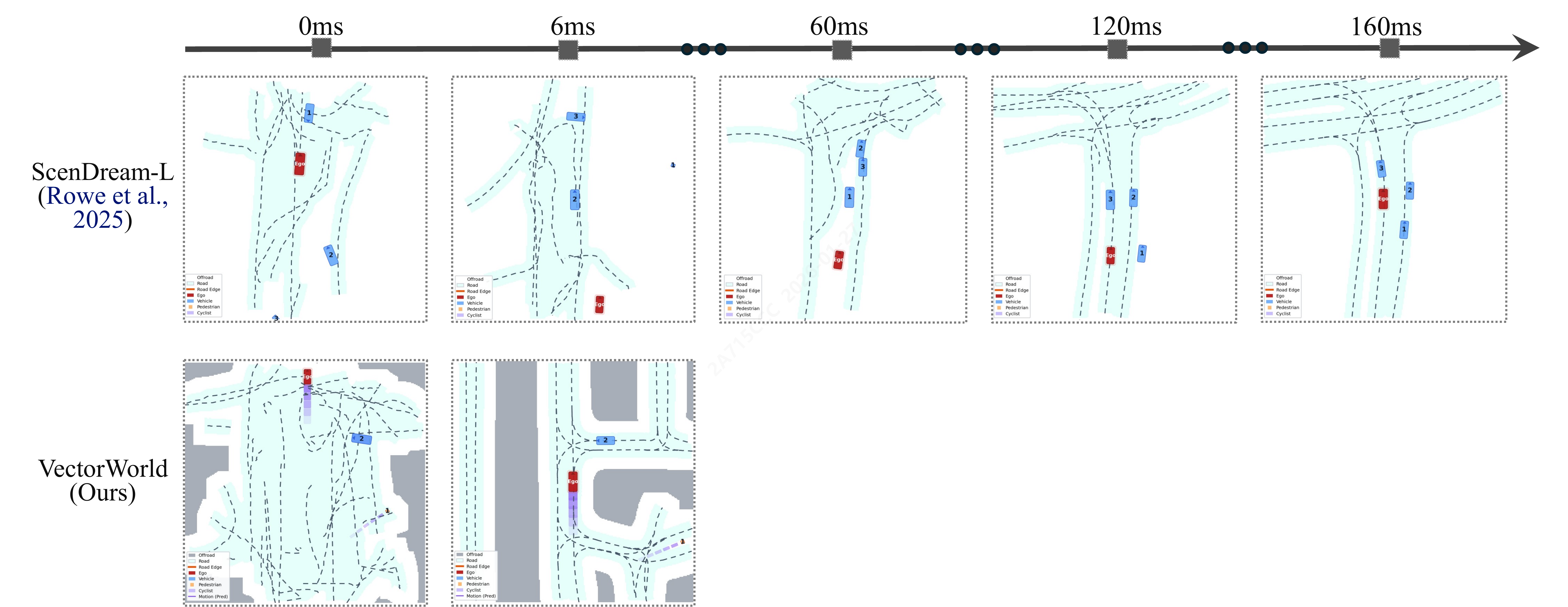}
  \caption{\textbf{End-to-end generation latency.} A different scene instance rendered with the same timing protocol (latent sampling + one VAE decode). The result is consistent: VectorWorld completes a full \(64\,\mathrm{m}\times 64\,\mathrm{m}\) lane--agent tile in \(\approx 6\)\,ms via a single MeanFlow update, while solver-based diffusion requires many iterative denoising calls.}
  \label{fig:app:latency2}
\end{figure}

\subsubsection{Waymo: initial-scene generation comparisons}
\label{app:vis:waymo}

\paragraph{Topology consistency at intersections is an edge-level property.}
The Waymo panels in \Cref{fig:app:waymo_compare} illustrate a recurrent failure mode of vector-graph generation: local lane geometry can appear plausible,
yet successor transitions become discontinuous or inconsistent at intersections and merges.
This is precisely an edge-defined constraint: connectivity is encoded by \(\mathcal{E}_{\mathrm{L2L}}\), not by marginal lane-node appearance.
VectorWorld reduces such discontinuities by explicitly modulating attention with typed edge attributes
(\Cref{sec:dit,eq:rel_attn}), which is consistent with the improvements in endpoint distance and route validity in
\Cref{tab:main_performance,tab:detailed_metrics}.
Qualitatively, this generates lane graphs that remain traversable across junctions without relying on heuristic stitching.

\paragraph{Agent--lane alignment and initialization feasibility co-vary in closed loop.}
The same figure also reveals that agent placement and orientation errors are not isolated artifacts: misalignment to nearby lane tangents
(e.g., heading inconsistent with centerline direction) increases the probability of overlaps and early collisions after the first few simulator steps.
VectorWorld’s generator benefits from cross-type relational modeling, which conditions agent tokens on lane context and vice versa,
making agent orientation and lane assignment more consistent at initialization.
This qualitative pattern matches the lower initialization collision rate in \Cref{tab:main_performance} and the improved angular/size/speed
distribution alignment in \Cref{tab:detailed_metrics} (Waymo).

\paragraph{Why interaction-state initialization is visually diagnostic (beyond FD).}
A key difference between VectorWorld and snapshot-only generators is that VectorWorld instantiates each agent with a compact motion-history code
(\Cref{sec:vae}).
While FD primarily measures distribution alignment in an embedding space, the \InteractionState\ determines whether the downstream
behavior model receives in-distribution temporal context at \(t=0\).
In \Cref{fig:app:waymo_compare}, VectorWorld’s initialized agents exhibit consistent short-horizon motion cues, which reduces the “zero-history” prior
implicitly assumed by history-conditioned policies.
This mechanism is causally linked to the reduced jerk and improved closed-loop success when WarmStart is enabled
(\Cref{tab:closed_loop_eval}, Section I).

\subsubsection{nuPlan: traffic-light semantics and lane--agent consistency}
\label{app:vis:nuplan}

\paragraph{Traffic-light semantics create structured conditional dependencies.}
Traffic signals in nuPlan induce a constraint that couples distant elements:
the lane state (red/green) constrains which lane segments are currently traversable, which in turn constrains feasible agent intent and placement
(e.g., queue formation at red lights vs.\ flowing traffic on green).
This dependency is not captured by purely local appearance matching, and is particularly brittle under repeated generation because
a single inconsistent lane state can propagate into unrealistic agent motion in closed loop.

\paragraph{VectorWorld preserves signal-conditioned lane structure and its downstream implications.}
In \Cref{fig:app:nuplan_compare}, VectorWorld produces lane centerlines whose traffic-light coloring is consistent across the intersection structure,
and the resulting agent configurations remain compatible with the implied right-of-way constraints.
Mechanistically, this is expected from two design choices:
(i) the heterogeneous graph representation makes lane semantics explicit at the node/edge level (\Cref{sec:repr}),
and (ii) edge-gated relational attention provides a controlled pathway for propagating lane semantics into cross-type interactions (\Cref{sec:dit}).
Qualitatively, this generates fewer cases where agents are initialized with headings or lane assignments that contradict the current signal state.

\paragraph{Why this matters for closed-loop evaluation.}
Signal-conditioned inconsistencies can appear benign in single-frame offline metrics, yet they directly affect closed-loop behavior:
a history-conditioned planner reacts strongly to right-of-way cues, and inconsistent initialization can trigger abrupt braking or unsafe merges.
The traffic-light visualization therefore serves as a targeted diagnostic for “control-relevant realism,” complementing FD/topology metrics.
It also explains why we emphasize structure and safety metrics alongside perceptual alignment in \Cref{tab:main_performance}.

\subsubsection{Map-conditioned controllability of agent generation}
\label{app:vis:control}
\paragraph{Controllability should be evaluated under a fixed map to avoid confounding.}
A common confound in scenario generation is that apparent “control” can be explained by changes in the generated map itself
(e.g., different lane topology induces different feasible behaviors).
\Cref{fig:app:lane_condition} fixes the lane graph as a condition and varies only the controllability input that affects agent generation.
This isolates whether the model can modulate agent intent and configuration while maintaining compatibility with the same geometric constraints.

\paragraph{Why VectorWorld supports map-conditioned control.}
VectorWorld’s interface separates (i) conditioned structure (clamped lane tokens) from (ii) sampled content (agent tokens),
and uses a unified masked-completion mechanism in latent space (\Cref{sec:repr,sec:streaming}).
As a result, the generator can keep lane topology invariant while sampling alternative agent sets that remain aligned with lane geometry.
The \MotionVAE\ further stabilizes the controllability effect by representing agent initialization as an interaction state:
changing the control input modifies not only the instantaneous pose distribution, but also the implied short-horizon momentum at \(t=0\),
which is the quantity that reactive NPC policies and ego planners are most sensitive to.

\paragraph{Feasibility is a necessary complement to controllability.}
Controllability is only useful if the resulting agent configurations remain feasible under rollout.
In VectorWorld, feasibility is enforced at two levels:
(i) the generator improves initialization validity (lower static collision rate), and
(ii) \(\Delta\)Sim suppresses kinematically infeasible actions over long horizons via physics-aligned learning and logit shaping
(\Cref{sec:deltasim,eq:deltasim_shaping,eq:deltasim_dkal}).
This is consistent with the fact that control sweeps in \Cref{tab:closed_loop_eval} can systematically modulate failure rates
without requiring shorter rollouts or early termination due to simulator instability.

\subsubsection{End-to-end latency comparisons (64m \(\times\) 64m scenes)}
\label{app:vis:latency}

\paragraph{Measurement protocol.}
We report wall-clock latency per \(64\,\mathrm{m}\times 64\,\mathrm{m}\) scene with GPU synchronization.
The timing includes (i) latent sampling and (ii) exactly one VAE decode into the vector-graph representation; it excludes data loading and any optional deterministic filtering so the reported numbers isolate model inference cost.

\paragraph{Key result.}
As shown in \Cref{fig:app:latency,fig:app:latency2}, \VectorWorld\ operates in the single-digit millisecond regime:
one MeanFlow update at \((t,\Delta)=(1,1)\) followed by one decode yields a complete lane--agent tile in \(\approx 6\)\,ms, (\Cref{app:vis:latency}).

\paragraph{Why the speedup is structural (not an implementation artifact).}
Our generator is explicitly trained for solver-free large-step transport and deployed with the one-step update
\(z_{0}=z_{1}-u_{\theta}^{\mathrm{cfg}}(z_{1};t{=}1,\Delta{=}1,c)\) (\Cref{eq:mf_update}), so inference cost is essentially one DiT forward pass plus one decode.
In contrast, diffusion-style baselines must iterate many denoising steps, each invoking a comparable backbone; therefore, latency scales roughly linearly with the solver step budget even when the decoder and output representation are matched.
This end-to-end latency is the binding constraint in streaming outpainting, where generation is invoked repeatedly during closed-loop rollout.

\subsubsection{Real-time streaming outpainting case studies (1km+)}
\label{app:vis:streaming}

\begin{figure}[t]
  \centering
  \includegraphics[width=\linewidth]{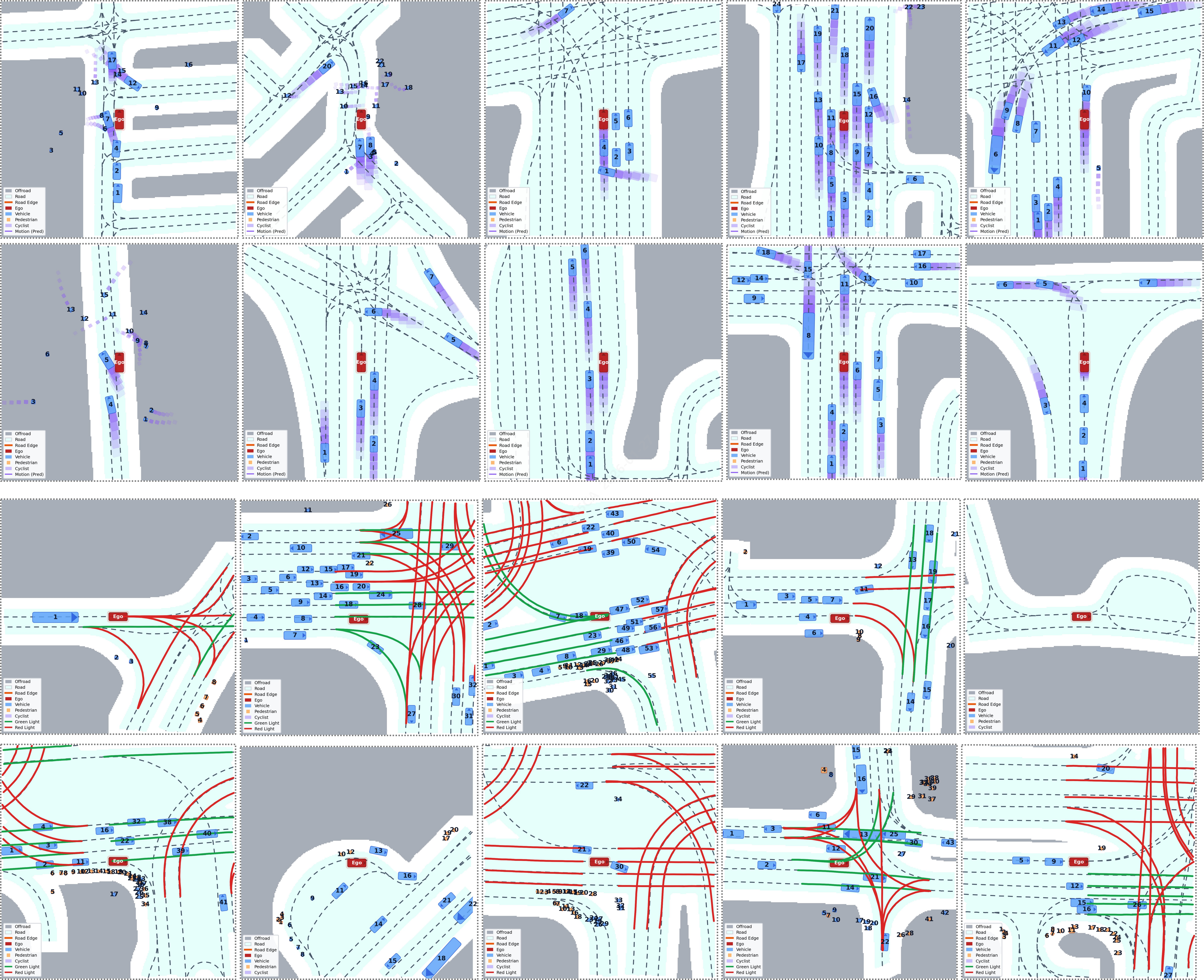}
  \caption{\textbf{Planner-relevant challenging scenarios sampled from \VectorWorld.}
  We render a subset of VectorWorld-generated scenes using the protocol in \Cref{app:vis}.
  These scenarios illustrate diverse intersection geometries, dense multi-agent interactions, and tight gaps that are valuable for stress testing and training an ego PPO planner in closed loop.}
  \label{fig:vis:gen_valuable_scenario}
\end{figure}

\paragraph{Challenging scenarios mined from VectorWorld generation.}
Figure~\ref{fig:vis:gen_valuable_scenario} shows representative scenarios selected from VectorWorld-generated environments.
These cases feature dense interactions, complex intersection topology, and/or tight spatiotemporal gaps that empirically stress a fixed PPO ego planner.
They complement the system-level statistics in \Cref{tab:closed_loop_eval} by visualizing the kinds of planner-relevant edge cases enabled by streaming outpainting.

\begin{figure}[t]
  \centering
  \includegraphics[width=\linewidth]{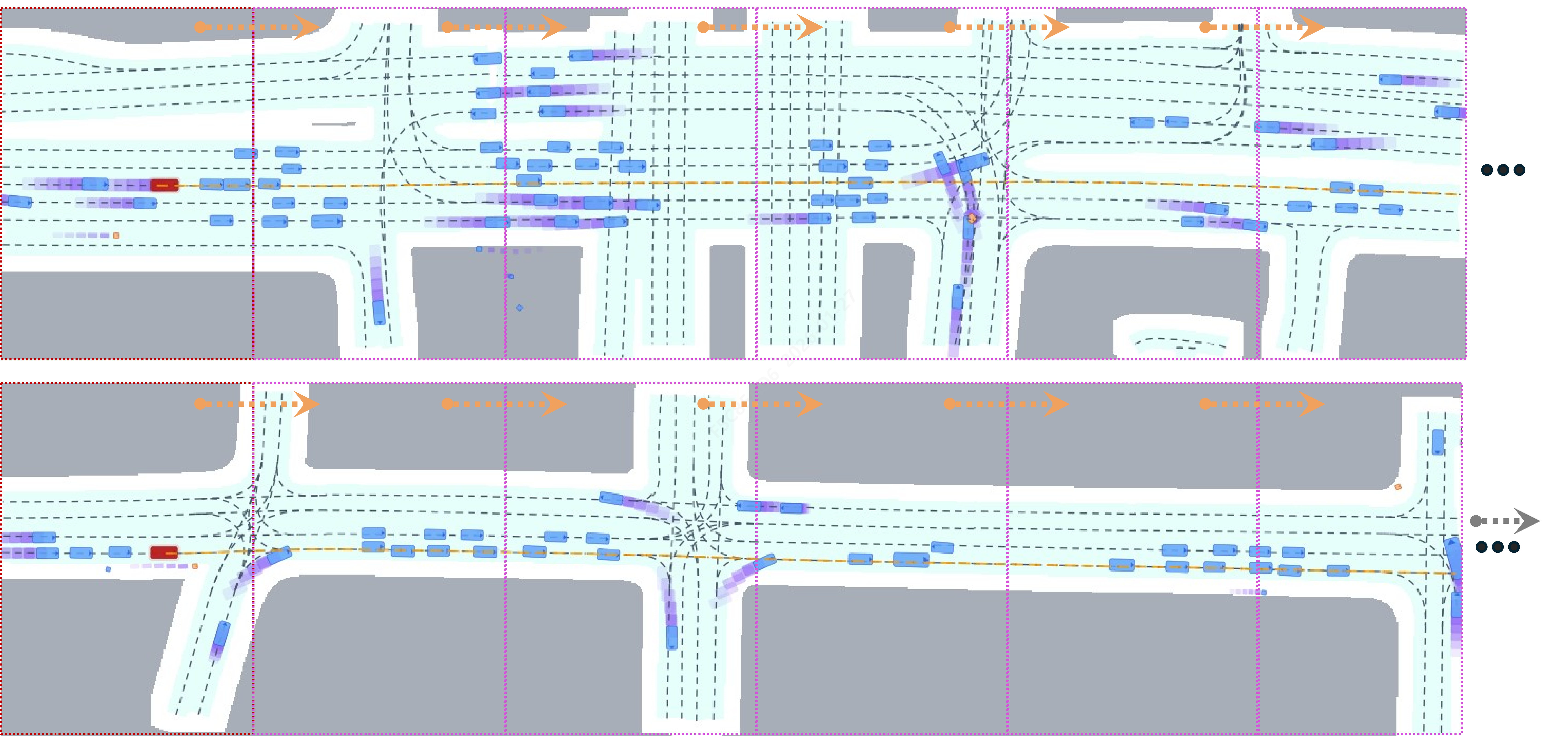}
  \caption{\textbf{Streaming outpainting in closed-loop rollout (case study 1).}
  The simulator repeatedly outpaints new lane--agent tiles near the frontier while clamping the observed region.
  The visualization highlights seam consistency and stable long-horizon progression.}
  \label{fig:app:streaming_a}
\end{figure}

\begin{figure}[t]
  \centering
  \includegraphics[width=\linewidth]{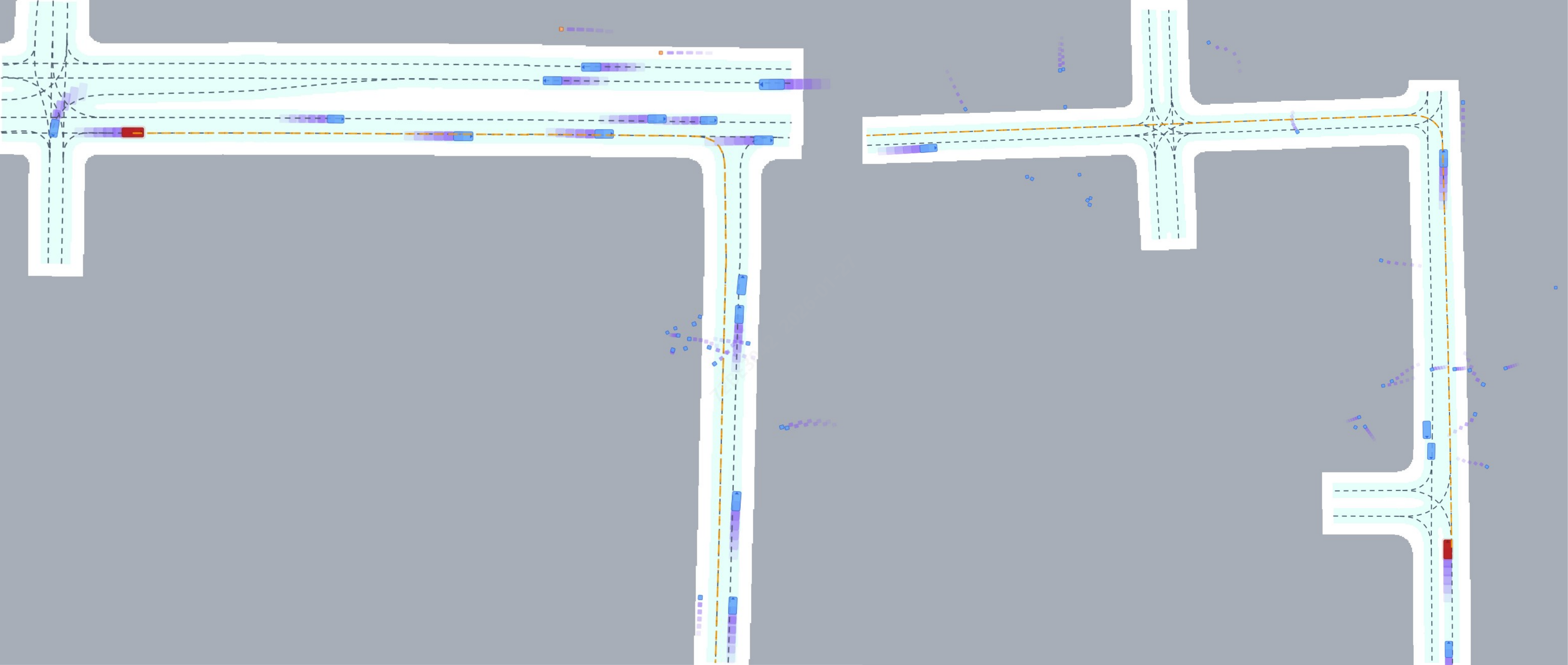}
  \caption{\textbf{Streaming outpainting with dynamic agent set and per-agent intent control (case study 2).}
  The visualization illustrates stable long-horizon expansion with agent insertion/removal and controllable behaviors,
  enabled by the \InteractionState\,.}
  \label{fig:app:streaming_b}
\end{figure}

\paragraph{Streaming exposes failure modes that single-tile generation cannot reveal.}
A single generated tile (e.g., 200m) can look plausible even if small structural errors exist.
Streaming outpainting amplifies these errors because the system repeatedly conditions on previously generated content.
This makes seam consistency, topology preservation, and long-horizon feasibility the dominant factors, rather than single-frame perceptual alignment.

\paragraph{Why clamping + constant-path conditioning reduces structural drift.}
VectorWorld’s outpainting clamps previously observed latents and enforces a constant probability path for conditioned tokens
(\Cref{sec:meanflow,eq:mf_path} and the masked-token construction).
This mechanism prevents conditioned structures from drifting over repeated updates.
In \Cref{fig:app:streaming_a,fig:app:streaming_b}, the lane graph remains continuous across tile boundaries,
which is consistent with the strong endpoint-distance and route-length metrics reported in the main paper.

\paragraph{Dynamic agent management and intent control require an \InteractionState.}
In long-horizon rollouts, the active agent set must change: agents behind the ego should be removed, and new agents should appear ahead.
However, naive insertion based on static states tends to introduce cold-start artifacts (non-physical jerks) that destabilize interactions.
VectorWorld mitigates this by generating agents as interaction states with motion-history codes (\Cref{sec:vae}),
so newly inserted agents can be warm-started with implicit momentum consistent with the behavior model’s context window.
This is qualitatively visible as stable agent motion around insertion events, and quantitatively reflected by reduced jerk with WarmStart
(\Cref{tab:closed_loop_eval}).

\paragraph{Why long-horizon stability depends on feasibility-aligned behavior modeling.}
Even if maps are outpainted consistently, long-horizon stability is ultimately limited by the compounding rate of kinematic violations in NPC actions.
\(\Delta\)Sim reduces these compounding errors via hybrid action tokenization and feasibility-aligned learning (\Cref{sec:deltasim}).
The streaming case studies therefore serve as a system-level validation: they demonstrate that the generator (\(\sim\)ms latency),
the interface (warm-started interaction states), and the behavior model (\(\Delta\)Sim) jointly satisfy the constraints required for
kilometer-scale closed-loop simulation.

\subsection{Limitations and Safety Considerations}
\label{app:limitations}

\paragraph{Representation coverage.}
VectorWorld currently operates on a compact centerline-based map representation and does not model all map elements
(e.g., detailed road boundaries, curb geometry, and some fine-grained traffic control semantics).
As a result, certain planner-relevant constraints may be under-specified in the generated environment, especially in dense urban layouts.

\paragraph{One-step streaming trade-offs.}
Our design prioritizes solver-free large-step sampling for streaming outpainting.
While this enables real-time deployment, it can leave residual gaps in fine-grained distribution matching.
For example, VectorWorld does not minimize every marginal agent JSD term on nuPlan (e.g., Near./Lat. in Table~\ref{tab:detailed_metrics}),
and the Conve. topology discrepancy can remain higher than some baselines.
These gaps motivate improving fine-grained alignment under the same streaming constraint, rather than relaxing latency via multi-step solvers.

\paragraph{Behavior-model fidelity under extreme maneuvers.}
\(\Delta\)Sim uses delta-based updates and feasibility-aligned shaping to reduce kinematic violations, but it does not guarantee strict physical feasibility
under all extreme conditions (e.g., rare high-speed evasive maneuvers).
Closed-loop results should therefore be interpreted as the fidelity of the proposed generative-simulation pipeline under the stated assumptions and thresholds.

\paragraph{Safety considerations and intended use.}
VectorWorld is designed for research on simulation-based evaluation and training, not for direct deployment in real vehicles.
Generated scenarios may contain rare, safety-critical interactions; this is useful for stress testing but also creates a misuse risk if employed to
design adversarial traffic behaviors outside a controlled research context.
We recommend (i) using VectorWorld-generated scenarios only within sandboxed simulation, (ii) reporting configured feasibility thresholds and filtering rules,
and (iii) validating any learned policy improvements with cross-backend or log-replay evaluation (e.g., Appendix~\ref{app:extraexp:closedloop})
to reduce simulator-specific overfitting.

\end{document}